%% file: arxiv_version.tex
\documentclass{article}

\usepackage[preprint]{neurips_2026}

\usepackage[utf8]{inputenc}
\usepackage[T1]{fontenc}
\usepackage{microtype}

\usepackage{xcolor}
\usepackage{colortbl}
\usepackage{tcolorbox}

\definecolor{colorA}{HTML}{1B9E77}
\definecolor{colorB}{HTML}{D95F02}
\definecolor{colorC}{HTML}{7570B3}
\definecolor{colorD}{HTML}{E7298A}
\definecolor{colorE}{HTML}{66A61E}
\definecolor{colorF}{HTML}{E6AB02}
\definecolor{colorG}{HTML}{A6761D}
\definecolor{colorH}{HTML}{666666}
\definecolor{fuchsia}{rgb}{1.0, 0.0, 1.0}
\definecolor{headergray}{gray}{0.95}
\definecolor{ourmethodgreen}{HTML}{E6F5E6}
\definecolor{sotablue}{HTML}{EBF5FF}
\definecolor{bestyellow}{HTML}{FFF9C4}
\definecolor{darkviolet}{RGB}{75,0,130}
\definecolor{softblue}{RGB}{70,130,180}
\definecolor{darkerteal}{RGB}{0,128,128}
\definecolor{forest}{RGB}{34,139,34}
\definecolor{slate}{RGB}{112,128,144}
\definecolor{navy}{RGB}{0,51,102}
\definecolor{colorE}{HTML}{66A61E}

\definecolor{darkgreen}{RGB}{27,120,55}
\definecolor{lightgreen}{RGB}{120,198,121}
\definecolor{darkblue}{RGB}{33,102,172}
\definecolor{lightblue}{RGB}{146,197,222}

\usepackage{amsmath,amssymb,amsfonts}
\usepackage{nicefrac}
\usepackage{siunitx}

\usepackage{graphicx}
\usepackage{booktabs}
\usepackage{multirow}
\usepackage{wrapfig}
\usepackage{sidecap}
\sidecaptionvpos{figure}{t}
\usepackage{subcaption}

\usepackage{enumitem}
\usepackage{soul}
\usepackage{comment}

\usepackage{url}
\usepackage{hyperref}
\hypersetup{
  colorlinks=true,
  citecolor=blue,
}

\usepackage{algorithm}
\usepackage{algpseudocode}

\usepackage{tikz}
\usepackage{pgfplots, pgfplotstable}
\usetikzlibrary{positioning, fit, shapes.geometric, bending, decorations.text, matrix, calc}
\usetikzlibrary{pgfplots.groupplots}
\usepgfplotslibrary{colorbrewer}
\pgfplotsset{compat=1.15}

\pgfplotsset{
  /pgfplots/legend image code/.code={%
    \draw[mark repeat=2,mark phase=2,#1]
      plot coordinates {(0cm,0cm) (0.19cm,0cm) (0.38cm,0cm)};
  },
}

\pgfplotsset{
  cycle list/Dark2,
  cycle multiindex* list={mark list*\nextlist Dark2\nextlist},
}

\pgfplotsset{every x tick label/.append style={font=\tiny, yshift=0.5ex}}
\pgfplotsset{every y tick label/.append style={font=\tiny, xshift=0.5ex}}

\usepgfplotslibrary{fillbetween}

\input{math_commands.tex}

\setlist[itemize]{leftmargin=.5cm}

\title{Not All Synthetic Data Is Yours to Learn From}

\author{%
  Sina Alemohammad\thanks{Corresponding author: \href{mailto:sina.alemohammad@austin.utexas.edu}{sina.alemohammad@austin.utexas.edu}} \\
  ECE Department \\
  The University of Texas at Austin \\
  \And
  Li Chen \\
  Apple \\
  \And
  Richard G.\ Baraniuk \\
  ECE Department \\
  Rice University \\
  \And
  Zhangyang Wang \\
  ECE Department \\
  The University of Texas at Austin \\
}

\begin{document}

\maketitle

\begin{abstract}

\textit{Can a language model improve from plain text sampled from itself, with \textbf{no} prompts, \textbf{no} teacher, \textbf{no} verifier, and \textbf{no} reward model}? Yes, but only when the synthetic corpus is compatible with the student, a relational property of the source-student pair rather than an intrinsic property of the data. We call this the \emph{latent capability resurfacing hypothesis}: weak self-training can amplify capabilities already present in the pretrained model, but only under this compatibility condition. We study this in the minimal setting of \emph{prompt-free unconditional self-training}, where base language models are fine-tuned on text generated from the BOS token alone, with no task specification or external supervision. We report three findings. \textit{First}, synthetic utility is relational rather than intrinsic: self-generated data is the most effective source, same-lineage transfer outperforms stronger but differently trained sources, and cross-family transfer is substantially weaker. \textit{Second}, common intrinsic proxies fail: neither benchmark-level semantic similarity nor average per-token likelihood under the student predicts which corpora help. \textit{Third}, this regime produces a surprising byproduct. In controlled Pythia experiments, capability and verbatim memorization decouple: benchmark utility is preserved or improved while held-out exact-match extraction drops by over 95\%, with no forget set, privacy objective, or targeted unlearning. Together, these results suggest that prompt-free self-training works by amplifying what the student already knows, not by importing structure from the data. They also reveal a regime in which capability and verbatim memorization can be separated without any explicit unlearning objective. \blfootnote{Code is available at \url{https://github.com/VITA-Group/latent-capability-resurfacing}.}


\end{abstract}

\section{Introduction}
\label{sec:intro}

Training a language model on its own outputs is supposed to fail. Recursive self-training can produce distributional drift and eventual collapse~\citep{shumailov2023curse,alemohammad2024self}, and modern post-training pipelines typically treat synthetic data as unreliable unless it is filtered through verification, reward models, or correctness checks~\citep{feng2025beyond,zelikman2022star,singh2024beyond}. Yet a growing body of work complicates this picture. Simple self-distillation on raw unverified generations substantially improves code generation without any filter~\citep{zhang2026selfdistill}. Reward-free self-training on reasoning responses yields gains without external verifiers~\citep{li2025selftrain}. Pure inference-time sampling recovers much of the benefit traditionally attributed to reinforcement learning on reasoning benchmarks~\citep{karan2025reasoning}. In the vision domain, unverified self-generated samples can improve generative models beyond their training distribution~\citep{alemohammad2024selfimproving,alemohammad2025neon}. Across these disparate settings, a pattern emerges: weak, unstructured training signals with no task-specific supervision and no explicit teacher can still help. Something already present in the pretrained model is being accessed or amplified.

Several of these works gesture, independently, at the same candidate explanation: that weak self-training \emph{resurfaces latent capabilities} already encoded during pretraining, rather than injecting new knowledge. The evidence, however, is scattered across prompt-conditioned code, reward-style reasoning, and inference-only sampling; each setting introduces its own confounds. \textit{What has been missing} is a controlled setting in which task prompts, external reward, and explicit teacher annotations are absent by construction, so that any observed gain must arise from the interaction between a pretrained student and a weak self-generated signal. 

We therefore study the simplest case we know: continuing training of base language models on plain text generated \emph{unconditionally} from the BOS token, with no prompts, no task specifications, no filtering, and no external supervision. Any reasoning traces, benchmark-family structure, or factual content in this data must arise endogenously from the base model rather than from the data collection procedure. If latent capability resurfacing is real, this is where it should appear in its cleanest form.

\begin{figure}[t]
  \centering
  \includegraphics[width=\textwidth]{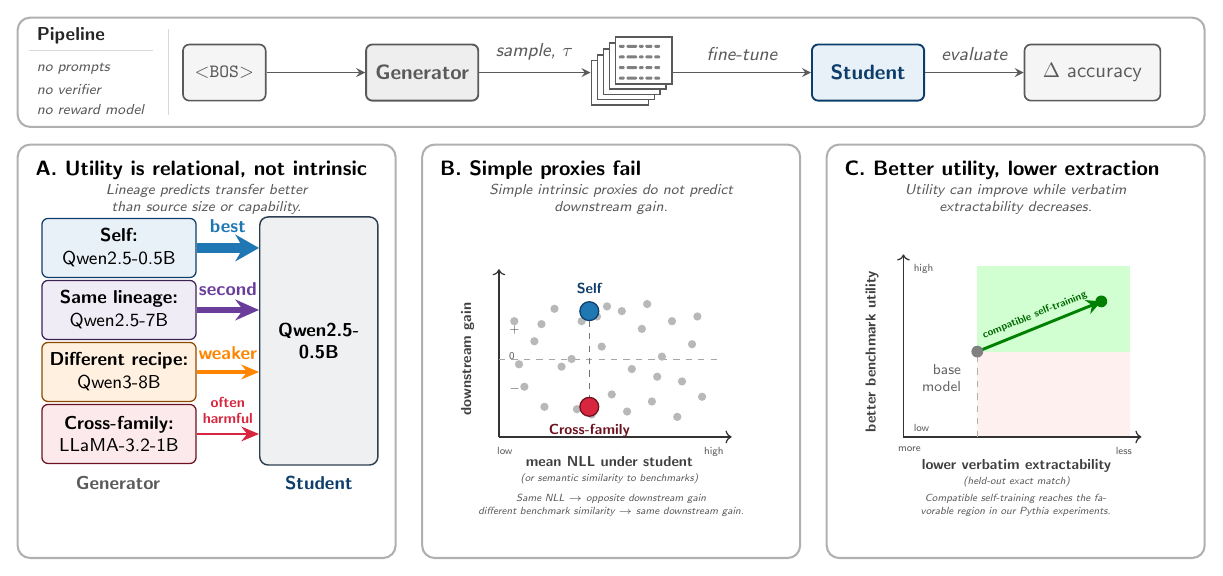}
  \caption{\textbf{Synthetic utility is relational, not intrinsic.} We study a prompt-free BOS-only pipeline in which a source model samples unconditional text and a student fine-tunes on it. \textbf{(A)} Transfer tracks student--source compatibility: self-generated data is strongest, same-lineage transfer is next, and cross-family transfer is weakest. \textbf{(B)} Intrinsic corpus proxies, including benchmark proximity and mean likelihood under the student, do not predict downstream utility. \textbf{(C)} In favorable self-training regimes, benchmark utility can improve while held-out exact-match verbatim extraction decreases.}
  \label{fig:overview}
\end{figure}

\begin{center}
\fbox{
\begin{minipage}{0.94\linewidth}
\textbf{Latent Capability Resurfacing Hypothesis.}
We use \textit{latent capability resurfacing} to denote three linked claims:
(i) pretrained language models contain useful capabilities that are not fully expressed by their base behavior;
(ii) weak synthetic fine-tuning can surface these capabilities when the synthetic corpus is compatible with the student, where compatibility is a relational property of the source-student pair rather than an intrinsic data property; and (iii) in favorable regimes, the update amplifies distributed task structure rather than sequence-specific recall, as evidenced by improved benchmark utility together with reduced verbatim extractability.
\end{minipage}
}
\end{center}

We test this hypothesis through three empirical findings. 
\underline{First}, synthetic utility is relational: self-generated data is strongest, a smaller same-lineage source (Qwen2.5-7B) transfers better than a larger but differently trained source (Qwen3-8B), and different-family sources transfer more weakly. Common intrinsic proxies fail to explain this ranking: neither benchmark similarity nor average per-token likelihood under the student predicts which corpora help, and scaling synthetic support yields little additional stable benefit. Within compatible pairings, temperature modulates a coverage--fidelity tradeoff. \underline{Second}, recoverable gains are strongly constrained by pretraining. Qwen shows broad gains across reasoning, math, and code, whereas the LLaMA student we study shows narrower gains concentrated on comprehension-style benchmarks and little reliable improvement on structured generation. In our setting, self-training amplifies latently supported behavior rather than reliably creating missing structured abilities. \underline{Third}, favorable self-training regimes improve the utility--extractability frontier: in controlled Pythia experiments, held-out exact-match verbatim extraction drops sharply while downstream capability is preserved or improved. This argues against simple rehearsal as the dominant explanation and points to reinforcement of less sequence-specific, more distributed structure. Our contributions can be outlined as follows: 
\begin{itemize}
    \item We formulate and test the \emph{latent capability resurfacing hypothesis} in a strictly minimal BOS-only regime: prompt-free unconditional self-training with no task prompts, verifiers, reward models, or external teachers.
    \item We show that synthetic utility is \emph{relational}: self-generated data is strongest, same-lineage transfer beats stronger but mismatched sources, and cross-family transfer is substantially weaker.
    \item We show that common intrinsic proxies do not explain the observed ranking: neither benchmark-level semantic similarity nor average per-token likelihood under the student predicts downstream gains, and increasing synthetic support yields little additional stable benefit once the source distribution is fixed.
    \item We identify a favorable self-training regime that improves the utility-extractability frontier: benchmark utility is preserved or improved while held-out exact-match extraction drops sharply, and true-continuation log-probability decreases under teacher forcing.
\end{itemize}

\section{Related Work}

Modern post-training pipelines treat synthetic data as useful but unsafe unless filtered through verifiers, reward models, or correctness checks~\citep{feng2025beyond, zelikman2022star, singh2024beyond}, and recursive self-consumption can produce distributional drift and ``model collapse''~\citep{shumailov2023curse, alemohammad2024self}. These results mainly study long-horizon or replacement-style regimes; our setting is short-horizon, low-learning-rate continued fine-tuning of an already-pretrained base model.

Recent work shows that unverified self-generated signals can also be beneficial. The closest prior result is \citet{zhang2026selfdistill}, who fine-tune on raw unverified solutions to competitive-programming prompts and improve code generation without a verifier or stronger teacher. Related evidence appears in reward-free self-training for reasoning~\citep{li2025selftrain} and training with unverified self-generated samples in vision~\citep{alemohammad2024selfimproving, alemohammad2025neon}.

A parallel line argues these gains come from capabilities already latent in the base model. \citet{shao2025spurious} show that random, incorrect, or format-only RLVR rewards still produce large gains on Qwen2.5-Math but fail on LLaMA and OLMo, mirroring the Qwen-versus-LLaMA split we observe. \citet{karan2025reasoning} reach a similar conclusion from inference alone, showing that training-free iterative sampling nearly matches RL-post-trained models, and \citet{ji2026scalable} obtain related gains through inference-time distribution sharpening. Most directly related, \citet{tan2026prerl} show that RL on the unconditional marginal $p(y)$ can serve as a surrogate for prompt-conditioned RL, concluding that RLVR is bounded by the base model's existing distribution.

These works collectively suggest that weak self-training amplifies structure already present in the pretrained model, but each relies on some form of external structure: task prompts, reward signals, question sampling, or inference-time orchestration. We study the strictly minimal case in which all of these are absent, isolating whether synthetic utility is an intrinsic property of the data or a relational property of the student-source pairing.

\section{Self-Generated Data Produces Model-Dependent Transient Gains}
\label{sec:transient}

We begin by asking whether plain unconditional self-generated text can produce transient capability gains in the strictly minimal BOS-only setting. All experiments use base language models immediately after pretraining. Synthetic corpora are generated unconditionally from the BOS token with no prompts, instructions, or task specifications. Each sample runs until an end-of-generation token or 1{,}024 tokens (minimum 50 tokens). We generate corpora at three sampling temperatures \(\tau \in \{0.75, 1.0, 1.25\}\) with no top-\(k\) or top-\(p\) truncation. Each corpus contains approximately 5M tokens. A generic replay baseline is constructed by sampling the publicly available multilingual Common Corpus~\citep{langlais2026common} to the same token budget.

All synthetic corpora undergo 8-gram decontamination against the test splits of every evaluation benchmark. From each decontaminated corpus we draw three independent, non-overlapping subsets of 5M tokens each (stratified across per-sample NLL bins) to estimate sensitivity to corpus subset selection under matched statistics. Fine-tuning uses full-parameter AdamW with a cosine schedule (10\% warmup,
decaying to zero), learning rate $1 \times 10^{-6}$, and effective batch
size 64, for 40 epochs, with bf16 mixed precision and sequence packing at
cutoff length 2048. Implementation uses LLaMA-Factory~\citep{zheng2024llamafactory}
with DeepSpeed ZeRO-3~\citep{9355301}; remaining optimizer hyperparameters
follow framework defaults. Only the training corpus varies across conditions.

We evaluate on two complementary suites. Suite A uses zero-shot evaluation on MMLU~\citep{hendrycks2021mmlu}, ARC-Challenge~\citep{clark2018arc}, GSM8K~\citep{cobbe2021gsm8k}, HellaSwag~\citep{zellers2019hellaswag}, TruthfulQA~\citep{lin2022truthfulqa}, Minerva-MATH~\citep{lewkowycz2022minerva}, and HumanEval@1~\citep{chen2021humaneval}. Suite B applies standard few-shot protocols (ARC-Challenge 25-shot, GSM8K 8-shot, HellaSwag 5-shot, Minerva-MATH 4-shot); the Suite B results mirror Suite A qualitatively and are reported in Appendix~\ref{app:fewshot}.

\input{figure_qwen_qwen.tex}

Figure~\ref{fig:qwen05b_cc_vs_syn} shows that self-generated text produces clear gains on Qwen2.5-0.5B across structured reasoning, math, and code, with task-dependent dynamics. On ARC-Challenge and HellaSwag, synthetic data produces steady positive deltas that saturate within 20 to 30 epochs and persist through training. On GSM8K, the \(\tau=1.25\) corpus produces a pronounced early spike before decaying toward baseline, the transient-then-degrade pattern characteristic of a weak-signal regime, while \(\tau=0.75\) and \(\tau=1.0\) produce smaller, more stable gains. On HumanEval, all three synthetic temperatures yield comparable modest improvements, with no clear temperature ordering. The generic replay baseline (Common Corpus) fails to reproduce these structured-reasoning gains: it is flat or negative on GSM8K, flat on Minerva-MATH, and steadily degrades on HumanEval. The Common Corpus is comparable to synthetic data only on MMLU and TruthfulQA, which is itself informative: generic pretraining text replays what was broadly learned, while self-generated text concentrates on the model's own high-probability modes. The signal in self-generated data is therefore specific to structured-reasoning and code capabilities, not a generic low-learning-rate regularization effect.

\input{figure_llama_llama.tex}

The effect is model-dependent. Figure~\ref{fig:llama_cc_vs_syn} shows that the identical protocol on LLaMA-3.2-1B does not reproduce the Qwen pattern. GSM8K and Minerva-MATH show no improvement under any condition; HumanEval degrades substantially under both Common Corpus and synthetic conditions. Only comprehension-style benchmarks (MMLU, ARC-Challenge, HellaSwag, TruthfulQA at \(\tau=1.25\)) show modest positive deltas, with the Common Corpus baseline often matching or exceeding synthetic data. The task-type split is qualitative: LLaMA gains selectively on multiple-choice comprehension and fails to gain (or degrades) on structured generation. The set of capabilities amenable to resurfacing appears to be fixed by the student's pretraining, a point we revisit in Section~\ref{sec:relational}.

\paragraph{Ruling out benchmark contamination.} To address the concern that transient gains could reflect accidental rehearsal of benchmark-like content, we conduct a semantic contamination audit in the style of~\citep{spiesberger2026soft}. We embed every synthetic sample and every benchmark test item using \texttt{llama-embed-nemotron-8b}~\citep{babakhin2025llamaembed}, and compute the maximum cosine similarity of each synthetic sample to any benchmark item.

\input{figure_hist.tex}

Figure~\ref{fig:hist_acc} reports the resulting max-similarity distributions. At \(\tau \leq 1.0\) the distributions are tight and centered well below 0.35; at \(\tau=1.25\) they shift modestly right, with a small but non-negligible mass above 0.35 on reasoning benchmarks. We select the 0.35 threshold as a conservative boundary: cosine values below 0.35 fall in the range that human inspection and the audit of~\citep{spiesberger2026soft} identify as unrelated, and subjectively distinct content frequently embeds at or above this value due to the geometric properties of the embedding space. To test whether the right-tail mass at \(\tau=1.25\) is responsible for the GSM8K gains, we partition the Qwen \(\tau=1.25\) corpus by per-sample max cosine similarity to GSM8K items (below 0.35, above 0.35) and fine-tune separate models on each subset under identical hyperparameters. The rightmost panel of Figure~\ref{fig:hist_acc} shows that the two learning curves are nearly indistinguishable: training on samples with \emph{no measurable} semantic relationship to GSM8K items produces the same gains as training on the nominally higher-similarity samples. This benchmark-proximity split does not explain the GSM8K gains.

Together, these results show that the gains are real and model-dependent, and that neither matched generic replay nor this benchmark-proximity proxy explains the observed ranking.

\section{Synthetic Utility Is Relational, Not Intrinsic}
\label{sec:relational}

Section~\ref{sec:transient} shows that prompt-free self-training can help, but only for some students and some tasks. We now ask what makes a synthetic corpus useful for a fixed student. Is utility an intrinsic property of the corpus, for example because the source model is larger, more capable, or more benchmark-like? Or is utility relational, depending on whether the source corpus is compatible with the student that consumes it? To test this, we fix the student to Qwen2.5-0.5B and train it on matched-token synthetic corpora generated by different source models. The sources include the student itself, Qwen2.5-7B, Qwen3-8B, and LLaMA-3.2-1B. Unless otherwise stated, corpora are generated at \(\tau=1.25\); for LLaMA we additionally sweep temperature to test whether the mismatch can be repaired by changing sampling diversity.

\input{figure_qwen_other.tex}

Figure~\ref{fig:qwen_other} shows a clear compatibility hierarchy. Self-generated data is strongest. Generations from Qwen2.5-7B, a larger same-lineage model, rank second. Generations from Qwen3-8B, which is larger and more capable but trained with a different recipe, transfer worse than Qwen2.5-7B despite standard capability metrics favoring the larger source. Generations from LLaMA-3.2-1B, a same-scale but different-family model, are weakest and often harm performance, especially on MMLU, TruthfulQA, Minerva-MATH, and HumanEval. The hierarchy is sharpest on structured-reasoning and code benchmarks, the same tasks where Qwen self-training produced the largest gains in Section~\ref{sec:transient}. On comprehension-style benchmarks the ordering is partially compressed, but the central pattern remains: source capability alone does not predict transfer. In this comparison, pretraining lineage predicts utility better than raw source strength.

One possible explanation is that Qwen-generated corpora are simply better in an absolute sense. The symmetric experiment in Appendix~\ref{app:llama_other} argues against this interpretation. A LLaMA-3.2-1B student trained on Qwen2.5-0.5B data at three temperatures fails to outperform LLaMA-self on the benchmarks where LLaMA gains at all, and underperforms on HellaSwag and ARC-Challenge. Thus Qwen data is not a generic upgrade. Its usefulness depends on the student, and the mirror pattern for LLaMA supports a relational view of synthetic utility.

A second intrinsic explanation is benchmark proximity. Section~\ref{sec:transient} tests this directly by partitioning Qwen \(\tau=1.25\) data according to maximum cosine similarity to GSM8K items. Training on low-similarity samples produces essentially the same GSM8K trajectory as training on higher-similarity samples, so this benchmark-proximity proxy does not explain the gains. A third natural explanation is likelihood under the student: perhaps useful corpora are simply those the student already assigns high probability to.

\input{figure_nll.tex}

Figure~\ref{fig:nll_dist} shows that mean student likelihood is also insufficient. We score every sample in the Qwen-own and LLaMA-cross corpora under the Qwen2.5-0.5B student. At lower temperatures, own-corpus samples receive lower NLL than cross-corpus samples, as expected. At \(\tau=1.25\), however, the two corpora have nearly identical mean per-token NLL under the same scorer (\(\mu=9.10\) for Qwen-own versus \(\mu=9.01\) for LLaMA-cross). Despite this likelihood match, their downstream effects differ sharply: Qwen-own improves structured reasoning, while LLaMA-cross leaves the student flat or worse. Mean likelihood under the student therefore does not predict utility.

This also clarifies the relationship to recent mechanistic accounts of RLVR. \citet{shao2025spurious} argue that GRPO can bias updates toward high-probability pretraining priors, suggesting that low-NLL modes may be important. Our prompt-free setting shows that likelihood cannot be the whole story. Within compatible pairings, higher-NLL samples at \(\tau=1.25\) can outperform lower-NLL samples at \(\tau=0.75\), and two corpora with nearly identical mean NLL can induce opposite training trajectories. The scalar likelihood of a corpus is therefore not enough; what matters is whether the corpus induces useful updates for the particular student.

Taken together, four simple intrinsic proxies do not explain the ranking we observe: source capability, student-independent corpus quality, benchmark-level semantic similarity, and mean likelihood under the student. A plausible remaining explanation is \emph{student-source compatibility}: different corpora induce different update directions under the same student, and those directions may align differently with useful latent structure already present in the model. Same-lineage sources appear more likely to produce data whose higher-order sequential structure is compatible with the student's learned representation, while distant sources can produce text that is fluent, high-likelihood, or high-capability in isolation but still induces unhelpful updates for that student. We do not directly measure gradient alignment here, so we treat compatibility as an empirically supported explanatory hypothesis rather than a demonstrated mechanism. Under this view, sampling temperature modulates which modes are emphasized within a compatible pairing, while compatibility determines whether the induced update is useful at all. This is the relational form of the latent capability resurfacing hypothesis: weak self-training helps when the synthetic signal reinforces structure already supported by the student, and a corpus's usefulness is inseparable from the student that consumes it.

\section{Self-Training Improves the Utility-Extractability Frontier}
\label{sec:memo}

If resurfacing amplifies distributed task structure rather than sequence-specific recall, then favorable self-training should not increase verbatim extraction, and may reduce it. We test this prediction on Pythia-1B and Pythia-6.9B, standard testbeds for memorization research because their pretraining corpus is documented~\citep{biderman2023pythia, carlini2022quantifying}. We apply our unconditional self-training protocol ($\tau=1.25$, 5M tokens, 40 epochs) across two learning rates ($\mathrm{lr}\in\{10^{-5}, 10^{-6}\}$). Pythia experiments use a separate training pipeline implemented in
HuggingFace Transformers; full hyperparameters are reported in
Appendix~\ref{app:reproducibility}.

Capability is tracked via ARC-Challenge, HellaSwag, ARC-Easy, and WinoGrande. Memorization is evaluated using the prefix-extraction attack of \citet{carlini2021extracting}: we prompt the model with 50-token prefixes from 10{,}000 fixed sequences
from each of Wikipedia, Enron, Pile-CC, and GitHub (40{,}000 total)

\input{fig_pythia_bench}
\input{fig_memo.tex}

At $\mathrm{lr}=10^{-5}$, verbatim extraction drops sharply on both models (Figure~\ref{fig:memo}, left two panels). On text, Pythia-1B falls from $\sim$184 memorized sequences to near zero ($\sim$97\% reduction), and Pythia-6.9B falls from $\sim$540 to $\sim$34 ($\sim$94\% reduction). On code, both models drop by roughly half (Pythia-1B: $\sim$1900 $\to$ $\sim$950; Pythia-6.9B: $\sim$2660 $\to$ $\sim$1390). At $\mathrm{lr}=10^{-6}$, Pythia-1B shows moderate reductions while Pythia-6.9B remains essentially unchanged.

Capability changes track the same axis (Figure~\ref{fig:capability}). Across every condition where either quantity shifts appreciably, capability and memorization move in opposite directions. Pythia-6.9B at $\mathrm{lr}=10^{-5}$ shows the cleanest co-occurrence: ARC-Challenge improves by $\sim$+2.5 and HellaSwag by $\sim$+1.5 while memorization crashes, with identical patterns on ARC-Easy and WinoGrande. Pythia-1B at $\mathrm{lr}=10^{-6}$ shows modest capability gains alongside moderate memorization reduction. The remaining two conditions bound the effect: at Pythia-6.9B with $\mathrm{lr}=10^{-6}$ both quantities barely move; at Pythia-1B with $\mathrm{lr}=10^{-5}$ the transient-then-degrade pattern from Section~\ref{sec:transient} reappears, yet memorization still drops steadily toward zero.

The extraction attack measures recoverability under greedy decoding. To confirm the effect is representational rather than a decoding artifact, we also compute the average log-probability assigned to the true continuation of the same held-out sequences under teacher-forcing: $\Delta \log p(x \mid y) = \log p_{\theta_t}(x \mid y) - \log p_{\theta_{\mathrm{base}}}(x \mid y)$. Figure~\ref{fig:memo} (right two panels) shows these log-probabilities drop under exactly the same conditions where extraction drops. Pythia-1B and Pythia-6.9B at $\mathrm{lr}=10^{-5}$ both exhibit steady negative $\Delta \log p$, while the sub-threshold condition ($\mathrm{lr}=10^{-6}$ on Pythia-6.9B) remains near zero. The model genuinely moves probability mass away from pretraining sequences.

A cohesive picture emerges across the four conditions: self-training shifts the model away from the pretraining solution into a neighboring latent mode. The learning rate controls travel distance; model size determines the quality of the destination. Pythia-6.9B at $\mathrm{lr}=10^{-5}$ moves decisively into a new mode that is structurally distinct from the pretraining basin, simultaneously producing capability gains and sharp memorization reductions.

Crucially, these observations rule out the hypothesis that capability gains arise from rehearsal of memorized training data. If rehearsal were the mechanism, extraction rates would rise alongside capability. Instead, capability and memorization move in opposite directions across every condition where either shifts appreciably. The log-probability measurements confirm this is a genuine redistribution of probability mass, not a decoding artifact. The modes reinforced by self-training are therefore distributed and generalizable, firmly supporting the resurfacing interpretation from Sections~\ref{sec:transient} and~\ref{sec:relational}.

While prior work shows that targeted unlearning often degrades utility, our BOS-only regime isolates a distinct mechanism: untargeted self-training organically decouples capability from verbatim recall. \citet{carlini2021extracting} and \citet{carlini2022quantifying} demonstrate that LLMs memorize and regurgitate verbatim training data, with extraction rates scaling with model size and duplication. Existing unlearning methods reinforce this entanglement view: gradient ascent~\citep{jang2022knowledge} or preference optimization on targeted forget sets~\citep{maini2024tofu} require explicitly specifying content to remove and typically degrade utility. Recently, \citet{scholten2026pmc} showed that deliberately triggering model collapse on specific forget queries can remove targeted information, preserving utility at best. Our regime operates entirely differently. No forget set is specified, no privacy objective is optimized, no reward is used, and no utility is sacrificed. The memorization reduction is entirely emergent---a byproduct of reinforcing distributed modes that compete with sequence-specific recall for the same representational capacity. To our knowledge, this is the first regime in which plain untargeted self-training jointly improves capability and reduces memorization without any specified forget set or privacy objective. Where prior work finds capability and memorization hopelessly entangled, we find a regime where they cleanly decouple.

\section{Conclusion}
\label{sec:conclusion}

We have studied prompt-free self-training in the minimal BOS-only regime, where base language models are continued on unconditional synthetic text without prompts, verifiers, reward models, or external teachers. In this setting, synthetic utility is relational: self-generated and close-lineage corpora are most effective, while source capability, benchmark proximity, and mean student likelihood do not explain which corpora help. The gains are selective and strongly student-dependent, suggesting that weak self-training amplifies structure already supported by the base model rather than reliably creating missing capabilities. The same regime also reveals a surprising utility--extractability effect. In controlled Pythia experiments, favorable self-training runs preserve or improve benchmark utility while sharply reducing held-out exact-match extraction and lowering the probability assigned to true pretraining continuations. These results support the {\bf latent capability resurfacing hypothesis}: compatible synthetic data can reinforce distributed task structure while moving probability mass away from sequence-specific recall. More broadly, our findings suggest that synthetic post-training should be designed around student--source compatibility, not synthetic data quality alone.

\paragraph{Limitations.}
Our study isolates prompt-free self-training in a deliberately minimal BOS-only regime, but several limitations remain. First, our main relational-utility experiments focus on a small set of open-weight model families and scales, so the compatibility hierarchy we observe should not be interpreted as a universal ordering across all pretrained models, such as those at excessive scales. Second, although our results support student-source compatibility as an explanatory hypothesis, we do not directly measure gradient alignment or the geometry of the induced update directions; doing so is an important direction for future work. Third, our memorization analysis measures held-out exact-match extraction and true-continuation likelihood on documented Pythia pretraining sequences, which are stronger than greedy extraction alone but still do not constitute a complete privacy audit. Finally, our downstream evaluations rely primarily on standard static benchmarks, so future work should test whether the same resurfacing and utility-extractability patterns hold under dynamic, contamination-resistant, and deployment-oriented evaluations.

\section*{Acknowledgments}

This work was supported by NSF Awards 2145346 (CAREER), 02133861 (DMS), and 2113904 (CCSS), the NSF AI Institute for Foundations of Machine Learning (IFML), ONR grant N00014-23-1-2714, DOE grant DE-SC0020345, and DOI grant 140D0423C0076. This work was also supported by computing resources on the Vista GPU Cluster through the Center for Generative AI (CGAI) and the Texas Advanced Computing Center (TACC) at The University of Texas at Austin.

\bibliography{reference}
\bibliographystyle{plainnat}

\newpage
\appendix
\section{Few-Shot Evaluation Results}
\label{app:fewshot}

The main text reports zero-shot evaluation throughout. To verify that the observed patterns are not artifacts of the zero-shot protocol, we repeat the Qwen2.5-0.5B self-generated versus real-data comparison from Section~\ref{sec:transient} under standard few-shot protocols: ARC-Challenge 25-shot, GSM8K 8-shot, HellaSwag 5-shot, and Minerva-MATH 4-shot. All other experimental details (synthetic corpora, temperatures, token budget, learning rate, optimizer, and subset structure) are identical to the zero-shot experiments.

Figure~\ref{fig:qwen_fewshot} shows that the few-shot results mirror the zero-shot pattern qualitatively. On ARC-Challenge and HellaSwag, self-generated data at $\tau{=}1.25$ produces the largest sustained gains, while the Common Corpus baseline is competitive but weaker. On GSM8K, synthetic data at $\tau{=}0.75$ and $\tau{=}1.0$ produces early transient gains that decay over continued training, reproducing the transient-then-degrade pattern observed in the zero-shot setting. On Minerva-MATH, synthetic conditions show modest early improvement followed by decay, while the Common Corpus baseline trends negative. The task-type split, structured-reasoning gains from self-generated data that generic replay cannot reproduce, is consistent across evaluation protocols.

\input{figure_qwen_fewshot.tex}

\section{Symmetric Cross-Model Experiment: LLaMA Student with Qwen Teacher}
\label{app:llama_other}

Section~\ref{sec:relational} establishes that Qwen-generated data transfers poorly to the Qwen2.5-0.5B student when the source model is from a different pretraining lineage. One might worry that this asymmetry reflects an intrinsic quality difference between Qwen and LLaMA corpora rather than a compatibility effect. To rule this out, we run the symmetric experiment: a LLaMA-3.2-1B student trained on Qwen2.5-0.5B teacher data at three temperatures ($\tau \in \{0.75, 1.0, 1.25\}$), compared against LLaMA-self at $\tau{=}1.25$.

Figure~\ref{fig:llama_qwen} shows that Qwen teacher data fails to outperform LLaMA-self on the benchmarks where LLaMA gains at all, and actively underperforms on HellaSwag and ARC-Challenge. On GSM8K, Minerva-MATH, and HumanEval, neither LLaMA-self nor Qwen-teacher data produces meaningful gains, consistent with the main-text finding that LLaMA's pretraining does not latently support structured generation in the way Qwen's does. Qwen data is therefore not a generic upgrade; its utility is specifically tied to compatibility with a Qwen-lineage student.

Figure~\ref{fig:nll_dist_llama} reports the per-sample NLL distributions under the LLaMA-3.2-1B scorer, mirroring the Qwen-scorer analysis in Figure~\ref{fig:nll_dist}. The pattern is the same: at $\tau{=}1.25$, the own-corpus and cross-corpus distributions converge to nearly identical mean NLL ($\mu{=}8.88$ vs $\mu{=}8.93$), yet produce different downstream effects. This confirms that the failure of per-token likelihood to predict utility is symmetric across model families and is not an artifact of the Qwen scoring direction.

\input{figure_llama_qwen.tex}
\input{figure_nll_llama.tex}

\section{Reproducibility Details}
\label{app:reproducibility}

\paragraph{Qwen and LLaMA experiments.}
All experiments were run on NVIDIA A100-SXM4 80GB GPU.
Training uses LLaMA-Factory~\citep{zheng2024llamafactory} with DeepSpeed
ZeRO-3~\citep{9355301} in bf16 mixed precision, full-parameter AdamW
with framework-default optimizer hyperparameters, a cosine schedule
with 10\% warmup decaying to zero, learning rate $1 \times 10^{-6}$,
effective batch size 64, 40 epochs, sequence packing, and cutoff length
2048. Generation uses vLLM~\citep{10.1145/3600006.3613165} in bf16 with
\texttt{top\_p}~$=1.0$, \texttt{top\_k}~$=-1$, \texttt{min\_tokens}~$=50$,
and \texttt{max\_tokens}~$=1024$; we override model-default stop tokens
because vLLM's \texttt{min\_tokens} does not block them, which would
otherwise truncate samples below the 50-token minimum. Evaluation uses
lm-evaluation-harness~\citep{eval-harness} with the vLLM backend, greedy
decoding, \texttt{seed}~$=42$, and deterministic CUDA settings. Each Qwen2.5-0.5B or LLaMA-3.2-1B fine-tuning run required approximately 4 GPU-hours.

\paragraph{Pythia experiments.}
Training uses HuggingFace Transformers in fp32 with AdamW
($\beta_1 = 0.9$, $\beta_2 = 0.95$, weight decay $0$, gradient clipping
$1.0$) and a linear warmup-then-decay schedule (50 warmup steps), with
batch size 4 $\times$ gradient accumulation 8 (effective batch size 32),
max sequence length 1024, learning rates $\{10^{-5}, 10^{-6}\}$, and
40 epochs. Generation uses HuggingFace Transformers in fp16 for GPT-NeoX
architecture compatibility, with the same BOS-only protocol,
$\tau = 1.25$, \texttt{min\_tokens}~$=50$, and \texttt{max\_tokens}~$=1024$
as in the main protocol. Pythia models are loaded from the \texttt{main}
revision of \texttt{EleutherAI/pythia-1b} and
\texttt{EleutherAI/pythia-6.9b}. Verbatim extraction and exposure
log-probability evaluation use HuggingFace Transformers in fp16 with
greedy decoding for extraction and single-pass teacher forcing for
exposure. Each Pythia-1B fine-tuning run required approximately 8 GPU-hours. Each Pythia-6.9B fine-tuning run required approximately 24 GPU-hours.

\end{document}

%% file: math_commands.tex
\usepackage{amsmath,amssymb,amsfonts,bm}
\usepackage{amsthm}

\def\eqref#1{equation~\ref{#1}}

\def\1{\bm{1}}

\DeclareMathAlphabet{\mathsfit}{\encodingdefault}{\sfdefault}{m}{sl}
\SetMathAlphabet{\mathsfit}{bold}{\encodingdefault}{\sfdefault}{bx}{n}

\newtheorem*{assumption*}{Assumption}
\theoremstyle{definition}

\theoremstyle{remark}

\makeatletter
\let\save@mathaccent\mathaccent
\newcommand*\if@single[3]{%
  \setbox0\hbox{${\mathaccent"0362{#1}}^H$}%
  \setbox2\hbox{${\mathaccent"0362{\kern0pt#1}}^H$}%
  \ifdim\ht0=\ht2 #3\else #2\fi
}
\newcommand*\rel@kern[1]{\kern#1\dimexpr\macc@kerna}
\newcommand*\widebar[1]{\@ifnextchar^{{\wide@bar{#1}{0}}}{\wide@bar{#1}{1}}}
\newcommand*\wide@bar[2]{\if@single{#1}{\wide@bar@{#1}{#2}{1}}{\wide@bar@{#1}{#2}{2}}}
\newcommand*\wide@bar@[3]{%
  \begingroup
  \def\mathaccent##1##2{%
    \let\mathaccent\save@mathaccent
    \if#32 \let\macc@nucleus\first@char \fi
    \setbox\z@\hbox{$\macc@style{\macc@nucleus}_{}$}%
    \setbox\tw@\hbox{$\macc@style{\macc@nucleus}{}_{}$}%
    \dimen@\wd\tw@
    \advance\dimen@-\wd\z@
    \divide\dimen@ 3
    \@tempdima\wd\tw@
    \advance\@tempdima-\scriptspace
    \divide\@tempdima 10
    \advance\dimen@-\@tempdima
    \ifdim\dimen@>\z@ \dimen@0pt\fi
    \rel@kern{0.6}\kern-\dimen@
    \if#31
      \overline{\rel@kern{-0.6}\kern\dimen@\macc@nucleus\rel@kern{0.4}\kern\dimen@}%
      \advance\dimen@0.4\dimexpr\macc@kerna
      \let\final@kern#2%
      \ifdim\dimen@<\z@ \let\final@kern1\fi
      \if\final@kern1 \kern-\dimen@\fi
    \else
      \overline{\rel@kern{-0.6}\kern\dimen@#1}%
    \fi
  }%
  \macc@depth\@ne
  \let\math@bgroup\@empty \let\math@egroup\macc@set@skewchar
  \mathsurround\z@ \frozen@everymath{\mathgroup\macc@group\relax}%
  \macc@set@skewchar\relax
  \let\mathaccentV\macc@nested@a
  \if#31
    \macc@nested@a\relax111{#1}%
  \else
    \def\gobble@till@marker##1\endmarker{}%
    \futurelet\first@char\gobble@till@marker#1\endmarker
    \ifcat\noexpand\first@char A\else \def\first@char{}\fi
    \macc@nested@a\relax111{\first@char}%
  \fi
  \endgroup
}
\makeatother

\definecolor{editone}{HTML}{0000FF}
\definecolor{edittwo}{HTML}{FF0099}
\definecolor{editthree}{HTML}{FF6600}

\definecolor{DarkGreen}{RGB}{158, 6, 151}
\def\hideNotes{0}
\if\hideNotes0
  
  \newcommand\richb[1]{{\color{red}\sf{[richb: #1]}}}
  
\else
  \renewcommand{\tableofcontents}{}
  \newcommand\lolo[1]{}
  \newcommand\richb[1]{}
  \newcommand\josue[1]{}
\fi

\newcommand\blfootnote[1]{%
  \begingroup
  \renewcommand\thefootnote{}%
  \footnote{#1}%
  \addtocounter{footnote}{-1}%
  \endgroup
}


%% file: figure_qwen_qwen.tex
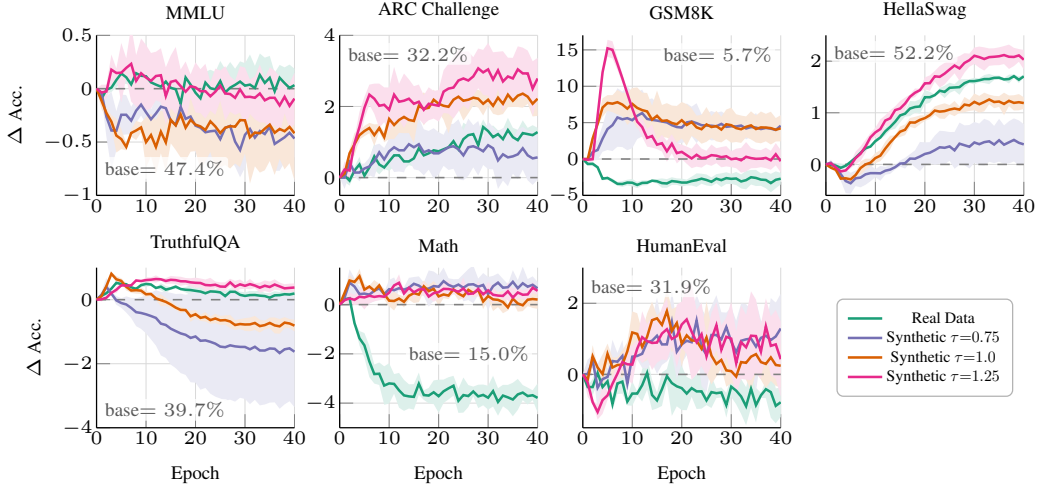
\begin{figure}[t]
\vspace{-0.5em}
\centering
\resizebox{\textwidth}{!}{%
\begin{tikzpicture}
\usepgfplotslibrary{fillbetween}
\begin{groupplot}[
  group style={group size=4 by 2, horizontal sep=5mm, vertical sep=8mm},
  width=0.27\linewidth, height=0.24\linewidth,
  xtick={0,10,20,30,40},
  tick label style={font=\fontsize{4}{4}\selectfont},
  label style={font=\tiny},
  title style={font=\tiny, yshift=-1ex},
]

  \nextgroupplot[
    title={\tiny MMLU},
    ylabel={\tiny $\Delta$ Acc.},
    xlabel={},
    axis x line*=bottom, axis y line*=left,
    xmin=0, xmax=40, ymin=-1.0, ymax=0.5,
    grid, grid style={line width=.2pt, draw=gray!30},
  ]
  \draw[gray,dashed,line width=0.5pt](axis cs:0,0)--(axis cs:40,0);
  \node[anchor=south west, font=\tiny, text=gray!70!black,
    fill=white, fill opacity=0.75, text opacity=1,
    inner sep=1pt, rounded corners=1pt]
    at (axis cs:1,-0.85) {base$=47.4\%$};
  \addplot[name path=ummluA, draw=none, forget plot]
    table[x=epoch, y expr=\thisrow{cfg0d}+\thisrow{cfg0s}, col sep=comma]{csv_tikz/qwen_qwen/mmlu.csv};
  \addplot[name path=lmmluA, draw=none, forget plot]
    table[x=epoch, y expr=\thisrow{cfg0d}-\thisrow{cfg0s}, col sep=comma]{csv_tikz/qwen_qwen/mmlu.csv};
  \addplot[colorA!15, draw=none, forget plot] fill between[of=ummluA and lmmluA];
  \addplot[colorA, line width=0.8pt]
    table[x=epoch, y=cfg0d, col sep=comma]{csv_tikz/qwen_qwen/mmlu.csv};
  \addplot[name path=ummluC, draw=none, forget plot]
    table[x=epoch, y expr=\thisrow{cfg1d}+\thisrow{cfg1s}, col sep=comma]{csv_tikz/qwen_qwen/mmlu.csv};
  \addplot[name path=lmmluC, draw=none, forget plot]
    table[x=epoch, y expr=\thisrow{cfg1d}-\thisrow{cfg1s}, col sep=comma]{csv_tikz/qwen_qwen/mmlu.csv};
  \addplot[colorC!15, draw=none, forget plot] fill between[of=ummluC and lmmluC];
  \addplot[colorC, line width=0.8pt]
    table[x=epoch, y=cfg1d, col sep=comma]{csv_tikz/qwen_qwen/mmlu.csv};
  \addplot[name path=ummluB, draw=none, forget plot]
    table[x=epoch, y expr=\thisrow{cfg2d}+\thisrow{cfg2s}, col sep=comma]{csv_tikz/qwen_qwen/mmlu.csv};
  \addplot[name path=lmmluB, draw=none, forget plot]
    table[x=epoch, y expr=\thisrow{cfg2d}-\thisrow{cfg2s}, col sep=comma]{csv_tikz/qwen_qwen/mmlu.csv};
  \addplot[colorB!15, draw=none, forget plot] fill between[of=ummluB and lmmluB];
  \addplot[colorB, line width=0.8pt]
    table[x=epoch, y=cfg2d, col sep=comma]{csv_tikz/qwen_qwen/mmlu.csv};
  \addplot[name path=ummluD, draw=none, forget plot]
    table[x=epoch, y expr=\thisrow{cfg3d}+\thisrow{cfg3s}, col sep=comma]{csv_tikz/qwen_qwen/mmlu.csv};
  \addplot[name path=lmmluD, draw=none, forget plot]
    table[x=epoch, y expr=\thisrow{cfg3d}-\thisrow{cfg3s}, col sep=comma]{csv_tikz/qwen_qwen/mmlu.csv};
  \addplot[colorD!15, draw=none, forget plot] fill between[of=ummluD and lmmluD];
  \addplot[colorD, line width=0.8pt]
    table[x=epoch, y=cfg3d, col sep=comma]{csv_tikz/qwen_qwen/mmlu.csv};

  \nextgroupplot[
    title={\tiny ARC Challenge},
    ylabel={},
    xlabel={},
    axis x line*=bottom, axis y line*=left,
    xmin=0, xmax=40, ymin=-0.5, ymax=4.0,
    grid, grid style={line width=.2pt, draw=gray!30},
  ]
  \draw[gray,dashed,line width=0.5pt](axis cs:0,0)--(axis cs:40,0);
  \node[anchor=north west, font=\tiny, text=gray!70!black,
    fill=white, fill opacity=0.75, text opacity=1,
    inner sep=1pt, rounded corners=1pt]
    at (axis cs:1,3.8) {base$=32.2\%$};
  \addplot[name path=uarcchA, draw=none, forget plot]
    table[x=epoch, y expr=\thisrow{cfg0d}+\thisrow{cfg0s}, col sep=comma]{csv_tikz/qwen_qwen/arc_challenge.csv};
  \addplot[name path=larcchA, draw=none, forget plot]
    table[x=epoch, y expr=\thisrow{cfg0d}-\thisrow{cfg0s}, col sep=comma]{csv_tikz/qwen_qwen/arc_challenge.csv};
  \addplot[colorA!15, draw=none, forget plot] fill between[of=uarcchA and larcchA];
  \addplot[colorA, line width=0.8pt]
    table[x=epoch, y=cfg0d, col sep=comma]{csv_tikz/qwen_qwen/arc_challenge.csv};
  \addplot[name path=uarcchC, draw=none, forget plot]
    table[x=epoch, y expr=\thisrow{cfg1d}+\thisrow{cfg1s}, col sep=comma]{csv_tikz/qwen_qwen/arc_challenge.csv};
  \addplot[name path=larcchC, draw=none, forget plot]
    table[x=epoch, y expr=\thisrow{cfg1d}-\thisrow{cfg1s}, col sep=comma]{csv_tikz/qwen_qwen/arc_challenge.csv};
  \addplot[colorC!15, draw=none, forget plot] fill between[of=uarcchC and larcchC];
  \addplot[colorC, line width=0.8pt]
    table[x=epoch, y=cfg1d, col sep=comma]{csv_tikz/qwen_qwen/arc_challenge.csv};
  \addplot[name path=uarcchB, draw=none, forget plot]
    table[x=epoch, y expr=\thisrow{cfg2d}+\thisrow{cfg2s}, col sep=comma]{csv_tikz/qwen_qwen/arc_challenge.csv};
  \addplot[name path=larcchB, draw=none, forget plot]
    table[x=epoch, y expr=\thisrow{cfg2d}-\thisrow{cfg2s}, col sep=comma]{csv_tikz/qwen_qwen/arc_challenge.csv};
  \addplot[colorB!15, draw=none, forget plot] fill between[of=uarcchB and larcchB];
  \addplot[colorB, line width=0.8pt]
    table[x=epoch, y=cfg2d, col sep=comma]{csv_tikz/qwen_qwen/arc_challenge.csv};
  \addplot[name path=uarcchD, draw=none, forget plot]
    table[x=epoch, y expr=\thisrow{cfg3d}+\thisrow{cfg3s}, col sep=comma]{csv_tikz/qwen_qwen/arc_challenge.csv};
  \addplot[name path=larcchD, draw=none, forget plot]
    table[x=epoch, y expr=\thisrow{cfg3d}-\thisrow{cfg3s}, col sep=comma]{csv_tikz/qwen_qwen/arc_challenge.csv};
  \addplot[colorD!15, draw=none, forget plot] fill between[of=uarcchD and larcchD];
  \addplot[colorD, line width=0.8pt]
    table[x=epoch, y=cfg3d, col sep=comma]{csv_tikz/qwen_qwen/arc_challenge.csv};

  \nextgroupplot[
    title={\tiny GSM8K},
    ylabel={},
    xlabel={},
    axis x line*=bottom, axis y line*=left,
    xmin=0, xmax=40, ymin=-5.0, ymax=17.0,
    grid, grid style={line width=.2pt, draw=gray!30},
  ]
  \draw[gray,dashed,line width=0.5pt](axis cs:0,0)--(axis cs:40,0);
  \node[anchor=north east, font=\tiny, text=gray!70!black,
    fill=white, fill opacity=0.75, text opacity=1,
    inner sep=1pt, rounded corners=1pt]
    at (axis cs:39,16.0) {base$=5.7\%$};
  \addplot[name path=ugsm8kA, draw=none, forget plot]
    table[x=epoch, y expr=\thisrow{cfg0d}+\thisrow{cfg0s}, col sep=comma]{csv_tikz/qwen_qwen/gsm8k.csv};
  \addplot[name path=lgsm8kA, draw=none, forget plot]
    table[x=epoch, y expr=\thisrow{cfg0d}-\thisrow{cfg0s}, col sep=comma]{csv_tikz/qwen_qwen/gsm8k.csv};
  \addplot[colorA!15, draw=none, forget plot] fill between[of=ugsm8kA and lgsm8kA];
  \addplot[colorA, line width=0.8pt]
    table[x=epoch, y=cfg0d, col sep=comma]{csv_tikz/qwen_qwen/gsm8k.csv};
  \addplot[name path=ugsm8kC, draw=none, forget plot]
    table[x=epoch, y expr=\thisrow{cfg1d}+\thisrow{cfg1s}, col sep=comma]{csv_tikz/qwen_qwen/gsm8k.csv};
  \addplot[name path=lgsm8kC, draw=none, forget plot]
    table[x=epoch, y expr=\thisrow{cfg1d}-\thisrow{cfg1s}, col sep=comma]{csv_tikz/qwen_qwen/gsm8k.csv};
  \addplot[colorC!15, draw=none, forget plot] fill between[of=ugsm8kC and lgsm8kC];
  \addplot[colorC, line width=0.8pt]
    table[x=epoch, y=cfg1d, col sep=comma]{csv_tikz/qwen_qwen/gsm8k.csv};
  \addplot[name path=ugsm8kB, draw=none, forget plot]
    table[x=epoch, y expr=\thisrow{cfg2d}+\thisrow{cfg2s}, col sep=comma]{csv_tikz/qwen_qwen/gsm8k.csv};
  \addplot[name path=lgsm8kB, draw=none, forget plot]
    table[x=epoch, y expr=\thisrow{cfg2d}-\thisrow{cfg2s}, col sep=comma]{csv_tikz/qwen_qwen/gsm8k.csv};
  \addplot[colorB!15, draw=none, forget plot] fill between[of=ugsm8kB and lgsm8kB];
  \addplot[colorB, line width=0.8pt]
    table[x=epoch, y=cfg2d, col sep=comma]{csv_tikz/qwen_qwen/gsm8k.csv};
  \addplot[name path=ugsm8kD, draw=none, forget plot]
    table[x=epoch, y expr=\thisrow{cfg3d}+\thisrow{cfg3s}, col sep=comma]{csv_tikz/qwen_qwen/gsm8k.csv};
  \addplot[name path=lgsm8kD, draw=none, forget plot]
    table[x=epoch, y expr=\thisrow{cfg3d}-\thisrow{cfg3s}, col sep=comma]{csv_tikz/qwen_qwen/gsm8k.csv};
  \addplot[colorD!15, draw=none, forget plot] fill between[of=ugsm8kD and lgsm8kD];
  \addplot[colorD, line width=0.8pt]
    table[x=epoch, y=cfg3d, col sep=comma]{csv_tikz/qwen_qwen/gsm8k.csv};

  \nextgroupplot[
    title={\tiny HellaSwag},
    ylabel={},
    xlabel={},
    axis x line*=bottom, axis y line*=left,
    xmin=0, xmax=40, ymin=-0.6, ymax=2.5,
    grid, grid style={line width=.2pt, draw=gray!30},
  ]
  \draw[gray,dashed,line width=0.5pt](axis cs:0,0)--(axis cs:40,0);
  \node[anchor=north west, font=\tiny, text=gray!70!black,
    fill=white, fill opacity=0.75, text opacity=1,
    inner sep=1pt, rounded corners=1pt]
    at (axis cs:1,2.4) {base$=52.2\%$};
  \addplot[name path=uhellaA, draw=none, forget plot]
    table[x=epoch, y expr=\thisrow{cfg0d}+\thisrow{cfg0s}, col sep=comma]{csv_tikz/qwen_qwen/hellaswag.csv};
  \addplot[name path=lhellaA, draw=none, forget plot]
    table[x=epoch, y expr=\thisrow{cfg0d}-\thisrow{cfg0s}, col sep=comma]{csv_tikz/qwen_qwen/hellaswag.csv};
  \addplot[colorA!15, draw=none, forget plot] fill between[of=uhellaA and lhellaA];
  \addplot[colorA, line width=0.8pt]
    table[x=epoch, y=cfg0d, col sep=comma]{csv_tikz/qwen_qwen/hellaswag.csv};
  \addplot[name path=uhellaC, draw=none, forget plot]
    table[x=epoch, y expr=\thisrow{cfg1d}+\thisrow{cfg1s}, col sep=comma]{csv_tikz/qwen_qwen/hellaswag.csv};
  \addplot[name path=lhellaC, draw=none, forget plot]
    table[x=epoch, y expr=\thisrow{cfg1d}-\thisrow{cfg1s}, col sep=comma]{csv_tikz/qwen_qwen/hellaswag.csv};
  \addplot[colorC!15, draw=none, forget plot] fill between[of=uhellaC and lhellaC];
  \addplot[colorC, line width=0.8pt]
    table[x=epoch, y=cfg1d, col sep=comma]{csv_tikz/qwen_qwen/hellaswag.csv};
  \addplot[name path=uhellaB, draw=none, forget plot]
    table[x=epoch, y expr=\thisrow{cfg2d}+\thisrow{cfg2s}, col sep=comma]{csv_tikz/qwen_qwen/hellaswag.csv};
  \addplot[name path=lhellaB, draw=none, forget plot]
    table[x=epoch, y expr=\thisrow{cfg2d}-\thisrow{cfg2s}, col sep=comma]{csv_tikz/qwen_qwen/hellaswag.csv};
  \addplot[colorB!15, draw=none, forget plot] fill between[of=uhellaB and lhellaB];
  \addplot[colorB, line width=0.8pt]
    table[x=epoch, y=cfg2d, col sep=comma]{csv_tikz/qwen_qwen/hellaswag.csv};
  \addplot[name path=uhellaD, draw=none, forget plot]
    table[x=epoch, y expr=\thisrow{cfg3d}+\thisrow{cfg3s}, col sep=comma]{csv_tikz/qwen_qwen/hellaswag.csv};
  \addplot[name path=lhellaD, draw=none, forget plot]
    table[x=epoch, y expr=\thisrow{cfg3d}-\thisrow{cfg3s}, col sep=comma]{csv_tikz/qwen_qwen/hellaswag.csv};
  \addplot[colorD!15, draw=none, forget plot] fill between[of=uhellaD and lhellaD];
  \addplot[colorD, line width=0.8pt]
    table[x=epoch, y=cfg3d, col sep=comma]{csv_tikz/qwen_qwen/hellaswag.csv};

  \nextgroupplot[
    title={\tiny TruthfulQA},
    ylabel={\tiny $\Delta$ Acc.},
    xlabel={\tiny Epoch},
    axis x line*=bottom, axis y line*=left,
    xmin=0, xmax=40, ymin=-4.0, ymax=1.0,
    grid, grid style={line width=.2pt, draw=gray!30},
  ]
  \draw[gray,dashed,line width=0.5pt](axis cs:0,0)--(axis cs:40,0);
  \node[anchor=south west, font=\tiny, text=gray!70!black,
    fill=white, fill opacity=0.75, text opacity=1,
    inner sep=1pt, rounded corners=1pt]
    at (axis cs:1,-3.8) {base$=39.7\%$};
  \addplot[name path=utruthA, draw=none, forget plot]
    table[x=epoch, y expr=\thisrow{cfg0d}+\thisrow{cfg0s}, col sep=comma]{csv_tikz/qwen_qwen/truthfulqa_mc2.csv};
  \addplot[name path=ltruthA, draw=none, forget plot]
    table[x=epoch, y expr=\thisrow{cfg0d}-\thisrow{cfg0s}, col sep=comma]{csv_tikz/qwen_qwen/truthfulqa_mc2.csv};
  \addplot[colorA!15, draw=none, forget plot] fill between[of=utruthA and ltruthA];
  \addplot[colorA, line width=0.8pt]
    table[x=epoch, y=cfg0d, col sep=comma]{csv_tikz/qwen_qwen/truthfulqa_mc2.csv};
  \addplot[name path=utruthC, draw=none, forget plot]
    table[x=epoch, y expr=\thisrow{cfg1d}+\thisrow{cfg1s}, col sep=comma]{csv_tikz/qwen_qwen/truthfulqa_mc2.csv};
  \addplot[name path=ltruthC, draw=none, forget plot]
    table[x=epoch, y expr=\thisrow{cfg1d}-\thisrow{cfg1s}, col sep=comma]{csv_tikz/qwen_qwen/truthfulqa_mc2.csv};
  \addplot[colorC!15, draw=none, forget plot] fill between[of=utruthC and ltruthC];
  \addplot[colorC, line width=0.8pt]
    table[x=epoch, y=cfg1d, col sep=comma]{csv_tikz/qwen_qwen/truthfulqa_mc2.csv};
  \addplot[name path=utruthB, draw=none, forget plot]
    table[x=epoch, y expr=\thisrow{cfg2d}+\thisrow{cfg2s}, col sep=comma]{csv_tikz/qwen_qwen/truthfulqa_mc2.csv};
  \addplot[name path=ltruthB, draw=none, forget plot]
    table[x=epoch, y expr=\thisrow{cfg2d}-\thisrow{cfg2s}, col sep=comma]{csv_tikz/qwen_qwen/truthfulqa_mc2.csv};
  \addplot[colorB!15, draw=none, forget plot] fill between[of=utruthB and ltruthB];
  \addplot[colorB, line width=0.8pt]
    table[x=epoch, y=cfg2d, col sep=comma]{csv_tikz/qwen_qwen/truthfulqa_mc2.csv};
  \addplot[name path=utruthD, draw=none, forget plot]
    table[x=epoch, y expr=\thisrow{cfg3d}+\thisrow{cfg3s}, col sep=comma]{csv_tikz/qwen_qwen/truthfulqa_mc2.csv};
  \addplot[name path=ltruthD, draw=none, forget plot]
    table[x=epoch, y expr=\thisrow{cfg3d}-\thisrow{cfg3s}, col sep=comma]{csv_tikz/qwen_qwen/truthfulqa_mc2.csv};
  \addplot[colorD!15, draw=none, forget plot] fill between[of=utruthD and ltruthD];
  \addplot[colorD, line width=0.8pt]
    table[x=epoch, y=cfg3d, col sep=comma]{csv_tikz/qwen_qwen/truthfulqa_mc2.csv};

  \nextgroupplot[
    title={\tiny Math},
    ylabel={},
    xlabel={\tiny Epoch},
    axis x line*=bottom, axis y line*=left,
    xmin=0, xmax=40, ymin=-5.0, ymax=1.5,
    grid, grid style={line width=.2pt, draw=gray!30},
  ]
  \draw[gray,dashed,line width=0.5pt](axis cs:0,0)--(axis cs:40,0);
  \node[anchor=east, font=\tiny, text=gray!70!black,
    fill=white, fill opacity=0.75, text opacity=1,
    inner sep=1pt, rounded corners=1pt]
    at (axis cs:39,-2.0) {base$=15.0\%$};
  \addplot[name path=uminerA, draw=none, forget plot]
    table[x=epoch, y expr=\thisrow{cfg0d}+\thisrow{cfg0s}, col sep=comma]{csv_tikz/qwen_qwen/minerva_math.csv};
  \addplot[name path=lminerA, draw=none, forget plot]
    table[x=epoch, y expr=\thisrow{cfg0d}-\thisrow{cfg0s}, col sep=comma]{csv_tikz/qwen_qwen/minerva_math.csv};
  \addplot[colorA!15, draw=none, forget plot] fill between[of=uminerA and lminerA];
  \addplot[colorA, line width=0.8pt]
    table[x=epoch, y=cfg0d, col sep=comma]{csv_tikz/qwen_qwen/minerva_math.csv};
  \addplot[name path=uminerC, draw=none, forget plot]
    table[x=epoch, y expr=\thisrow{cfg1d}+\thisrow{cfg1s}, col sep=comma]{csv_tikz/qwen_qwen/minerva_math.csv};
  \addplot[name path=lminerC, draw=none, forget plot]
    table[x=epoch, y expr=\thisrow{cfg1d}-\thisrow{cfg1s}, col sep=comma]{csv_tikz/qwen_qwen/minerva_math.csv};
  \addplot[colorC!15, draw=none, forget plot] fill between[of=uminerC and lminerC];
  \addplot[colorC, line width=0.8pt]
    table[x=epoch, y=cfg1d, col sep=comma]{csv_tikz/qwen_qwen/minerva_math.csv};
  \addplot[name path=uminerB, draw=none, forget plot]
    table[x=epoch, y expr=\thisrow{cfg2d}+\thisrow{cfg2s}, col sep=comma]{csv_tikz/qwen_qwen/minerva_math.csv};
  \addplot[name path=lminerB, draw=none, forget plot]
    table[x=epoch, y expr=\thisrow{cfg2d}-\thisrow{cfg2s}, col sep=comma]{csv_tikz/qwen_qwen/minerva_math.csv};
  \addplot[colorB!15, draw=none, forget plot] fill between[of=uminerB and lminerB];
  \addplot[colorB, line width=0.8pt]
    table[x=epoch, y=cfg2d, col sep=comma]{csv_tikz/qwen_qwen/minerva_math.csv};
  \addplot[name path=uminerD, draw=none, forget plot]
    table[x=epoch, y expr=\thisrow{cfg3d}+\thisrow{cfg3s}, col sep=comma]{csv_tikz/qwen_qwen/minerva_math.csv};
  \addplot[name path=lminerD, draw=none, forget plot]
    table[x=epoch, y expr=\thisrow{cfg3d}-\thisrow{cfg3s}, col sep=comma]{csv_tikz/qwen_qwen/minerva_math.csv};
  \addplot[colorD!15, draw=none, forget plot] fill between[of=uminerD and lminerD];
  \addplot[colorD, line width=0.8pt]
    table[x=epoch, y=cfg3d, col sep=comma]{csv_tikz/qwen_qwen/minerva_math.csv};

  \nextgroupplot[
    title={\tiny HumanEval},
    ylabel={},
    xlabel={\tiny Epoch},
    axis x line*=bottom, axis y line*=left,
    xmin=0, xmax=40, ymin=-1.5, ymax=3.0,
    grid, grid style={line width=.2pt, draw=gray!30},
    legend to name=figlegend,
    legend style={
      draw=gray!60, fill=white,
      nodes={scale=0.50, transform shape},
      text=black, cells={align=left}, row sep=0pt,
      inner sep=3pt, rounded corners=2pt,
      /tikz/every even column/.append style={column sep=0.08cm}},
  ]
  \draw[gray,dashed,line width=0.5pt](axis cs:0,0)--(axis cs:40,0);
  \node[anchor=north west, font=\tiny, text=gray!70!black,
    fill=white, fill opacity=0.75, text opacity=1,
    inner sep=1pt, rounded corners=1pt]
    at (axis cs:1,2.8) {base$=31.9\%$};
  \addplot[name path=uhumanA, draw=none, forget plot]
    table[x=epoch, y expr=\thisrow{cfg0d}+\thisrow{cfg0s}, col sep=comma]{csv_tikz/qwen_qwen/humaneval_1.csv};
  \addplot[name path=lhumanA, draw=none, forget plot]
    table[x=epoch, y expr=\thisrow{cfg0d}-\thisrow{cfg0s}, col sep=comma]{csv_tikz/qwen_qwen/humaneval_1.csv};
  \addplot[colorA!15, draw=none, forget plot] fill between[of=uhumanA and lhumanA];
  \addplot[colorA, line width=0.8pt]
    table[x=epoch, y=cfg0d, col sep=comma]{csv_tikz/qwen_qwen/humaneval_1.csv};
  \addlegendentry{Real Data};
  \addplot[name path=uhumanC, draw=none, forget plot]
    table[x=epoch, y expr=\thisrow{cfg1d}+\thisrow{cfg1s}, col sep=comma]{csv_tikz/qwen_qwen/humaneval_1.csv};
  \addplot[name path=lhumanC, draw=none, forget plot]
    table[x=epoch, y expr=\thisrow{cfg1d}-\thisrow{cfg1s}, col sep=comma]{csv_tikz/qwen_qwen/humaneval_1.csv};
  \addplot[colorC!15, draw=none, forget plot] fill between[of=uhumanC and lhumanC];
  \addplot[colorC, line width=0.8pt]
    table[x=epoch, y=cfg1d, col sep=comma]{csv_tikz/qwen_qwen/humaneval_1.csv};
  \addlegendentry{Synthetic $\tau{=}0.75$};
  \addplot[name path=uhumanB, draw=none, forget plot]
    table[x=epoch, y expr=\thisrow{cfg2d}+\thisrow{cfg2s}, col sep=comma]{csv_tikz/qwen_qwen/humaneval_1.csv};
  \addplot[name path=lhumanB, draw=none, forget plot]
    table[x=epoch, y expr=\thisrow{cfg2d}-\thisrow{cfg2s}, col sep=comma]{csv_tikz/qwen_qwen/humaneval_1.csv};
  \addplot[colorB!15, draw=none, forget plot] fill between[of=uhumanB and lhumanB];
  \addplot[colorB, line width=0.8pt]
    table[x=epoch, y=cfg2d, col sep=comma]{csv_tikz/qwen_qwen/humaneval_1.csv};
  \addlegendentry{Synthetic $\tau{=}1.0$};
  \addplot[name path=uhumanD, draw=none, forget plot]
    table[x=epoch, y expr=\thisrow{cfg3d}+\thisrow{cfg3s}, col sep=comma]{csv_tikz/qwen_qwen/humaneval_1.csv};
  \addplot[name path=lhumanD, draw=none, forget plot]
    table[x=epoch, y expr=\thisrow{cfg3d}-\thisrow{cfg3s}, col sep=comma]{csv_tikz/qwen_qwen/humaneval_1.csv};
  \addplot[colorD!15, draw=none, forget plot] fill between[of=uhumanD and lhumanD];
  \addplot[colorD, line width=0.8pt]
    table[x=epoch, y=cfg3d, col sep=comma]{csv_tikz/qwen_qwen/humaneval_1.csv};
  \addlegendentry{Synthetic $\tau{=}1.25$};

  \nextgroupplot[
    axis lines=none, xtick=\empty, ytick=\empty,
    xmin=0,xmax=1,ymin=0,ymax=1,
    xlabel={}, ylabel={},
  ]
\end{groupplot}
\node[anchor=center] at (group c4r2.center) {\pgfplotslegendfromname{figlegend}};
\end{tikzpicture}
}
\caption{\small \textbf{Self-generated data produces transient gains on structured reasoning, math, and code for Qwen2.5-0.5B, while generic replay does not.}
$\Delta$ performance relative to the frozen base model over 40 epochs. Synthetic self-generated corpora at three temperatures are compared against a matched Common Corpus replay baseline. Shaded bands show $\pm 1$ std.\ across subsets.}
\label{fig:qwen05b_cc_vs_syn}
\vspace{-0.5em}
\end{figure}

%% file: figure_llama_llama.tex
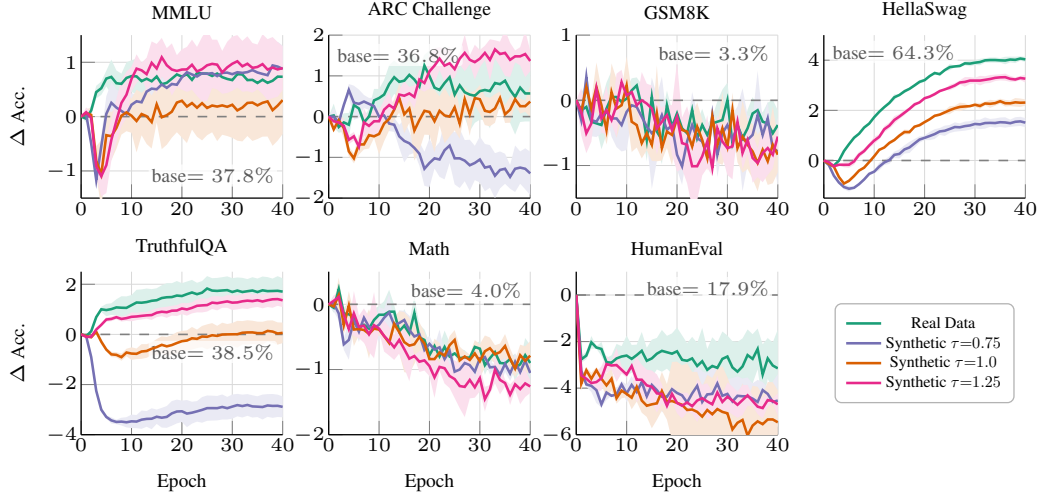
\begin{figure}[t]
\vspace{-0.5em}
\centering
\resizebox{\textwidth}{!}{%
\begin{tikzpicture}
\begin{groupplot}[
  group style={group size=4 by 2, horizontal sep=5mm, vertical sep=8mm},
  width=0.27\linewidth, height=0.24\linewidth,
  xtick={0,10,20,30,40},
  tick label style={font=\fontsize{4}{4}\selectfont},
  label style={font=\tiny},
  title style={font=\tiny, yshift=-1ex},
]

  \nextgroupplot[
    title={\tiny MMLU},
    ylabel={\tiny $\Delta$ Acc.},
    xlabel={},
    axis x line*=bottom, axis y line*=left,
    xmin=0, xmax=40, ymin=-1.5, ymax=1.5,
    grid, grid style={line width=.2pt, draw=gray!30},
  ]
  \draw[gray,dashed,line width=0.5pt](axis cs:0,0)--(axis cs:40,0);
  \node[anchor=south east, font=\tiny, text=gray!70!black,
    fill=white, fill opacity=0.75, text opacity=1,
    inner sep=1pt, rounded corners=1pt]
    at (axis cs:39,-1.3) {base$=37.8\%$};
  \addplot[name path=ummluA, draw=none, forget plot]
    table[x=epoch, y expr=\thisrow{cfg0d}+\thisrow{cfg0s}, col sep=comma]{csv_tikz/llama-llama/mmlu.csv};
  \addplot[name path=lmmluA, draw=none, forget plot]
    table[x=epoch, y expr=\thisrow{cfg0d}-\thisrow{cfg0s}, col sep=comma]{csv_tikz/llama-llama/mmlu.csv};
  \addplot[colorA!15, draw=none, forget plot] fill between[of=ummluA and lmmluA];
  \addplot[colorA, line width=0.8pt]
    table[x=epoch, y=cfg0d, col sep=comma]{csv_tikz/llama-llama/mmlu.csv};
  \addplot[name path=ummluC, draw=none, forget plot]
    table[x=epoch, y expr=\thisrow{cfg1d}+\thisrow{cfg1s}, col sep=comma]{csv_tikz/llama-llama/mmlu.csv};
  \addplot[name path=lmmluC, draw=none, forget plot]
    table[x=epoch, y expr=\thisrow{cfg1d}-\thisrow{cfg1s}, col sep=comma]{csv_tikz/llama-llama/mmlu.csv};
  \addplot[colorC!15, draw=none, forget plot] fill between[of=ummluC and lmmluC];
  \addplot[colorC, line width=0.8pt]
    table[x=epoch, y=cfg1d, col sep=comma]{csv_tikz/llama-llama/mmlu.csv};
  \addplot[name path=ummluB, draw=none, forget plot]
    table[x=epoch, y expr=\thisrow{cfg2d}+\thisrow{cfg2s}, col sep=comma]{csv_tikz/llama-llama/mmlu.csv};
  \addplot[name path=lmmluB, draw=none, forget plot]
    table[x=epoch, y expr=\thisrow{cfg2d}-\thisrow{cfg2s}, col sep=comma]{csv_tikz/llama-llama/mmlu.csv};
  \addplot[colorB!15, draw=none, forget plot] fill between[of=ummluB and lmmluB];
  \addplot[colorB, line width=0.8pt]
    table[x=epoch, y=cfg2d, col sep=comma]{csv_tikz/llama-llama/mmlu.csv};
  \addplot[name path=ummluD, draw=none, forget plot]
    table[x=epoch, y expr=\thisrow{cfg3d}+\thisrow{cfg3s}, col sep=comma]{csv_tikz/llama-llama/mmlu.csv};
  \addplot[name path=lmmluD, draw=none, forget plot]
    table[x=epoch, y expr=\thisrow{cfg3d}-\thisrow{cfg3s}, col sep=comma]{csv_tikz/llama-llama/mmlu.csv};
  \addplot[colorD!15, draw=none, forget plot] fill between[of=ummluD and lmmluD];
  \addplot[colorD, line width=0.8pt]
    table[x=epoch, y=cfg3d, col sep=comma]{csv_tikz/llama-llama/mmlu.csv};

  \nextgroupplot[
    title={\tiny ARC Challenge},
    ylabel={},
    xlabel={},
    axis x line*=bottom, axis y line*=left,
    xmin=0, xmax=40, ymin=-2.0, ymax=2.0,
    grid, grid style={line width=.2pt, draw=gray!30},
  ]
  \draw[gray,dashed,line width=0.5pt](axis cs:0,0)--(axis cs:40,0);
  \node[anchor=north west, font=\tiny, text=gray!70!black,
    fill=white, fill opacity=0.75, text opacity=1,
    inner sep=1pt, rounded corners=1pt]
    at (axis cs:1,1.8) {base$=36.8\%$};
  \addplot[name path=uarcchA, draw=none, forget plot]
    table[x=epoch, y expr=\thisrow{cfg0d}+\thisrow{cfg0s}, col sep=comma]{csv_tikz/llama-llama/arc_challenge.csv};
  \addplot[name path=larcchA, draw=none, forget plot]
    table[x=epoch, y expr=\thisrow{cfg0d}-\thisrow{cfg0s}, col sep=comma]{csv_tikz/llama-llama/arc_challenge.csv};
  \addplot[colorA!15, draw=none, forget plot] fill between[of=uarcchA and larcchA];
  \addplot[colorA, line width=0.8pt]
    table[x=epoch, y=cfg0d, col sep=comma]{csv_tikz/llama-llama/arc_challenge.csv};
  \addplot[name path=uarcchC, draw=none, forget plot]
    table[x=epoch, y expr=\thisrow{cfg1d}+\thisrow{cfg1s}, col sep=comma]{csv_tikz/llama-llama/arc_challenge.csv};
  \addplot[name path=larcchC, draw=none, forget plot]
    table[x=epoch, y expr=\thisrow{cfg1d}-\thisrow{cfg1s}, col sep=comma]{csv_tikz/llama-llama/arc_challenge.csv};
  \addplot[colorC!15, draw=none, forget plot] fill between[of=uarcchC and larcchC];
  \addplot[colorC, line width=0.8pt]
    table[x=epoch, y=cfg1d, col sep=comma]{csv_tikz/llama-llama/arc_challenge.csv};
  \addplot[name path=uarcchB, draw=none, forget plot]
    table[x=epoch, y expr=\thisrow{cfg2d}+\thisrow{cfg2s}, col sep=comma]{csv_tikz/llama-llama/arc_challenge.csv};
  \addplot[name path=larcchB, draw=none, forget plot]
    table[x=epoch, y expr=\thisrow{cfg2d}-\thisrow{cfg2s}, col sep=comma]{csv_tikz/llama-llama/arc_challenge.csv};
  \addplot[colorB!15, draw=none, forget plot] fill between[of=uarcchB and larcchB];
  \addplot[colorB, line width=0.8pt]
    table[x=epoch, y=cfg2d, col sep=comma]{csv_tikz/llama-llama/arc_challenge.csv};
  \addplot[name path=uarcchD, draw=none, forget plot]
    table[x=epoch, y expr=\thisrow{cfg3d}+\thisrow{cfg3s}, col sep=comma]{csv_tikz/llama-llama/arc_challenge.csv};
  \addplot[name path=larcchD, draw=none, forget plot]
    table[x=epoch, y expr=\thisrow{cfg3d}-\thisrow{cfg3s}, col sep=comma]{csv_tikz/llama-llama/arc_challenge.csv};
  \addplot[colorD!15, draw=none, forget plot] fill between[of=uarcchD and larcchD];
  \addplot[colorD, line width=0.8pt]
    table[x=epoch, y=cfg3d, col sep=comma]{csv_tikz/llama-llama/arc_challenge.csv};

  \nextgroupplot[
    title={\tiny GSM8K},
    ylabel={},
    xlabel={},
    axis x line*=bottom, axis y line*=left,
    xmin=0, xmax=40, ymin=-1.5, ymax=1.0,
    grid, grid style={line width=.2pt, draw=gray!30},
  ]
  \draw[gray,dashed,line width=0.5pt](axis cs:0,0)--(axis cs:40,0);
  \node[anchor=north east, font=\tiny, text=gray!70!black,
    fill=white, fill opacity=0.75, text opacity=1,
    inner sep=1pt, rounded corners=1pt]
    at (axis cs:39,0.9) {base$=3.3\%$};
  \addplot[name path=ugsm8kA, draw=none, forget plot]
    table[x=epoch, y expr=\thisrow{cfg0d}+\thisrow{cfg0s}, col sep=comma]{csv_tikz/llama-llama/gsm8k.csv};
  \addplot[name path=lgsm8kA, draw=none, forget plot]
    table[x=epoch, y expr=\thisrow{cfg0d}-\thisrow{cfg0s}, col sep=comma]{csv_tikz/llama-llama/gsm8k.csv};
  \addplot[colorA!15, draw=none, forget plot] fill between[of=ugsm8kA and lgsm8kA];
  \addplot[colorA, line width=0.8pt]
    table[x=epoch, y=cfg0d, col sep=comma]{csv_tikz/llama-llama/gsm8k.csv};
  \addplot[name path=ugsm8kC, draw=none, forget plot]
    table[x=epoch, y expr=\thisrow{cfg1d}+\thisrow{cfg1s}, col sep=comma]{csv_tikz/llama-llama/gsm8k.csv};
  \addplot[name path=lgsm8kC, draw=none, forget plot]
    table[x=epoch, y expr=\thisrow{cfg1d}-\thisrow{cfg1s}, col sep=comma]{csv_tikz/llama-llama/gsm8k.csv};
  \addplot[colorC!15, draw=none, forget plot] fill between[of=ugsm8kC and lgsm8kC];
  \addplot[colorC, line width=0.8pt]
    table[x=epoch, y=cfg1d, col sep=comma]{csv_tikz/llama-llama/gsm8k.csv};
  \addplot[name path=ugsm8kB, draw=none, forget plot]
    table[x=epoch, y expr=\thisrow{cfg2d}+\thisrow{cfg2s}, col sep=comma]{csv_tikz/llama-llama/gsm8k.csv};
  \addplot[name path=lgsm8kB, draw=none, forget plot]
    table[x=epoch, y expr=\thisrow{cfg2d}-\thisrow{cfg2s}, col sep=comma]{csv_tikz/llama-llama/gsm8k.csv};
  \addplot[colorB!15, draw=none, forget plot] fill between[of=ugsm8kB and lgsm8kB];
  \addplot[colorB, line width=0.8pt]
    table[x=epoch, y=cfg2d, col sep=comma]{csv_tikz/llama-llama/gsm8k.csv};
  \addplot[name path=ugsm8kD, draw=none, forget plot]
    table[x=epoch, y expr=\thisrow{cfg3d}+\thisrow{cfg3s}, col sep=comma]{csv_tikz/llama-llama/gsm8k.csv};
  \addplot[name path=lgsm8kD, draw=none, forget plot]
    table[x=epoch, y expr=\thisrow{cfg3d}-\thisrow{cfg3s}, col sep=comma]{csv_tikz/llama-llama/gsm8k.csv};
  \addplot[colorD!15, draw=none, forget plot] fill between[of=ugsm8kD and lgsm8kD];
  \addplot[colorD, line width=0.8pt]
    table[x=epoch, y=cfg3d, col sep=comma]{csv_tikz/llama-llama/gsm8k.csv};

  \nextgroupplot[
    title={\tiny HellaSwag},
    ylabel={},
    xlabel={},
    axis x line*=bottom, axis y line*=left,
    xmin=0, xmax=40, ymin=-1.5, ymax=5.0,
    grid, grid style={line width=.2pt, draw=gray!30},
  ]
  \draw[gray,dashed,line width=0.5pt](axis cs:0,0)--(axis cs:40,0);
  \node[anchor=north west, font=\tiny, text=gray!70!black,
    fill=white, fill opacity=0.75, text opacity=1,
    inner sep=1pt, rounded corners=1pt]
    at (axis cs:1,4.8) {base$=64.3\%$};
  \addplot[name path=uhellaA, draw=none, forget plot]
    table[x=epoch, y expr=\thisrow{cfg0d}+\thisrow{cfg0s}, col sep=comma]{csv_tikz/llama-llama/hellaswag.csv};
  \addplot[name path=lhellaA, draw=none, forget plot]
    table[x=epoch, y expr=\thisrow{cfg0d}-\thisrow{cfg0s}, col sep=comma]{csv_tikz/llama-llama/hellaswag.csv};
  \addplot[colorA!15, draw=none, forget plot] fill between[of=uhellaA and lhellaA];
  \addplot[colorA, line width=0.8pt]
    table[x=epoch, y=cfg0d, col sep=comma]{csv_tikz/llama-llama/hellaswag.csv};
  \addplot[name path=uhellaC, draw=none, forget plot]
    table[x=epoch, y expr=\thisrow{cfg1d}+\thisrow{cfg1s}, col sep=comma]{csv_tikz/llama-llama/hellaswag.csv};
  \addplot[name path=lhellaC, draw=none, forget plot]
    table[x=epoch, y expr=\thisrow{cfg1d}-\thisrow{cfg1s}, col sep=comma]{csv_tikz/llama-llama/hellaswag.csv};
  \addplot[colorC!15, draw=none, forget plot] fill between[of=uhellaC and lhellaC];
  \addplot[colorC, line width=0.8pt]
    table[x=epoch, y=cfg1d, col sep=comma]{csv_tikz/llama-llama/hellaswag.csv};
  \addplot[name path=uhellaB, draw=none, forget plot]
    table[x=epoch, y expr=\thisrow{cfg2d}+\thisrow{cfg2s}, col sep=comma]{csv_tikz/llama-llama/hellaswag.csv};
  \addplot[name path=lhellaB, draw=none, forget plot]
    table[x=epoch, y expr=\thisrow{cfg2d}-\thisrow{cfg2s}, col sep=comma]{csv_tikz/llama-llama/hellaswag.csv};
  \addplot[colorB!15, draw=none, forget plot] fill between[of=uhellaB and lhellaB];
  \addplot[colorB, line width=0.8pt]
    table[x=epoch, y=cfg2d, col sep=comma]{csv_tikz/llama-llama/hellaswag.csv};
  \addplot[name path=uhellaD, draw=none, forget plot]
    table[x=epoch, y expr=\thisrow{cfg3d}+\thisrow{cfg3s}, col sep=comma]{csv_tikz/llama-llama/hellaswag.csv};
  \addplot[name path=lhellaD, draw=none, forget plot]
    table[x=epoch, y expr=\thisrow{cfg3d}-\thisrow{cfg3s}, col sep=comma]{csv_tikz/llama-llama/hellaswag.csv};
  \addplot[colorD!15, draw=none, forget plot] fill between[of=uhellaD and lhellaD];
  \addplot[colorD, line width=0.8pt]
    table[x=epoch, y=cfg3d, col sep=comma]{csv_tikz/llama-llama/hellaswag.csv};

  \nextgroupplot[
    title={\tiny TruthfulQA},
    ylabel={\tiny $\Delta$ Acc.},
    xlabel={\tiny Epoch},
    axis x line*=bottom, axis y line*=left,
    xmin=0, xmax=40, ymin=-4.0, ymax=2.5,
    grid, grid style={line width=.2pt, draw=gray!30},
  ]
  \draw[gray,dashed,line width=0.5pt](axis cs:0,0)--(axis cs:40,0);
  \node[anchor=east, font=\tiny, text=gray!70!black,
    fill=white, fill opacity=0.75, text opacity=1,
    inner sep=1pt, rounded corners=1pt]
    at (axis cs:39,-0.7) {base$=38.5\%$};
  \addplot[name path=utruthA, draw=none, forget plot]
    table[x=epoch, y expr=\thisrow{cfg0d}+\thisrow{cfg0s}, col sep=comma]{csv_tikz/llama-llama/truthfulqa_mc2.csv};
  \addplot[name path=ltruthA, draw=none, forget plot]
    table[x=epoch, y expr=\thisrow{cfg0d}-\thisrow{cfg0s}, col sep=comma]{csv_tikz/llama-llama/truthfulqa_mc2.csv};
  \addplot[colorA!15, draw=none, forget plot] fill between[of=utruthA and ltruthA];
  \addplot[colorA, line width=0.8pt]
    table[x=epoch, y=cfg0d, col sep=comma]{csv_tikz/llama-llama/truthfulqa_mc2.csv};
  \addplot[name path=utruthC, draw=none, forget plot]
    table[x=epoch, y expr=\thisrow{cfg1d}+\thisrow{cfg1s}, col sep=comma]{csv_tikz/llama-llama/truthfulqa_mc2.csv};
  \addplot[name path=ltruthC, draw=none, forget plot]
    table[x=epoch, y expr=\thisrow{cfg1d}-\thisrow{cfg1s}, col sep=comma]{csv_tikz/llama-llama/truthfulqa_mc2.csv};
  \addplot[colorC!15, draw=none, forget plot] fill between[of=utruthC and ltruthC];
  \addplot[colorC, line width=0.8pt]
    table[x=epoch, y=cfg1d, col sep=comma]{csv_tikz/llama-llama/truthfulqa_mc2.csv};
  \addplot[name path=utruthB, draw=none, forget plot]
    table[x=epoch, y expr=\thisrow{cfg2d}+\thisrow{cfg2s}, col sep=comma]{csv_tikz/llama-llama/truthfulqa_mc2.csv};
  \addplot[name path=ltruthB, draw=none, forget plot]
    table[x=epoch, y expr=\thisrow{cfg2d}-\thisrow{cfg2s}, col sep=comma]{csv_tikz/llama-llama/truthfulqa_mc2.csv};
  \addplot[colorB!15, draw=none, forget plot] fill between[of=utruthB and ltruthB];
  \addplot[colorB, line width=0.8pt]
    table[x=epoch, y=cfg2d, col sep=comma]{csv_tikz/llama-llama/truthfulqa_mc2.csv};
  \addplot[name path=utruthD, draw=none, forget plot]
    table[x=epoch, y expr=\thisrow{cfg3d}+\thisrow{cfg3s}, col sep=comma]{csv_tikz/llama-llama/truthfulqa_mc2.csv};
  \addplot[name path=ltruthD, draw=none, forget plot]
    table[x=epoch, y expr=\thisrow{cfg3d}-\thisrow{cfg3s}, col sep=comma]{csv_tikz/llama-llama/truthfulqa_mc2.csv};
  \addplot[colorD!15, draw=none, forget plot] fill between[of=utruthD and ltruthD];
  \addplot[colorD, line width=0.8pt]
    table[x=epoch, y=cfg3d, col sep=comma]{csv_tikz/llama-llama/truthfulqa_mc2.csv};

  \nextgroupplot[
    title={\tiny Math},
    ylabel={},
    xlabel={\tiny Epoch},
    axis x line*=bottom, axis y line*=left,
    xmin=0, xmax=40, ymin=-2.0, ymax=0.5,
    grid, grid style={line width=.2pt, draw=gray!30},
  ]
  \draw[gray,dashed,line width=0.5pt](axis cs:0,0)--(axis cs:40,0);
  \node[anchor=north east, font=\tiny, text=gray!70!black,
    fill=white, fill opacity=0.75, text opacity=1,
    inner sep=1pt, rounded corners=1pt]
    at (axis cs:39,0.4) {base$=4.0\%$};
  \addplot[name path=uminerA, draw=none, forget plot]
    table[x=epoch, y expr=\thisrow{cfg0d}+\thisrow{cfg0s}, col sep=comma]{csv_tikz/llama-llama/minerva_math.csv};
  \addplot[name path=lminerA, draw=none, forget plot]
    table[x=epoch, y expr=\thisrow{cfg0d}-\thisrow{cfg0s}, col sep=comma]{csv_tikz/llama-llama/minerva_math.csv};
  \addplot[colorA!15, draw=none, forget plot] fill between[of=uminerA and lminerA];
  \addplot[colorA, line width=0.8pt]
    table[x=epoch, y=cfg0d, col sep=comma]{csv_tikz/llama-llama/minerva_math.csv};
  \addplot[name path=uminerC, draw=none, forget plot]
    table[x=epoch, y expr=\thisrow{cfg1d}+\thisrow{cfg1s}, col sep=comma]{csv_tikz/llama-llama/minerva_math.csv};
  \addplot[name path=lminerC, draw=none, forget plot]
    table[x=epoch, y expr=\thisrow{cfg1d}-\thisrow{cfg1s}, col sep=comma]{csv_tikz/llama-llama/minerva_math.csv};
  \addplot[colorC!15, draw=none, forget plot] fill between[of=uminerC and lminerC];
  \addplot[colorC, line width=0.8pt]
    table[x=epoch, y=cfg1d, col sep=comma]{csv_tikz/llama-llama/minerva_math.csv};
  \addplot[name path=uminerB, draw=none, forget plot]
    table[x=epoch, y expr=\thisrow{cfg2d}+\thisrow{cfg2s}, col sep=comma]{csv_tikz/llama-llama/minerva_math.csv};
  \addplot[name path=lminerB, draw=none, forget plot]
    table[x=epoch, y expr=\thisrow{cfg2d}-\thisrow{cfg2s}, col sep=comma]{csv_tikz/llama-llama/minerva_math.csv};
  \addplot[colorB!15, draw=none, forget plot] fill between[of=uminerB and lminerB];
  \addplot[colorB, line width=0.8pt]
    table[x=epoch, y=cfg2d, col sep=comma]{csv_tikz/llama-llama/minerva_math.csv};
  \addplot[name path=uminerD, draw=none, forget plot]
    table[x=epoch, y expr=\thisrow{cfg3d}+\thisrow{cfg3s}, col sep=comma]{csv_tikz/llama-llama/minerva_math.csv};
  \addplot[name path=lminerD, draw=none, forget plot]
    table[x=epoch, y expr=\thisrow{cfg3d}-\thisrow{cfg3s}, col sep=comma]{csv_tikz/llama-llama/minerva_math.csv};
  \addplot[colorD!15, draw=none, forget plot] fill between[of=uminerD and lminerD];
  \addplot[colorD, line width=0.8pt]
    table[x=epoch, y=cfg3d, col sep=comma]{csv_tikz/llama-llama/minerva_math.csv};

  \nextgroupplot[
    title={\tiny HumanEval},
    ylabel={},
    xlabel={\tiny Epoch},
    axis x line*=bottom, axis y line*=left,
    xmin=0, xmax=40, ymin=-6.0, ymax=1.0,
    grid, grid style={line width=.2pt, draw=gray!30},
    legend to name=figlegend,
    legend style={
      draw=gray!60, fill=white,
      nodes={scale=0.50, transform shape},
      text=black, cells={align=left}, row sep=0pt,
      inner sep=3pt, rounded corners=2pt,
      /tikz/every even column/.append style={column sep=0.08cm}},
  ]
  \draw[gray,dashed,line width=0.5pt](axis cs:0,0)--(axis cs:40,0);
  \node[anchor=north east, font=\tiny, text=gray!70!black,
    fill=white, fill opacity=0.75, text opacity=1,
    inner sep=1pt, rounded corners=1pt]
    at (axis cs:39,0.8) {base$=17.9\%$};
  \addplot[name path=uhumanA, draw=none, forget plot]
    table[x=epoch, y expr=\thisrow{cfg0d}+\thisrow{cfg0s}, col sep=comma]{csv_tikz/llama-llama/humaneval_1.csv};
  \addplot[name path=lhumanA, draw=none, forget plot]
    table[x=epoch, y expr=\thisrow{cfg0d}-\thisrow{cfg0s}, col sep=comma]{csv_tikz/llama-llama/humaneval_1.csv};
  \addplot[colorA!15, draw=none, forget plot] fill between[of=uhumanA and lhumanA];
  \addplot[colorA, line width=0.8pt]
    table[x=epoch, y=cfg0d, col sep=comma]{csv_tikz/llama-llama/humaneval_1.csv};
  \addlegendentry{Real Data};
  \addplot[name path=uhumanC, draw=none, forget plot]
    table[x=epoch, y expr=\thisrow{cfg1d}+\thisrow{cfg1s}, col sep=comma]{csv_tikz/llama-llama/humaneval_1.csv};
  \addplot[name path=lhumanC, draw=none, forget plot]
    table[x=epoch, y expr=\thisrow{cfg1d}-\thisrow{cfg1s}, col sep=comma]{csv_tikz/llama-llama/humaneval_1.csv};
  \addplot[colorC!15, draw=none, forget plot] fill between[of=uhumanC and lhumanC];
  \addplot[colorC, line width=0.8pt]
    table[x=epoch, y=cfg1d, col sep=comma]{csv_tikz/llama-llama/humaneval_1.csv};
  \addlegendentry{Synthetic $\tau{=}0.75$};
  \addplot[name path=uhumanB, draw=none, forget plot]
    table[x=epoch, y expr=\thisrow{cfg2d}+\thisrow{cfg2s}, col sep=comma]{csv_tikz/llama-llama/humaneval_1.csv};
  \addplot[name path=lhumanB, draw=none, forget plot]
    table[x=epoch, y expr=\thisrow{cfg2d}-\thisrow{cfg2s}, col sep=comma]{csv_tikz/llama-llama/humaneval_1.csv};
  \addplot[colorB!15, draw=none, forget plot] fill between[of=uhumanB and lhumanB];
  \addplot[colorB, line width=0.8pt]
    table[x=epoch, y=cfg2d, col sep=comma]{csv_tikz/llama-llama/humaneval_1.csv};
  \addlegendentry{Synthetic $\tau{=}1.0$};
  \addplot[name path=uhumanD, draw=none, forget plot]
    table[x=epoch, y expr=\thisrow{cfg3d}+\thisrow{cfg3s}, col sep=comma]{csv_tikz/llama-llama/humaneval_1.csv};
  \addplot[name path=lhumanD, draw=none, forget plot]
    table[x=epoch, y expr=\thisrow{cfg3d}-\thisrow{cfg3s}, col sep=comma]{csv_tikz/llama-llama/humaneval_1.csv};
  \addplot[colorD!15, draw=none, forget plot] fill between[of=uhumanD and lhumanD];
  \addplot[colorD, line width=0.8pt]
    table[x=epoch, y=cfg3d, col sep=comma]{csv_tikz/llama-llama/humaneval_1.csv};
  \addlegendentry{Synthetic $\tau{=}1.25$};

  \nextgroupplot[
    axis lines=none, xtick=\empty, ytick=\empty,
    xmin=0,xmax=1,ymin=0,ymax=1,
    xlabel={}, ylabel={},
  ]
\end{groupplot}
\node[anchor=center] at (group c4r2.center) {\pgfplotslegendfromname{figlegend}};
\end{tikzpicture}
}
\caption{\small \textbf{The identical protocol on LLaMA-3.2-1B yields only narrow comprehension gains, with no improvement on math or code.}
$\Delta$ performance relative to the frozen base model over 40 epochs. Unlike Qwen, LLaMA shows no GSM8K or Minerva-MATH gains under any condition, and HumanEval degrades substantially. Shaded bands show $\pm 1$ std.\ across subsets.}
\vspace{-0.5em}
\label{fig:llama_cc_vs_syn}
\end{figure}

%% file: figure_hist.tex
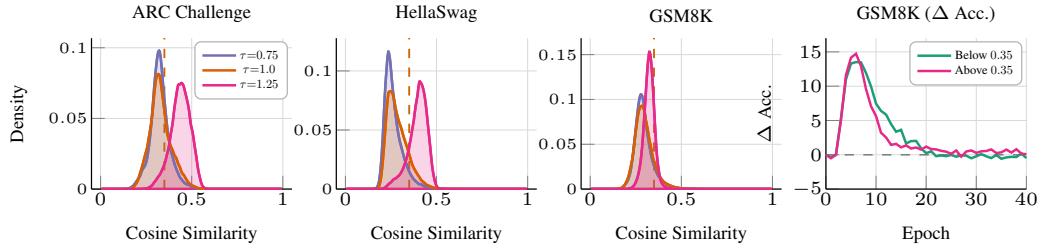
\begin{figure}[t]
\centering
\resizebox{\textwidth}{!}{%
\begin{tikzpicture}
\begin{groupplot}[
  group style={group size=4 by 1, horizontal sep=5mm},
  width=0.275\linewidth, height=0.235\linewidth,
  label style={font=\tiny},
  title style={font=\tiny, yshift=-1ex},
  tick label style={font=\fontsize{4}{4}\selectfont},
]

  \nextgroupplot[
    title={\tiny ARC Challenge},
    ylabel={\tiny Density},
    xlabel={\tiny Cosine Similarity},
    axis x line*=bottom, axis y line*=left,
    xmin=-0.05, xmax=1.05, ymin=0,
    scaled y ticks=false,
    yticklabel style={/pgf/number format/fixed, font=\fontsize{4}{4}\selectfont},
    grid, grid style={line width=.2pt, draw=gray!30},
    legend style={at={(0.98,0.98)}, anchor=north east,
      nodes={scale=0.40, transform shape}, draw=gray!60, fill=white,
      cells={align=left}, row sep=0pt, inner sep=2pt, rounded corners=2pt},
  ]
  \draw[orange!80!black, dashed, line width=0.6pt]
    (axis cs:0.35,0) -- (axis cs:0.35,1);
  \addplot[colorC, fill=colorC, fill opacity=0.20, line width=0.8pt, smooth]
    table[x=bin_center, y=t075, col sep=comma]{csv_tikz/hist/arc_challenge.csv} \closedcycle;
  \addplot[colorB, fill=colorB, fill opacity=0.20, line width=0.8pt, smooth]
    table[x=bin_center, y=t10,  col sep=comma]{csv_tikz/hist/arc_challenge.csv} \closedcycle;
  \addplot[colorD, fill=colorD, fill opacity=0.20, line width=0.8pt, smooth]
    table[x=bin_center, y=t125, col sep=comma]{csv_tikz/hist/arc_challenge.csv} \closedcycle;
  \addplot[colorC, line width=0.8pt, smooth, forget plot]
    table[x=bin_center, y=t075, col sep=comma]{csv_tikz/hist/arc_challenge.csv};
  \addplot[colorB, line width=0.8pt, smooth, forget plot]
    table[x=bin_center, y=t10,  col sep=comma]{csv_tikz/hist/arc_challenge.csv};
  \addplot[colorD, line width=0.8pt, smooth, forget plot]
    table[x=bin_center, y=t125, col sep=comma]{csv_tikz/hist/arc_challenge.csv};
  \addlegendentry{$\tau{=}0.75$};
  \addlegendentry{$\tau{=}1.0$};
  \addlegendentry{$\tau{=}1.25$};

  \nextgroupplot[
    title={\tiny HellaSwag},
    ylabel={},
    xlabel={\tiny Cosine Similarity},
    axis x line*=bottom, axis y line*=left,
    xmin=-0.05, xmax=1.05, ymin=0,
    scaled y ticks=false,
    yticklabel style={/pgf/number format/fixed, font=\fontsize{4}{4}\selectfont},
    grid, grid style={line width=.2pt, draw=gray!30},
  ]
  \draw[orange!80!black, dashed, line width=0.6pt]
    (axis cs:0.35,0) -- (axis cs:0.35,1);
  \addplot[colorC, fill=colorC, fill opacity=0.20, line width=0.8pt, smooth]
    table[x=bin_center, y=t075, col sep=comma]{csv_tikz/hist/hellaswag.csv} \closedcycle;
  \addplot[colorB, fill=colorB, fill opacity=0.20, line width=0.8pt, smooth]
    table[x=bin_center, y=t10,  col sep=comma]{csv_tikz/hist/hellaswag.csv} \closedcycle;
  \addplot[colorD, fill=colorD, fill opacity=0.20, line width=0.8pt, smooth]
    table[x=bin_center, y=t125, col sep=comma]{csv_tikz/hist/hellaswag.csv} \closedcycle;
  \addplot[colorC, line width=0.8pt, smooth, forget plot]
    table[x=bin_center, y=t075, col sep=comma]{csv_tikz/hist/hellaswag.csv};
  \addplot[colorB, line width=0.8pt, smooth, forget plot]
    table[x=bin_center, y=t10,  col sep=comma]{csv_tikz/hist/hellaswag.csv};
  \addplot[colorD, line width=0.8pt, smooth, forget plot]
    table[x=bin_center, y=t125, col sep=comma]{csv_tikz/hist/hellaswag.csv};

  \nextgroupplot[
    title={\tiny GSM8K},
    ylabel={},
    xlabel={\tiny Cosine Similarity},
    axis x line*=bottom, axis y line*=left,
    xmin=-0.05, xmax=1.05, ymin=0,
    scaled y ticks=false,
    yticklabel style={/pgf/number format/fixed, font=\fontsize{4}{4}\selectfont},
    grid, grid style={line width=.2pt, draw=gray!30},
  ]
  \draw[orange!80!black, dashed, line width=0.6pt]
    (axis cs:0.35,0) -- (axis cs:0.35,1);
  \addplot[colorC, fill=colorC, fill opacity=0.20, line width=0.8pt, smooth]
    table[x=bin_center, y=t075, col sep=comma]{csv_tikz/hist/gsm8k.csv} \closedcycle;
  \addplot[colorB, fill=colorB, fill opacity=0.20, line width=0.8pt, smooth]
    table[x=bin_center, y=t10,  col sep=comma]{csv_tikz/hist/gsm8k.csv} \closedcycle;
  \addplot[colorD, fill=colorD, fill opacity=0.20, line width=0.8pt, smooth]
    table[x=bin_center, y=t125, col sep=comma]{csv_tikz/hist/gsm8k.csv} \closedcycle;
  \addplot[colorC, line width=0.8pt, smooth, forget plot]
    table[x=bin_center, y=t075, col sep=comma]{csv_tikz/hist/gsm8k.csv};
  \addplot[colorB, line width=0.8pt, smooth, forget plot]
    table[x=bin_center, y=t10,  col sep=comma]{csv_tikz/hist/gsm8k.csv};
  \addplot[colorD, line width=0.8pt, smooth, forget plot]
    table[x=bin_center, y=t125, col sep=comma]{csv_tikz/hist/gsm8k.csv};

  \nextgroupplot[
    title={\tiny GSM8K ($\Delta$ Acc.)},
    ylabel={\tiny $\Delta$ Acc.},
    xlabel={\tiny Epoch},
    axis x line*=bottom, axis y line*=left,
    xmin=0, xmax=40, ymin=-5.0, ymax=17.0,
    grid, grid style={line width=.2pt, draw=gray!30},
    legend style={at={(0.98,0.98)}, anchor=north east,
      nodes={scale=0.40, transform shape}, draw=gray!60, fill=white,
      cells={align=left}, row sep=0pt, inner sep=2pt, rounded corners=2pt},
  ]
  \draw[gray, dashed, line width=0.5pt](axis cs:0,0)--(axis cs:40,0);
  \addplot[colorA, line width=0.8pt]
    table[x=epoch, y=cfg0d, col sep=comma]{csv_tikz/gsm8k/gsm8k.csv};
  \addlegendentry{Below $0.35$};
  \addplot[colorD, line width=0.8pt]
    table[x=epoch, y=cfg1d, col sep=comma]{csv_tikz/gsm8k/gsm8k.csv};
  \addlegendentry{Above $0.35$};

\end{groupplot}
\end{tikzpicture}
}
\caption{\small \textbf{Benchmark proximity does not explain the gains: training on semantically distant samples produces the same GSM8K improvement as training on higher-similarity samples.}
Left three panels: max cosine similarity distributions between synthetic corpus and benchmark items. Right panel: GSM8K $\Delta$ accuracy for Qwen $\tau{=}1.25$ subsets partitioned below vs.\ above the 0.35 similarity threshold.}
\label{fig:hist_acc}
\vspace{-0.5em}
\end{figure}

%% file: figure_qwen_other.tex
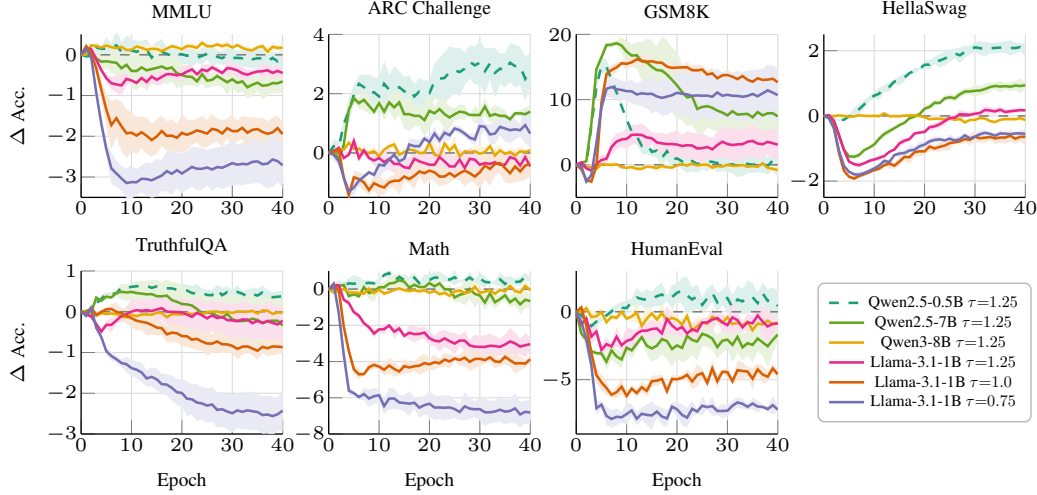
\begin{figure}[t]
\centering
\resizebox{\textwidth}{!}{%
\begin{tikzpicture}
\usepgfplotslibrary{fillbetween}
\begin{groupplot}[
  group style={group size=4 by 2, horizontal sep=5mm, vertical sep=8mm},
  width=0.27\linewidth, height=0.24\linewidth,
  xtick={0,10,20,30,40},
  tick label style={font=\fontsize{4}{4}\selectfont},
  label style={font=\tiny},
  title style={font=\tiny, yshift=-1ex},
]

  \nextgroupplot[
    title={\tiny MMLU},
    ylabel={\tiny $\Delta$ Acc.},
    xlabel={},
    axis x line*=bottom, axis y line*=left,
    xmin=0, xmax=40, ymin=-3.5, ymax=0.5,
    grid, grid style={line width=.2pt, draw=gray!30},
  ]
  \draw[gray,dashed,line width=0.5pt](axis cs:0,0)--(axis cs:40,0);
  \addplot[name path=ummluA, draw=none, forget plot]
    table[x=epoch, y expr=\thisrow{cfg0d}+\thisrow{cfg0s}, col sep=comma]{csv_tikz/qwen_other/mmlu.csv};
  \addplot[name path=lmmluA, draw=none, forget plot]
    table[x=epoch, y expr=\thisrow{cfg0d}-\thisrow{cfg0s}, col sep=comma]{csv_tikz/qwen_other/mmlu.csv};
  \addplot[colorA!15, draw=none, forget plot] fill between[of=ummluA and lmmluA];
  \addplot[colorA, line width=0.8pt, dashed]
    table[x=epoch, y=cfg0d, col sep=comma]{csv_tikz/qwen_other/mmlu.csv};
  \addplot[name path=ummluE, draw=none, forget plot]
    table[x=epoch, y expr=\thisrow{cfg1d}+\thisrow{cfg1s}, col sep=comma]{csv_tikz/qwen_other/mmlu.csv};
  \addplot[name path=lmmluE, draw=none, forget plot]
    table[x=epoch, y expr=\thisrow{cfg1d}-\thisrow{cfg1s}, col sep=comma]{csv_tikz/qwen_other/mmlu.csv};
  \addplot[colorE!15, draw=none, forget plot] fill between[of=ummluE and lmmluE];
  \addplot[colorE, line width=0.8pt]
    table[x=epoch, y=cfg1d, col sep=comma]{csv_tikz/qwen_other/mmlu.csv};
  \addplot[name path=ummluF, draw=none, forget plot]
    table[x=epoch, y expr=\thisrow{cfg2d}+\thisrow{cfg2s}, col sep=comma]{csv_tikz/qwen_other/mmlu.csv};
  \addplot[name path=lmmluF, draw=none, forget plot]
    table[x=epoch, y expr=\thisrow{cfg2d}-\thisrow{cfg2s}, col sep=comma]{csv_tikz/qwen_other/mmlu.csv};
  \addplot[colorF!15, draw=none, forget plot] fill between[of=ummluF and lmmluF];
  \addplot[colorF, line width=0.8pt]
    table[x=epoch, y=cfg2d, col sep=comma]{csv_tikz/qwen_other/mmlu.csv};
  \addplot[name path=ummluD, draw=none, forget plot]
    table[x=epoch, y expr=\thisrow{cfg3d}+\thisrow{cfg3s}, col sep=comma]{csv_tikz/qwen_other/mmlu.csv};
  \addplot[name path=lmmluD, draw=none, forget plot]
    table[x=epoch, y expr=\thisrow{cfg3d}-\thisrow{cfg3s}, col sep=comma]{csv_tikz/qwen_other/mmlu.csv};
  \addplot[colorD!15, draw=none, forget plot] fill between[of=ummluD and lmmluD];
  \addplot[colorD, line width=0.8pt]
    table[x=epoch, y=cfg3d, col sep=comma]{csv_tikz/qwen_other/mmlu.csv};
  \addplot[name path=ummluB, draw=none, forget plot]
    table[x=epoch, y expr=\thisrow{cfg4d}+\thisrow{cfg4s}, col sep=comma]{csv_tikz/qwen_other/mmlu.csv};
  \addplot[name path=lmmluB, draw=none, forget plot]
    table[x=epoch, y expr=\thisrow{cfg4d}-\thisrow{cfg4s}, col sep=comma]{csv_tikz/qwen_other/mmlu.csv};
  \addplot[colorB!15, draw=none, forget plot] fill between[of=ummluB and lmmluB];
  \addplot[colorB, line width=0.8pt]
    table[x=epoch, y=cfg4d, col sep=comma]{csv_tikz/qwen_other/mmlu.csv};
  \addplot[name path=ummluC, draw=none, forget plot]
    table[x=epoch, y expr=\thisrow{cfg5d}+\thisrow{cfg5s}, col sep=comma]{csv_tikz/qwen_other/mmlu.csv};
  \addplot[name path=lmmluC, draw=none, forget plot]
    table[x=epoch, y expr=\thisrow{cfg5d}-\thisrow{cfg5s}, col sep=comma]{csv_tikz/qwen_other/mmlu.csv};
  \addplot[colorC!15, draw=none, forget plot] fill between[of=ummluC and lmmluC];
  \addplot[colorC, line width=0.8pt]
    table[x=epoch, y=cfg5d, col sep=comma]{csv_tikz/qwen_other/mmlu.csv};

  \nextgroupplot[
    title={\tiny ARC Challenge},
    ylabel={},
    xlabel={},
    axis x line*=bottom, axis y line*=left,
    xmin=0, xmax=40, ymin=-1.5, ymax=4.0,
    grid, grid style={line width=.2pt, draw=gray!30},
  ]
  \draw[gray,dashed,line width=0.5pt](axis cs:0,0)--(axis cs:40,0);
  \addplot[name path=uarcchA, draw=none, forget plot]
    table[x=epoch, y expr=\thisrow{cfg0d}+\thisrow{cfg0s}, col sep=comma]{csv_tikz/qwen_other/arc_challenge.csv};
  \addplot[name path=larcchA, draw=none, forget plot]
    table[x=epoch, y expr=\thisrow{cfg0d}-\thisrow{cfg0s}, col sep=comma]{csv_tikz/qwen_other/arc_challenge.csv};
  \addplot[colorA!15, draw=none, forget plot] fill between[of=uarcchA and larcchA];
  \addplot[colorA, line width=0.8pt, dashed]
    table[x=epoch, y=cfg0d, col sep=comma]{csv_tikz/qwen_other/arc_challenge.csv};
  \addplot[name path=uarcchE, draw=none, forget plot]
    table[x=epoch, y expr=\thisrow{cfg1d}+\thisrow{cfg1s}, col sep=comma]{csv_tikz/qwen_other/arc_challenge.csv};
  \addplot[name path=larcchE, draw=none, forget plot]
    table[x=epoch, y expr=\thisrow{cfg1d}-\thisrow{cfg1s}, col sep=comma]{csv_tikz/qwen_other/arc_challenge.csv};
  \addplot[colorE!15, draw=none, forget plot] fill between[of=uarcchE and larcchE];
  \addplot[colorE, line width=0.8pt]
    table[x=epoch, y=cfg1d, col sep=comma]{csv_tikz/qwen_other/arc_challenge.csv};
  \addplot[name path=uarcchF, draw=none, forget plot]
    table[x=epoch, y expr=\thisrow{cfg2d}+\thisrow{cfg2s}, col sep=comma]{csv_tikz/qwen_other/arc_challenge.csv};
  \addplot[name path=larcchF, draw=none, forget plot]
    table[x=epoch, y expr=\thisrow{cfg2d}-\thisrow{cfg2s}, col sep=comma]{csv_tikz/qwen_other/arc_challenge.csv};
  \addplot[colorF!15, draw=none, forget plot] fill between[of=uarcchF and larcchF];
  \addplot[colorF, line width=0.8pt]
    table[x=epoch, y=cfg2d, col sep=comma]{csv_tikz/qwen_other/arc_challenge.csv};
  \addplot[name path=uarcchD, draw=none, forget plot]
    table[x=epoch, y expr=\thisrow{cfg3d}+\thisrow{cfg3s}, col sep=comma]{csv_tikz/qwen_other/arc_challenge.csv};
  \addplot[name path=larcchD, draw=none, forget plot]
    table[x=epoch, y expr=\thisrow{cfg3d}-\thisrow{cfg3s}, col sep=comma]{csv_tikz/qwen_other/arc_challenge.csv};
  \addplot[colorD!15, draw=none, forget plot] fill between[of=uarcchD and larcchD];
  \addplot[colorD, line width=0.8pt]
    table[x=epoch, y=cfg3d, col sep=comma]{csv_tikz/qwen_other/arc_challenge.csv};
  \addplot[name path=uarcchB, draw=none, forget plot]
    table[x=epoch, y expr=\thisrow{cfg4d}+\thisrow{cfg4s}, col sep=comma]{csv_tikz/qwen_other/arc_challenge.csv};
  \addplot[name path=larcchB, draw=none, forget plot]
    table[x=epoch, y expr=\thisrow{cfg4d}-\thisrow{cfg4s}, col sep=comma]{csv_tikz/qwen_other/arc_challenge.csv};
  \addplot[colorB!15, draw=none, forget plot] fill between[of=uarcchB and larcchB];
  \addplot[colorB, line width=0.8pt]
    table[x=epoch, y=cfg4d, col sep=comma]{csv_tikz/qwen_other/arc_challenge.csv};
  \addplot[name path=uarcchC, draw=none, forget plot]
    table[x=epoch, y expr=\thisrow{cfg5d}+\thisrow{cfg5s}, col sep=comma]{csv_tikz/qwen_other/arc_challenge.csv};
  \addplot[name path=larcchC, draw=none, forget plot]
    table[x=epoch, y expr=\thisrow{cfg5d}-\thisrow{cfg5s}, col sep=comma]{csv_tikz/qwen_other/arc_challenge.csv};
  \addplot[colorC!15, draw=none, forget plot] fill between[of=uarcchC and larcchC];
  \addplot[colorC, line width=0.8pt]
    table[x=epoch, y=cfg5d, col sep=comma]{csv_tikz/qwen_other/arc_challenge.csv};

  \nextgroupplot[
    title={\tiny GSM8K},
    ylabel={},
    xlabel={},
    axis x line*=bottom, axis y line*=left,
    xmin=0, xmax=40, ymin=-5.0, ymax=20.0,
    grid, grid style={line width=.2pt, draw=gray!30},
  ]
  \draw[gray,dashed,line width=0.5pt](axis cs:0,0)--(axis cs:40,0);
  \addplot[name path=ugsm8kA, draw=none, forget plot]
    table[x=epoch, y expr=\thisrow{cfg0d}+\thisrow{cfg0s}, col sep=comma]{csv_tikz/qwen_other/gsm8k.csv};
  \addplot[name path=lgsm8kA, draw=none, forget plot]
    table[x=epoch, y expr=\thisrow{cfg0d}-\thisrow{cfg0s}, col sep=comma]{csv_tikz/qwen_other/gsm8k.csv};
  \addplot[colorA!15, draw=none, forget plot] fill between[of=ugsm8kA and lgsm8kA];
  \addplot[colorA, line width=0.8pt, dashed]
    table[x=epoch, y=cfg0d, col sep=comma]{csv_tikz/qwen_other/gsm8k.csv};
  \addplot[name path=ugsm8kE, draw=none, forget plot]
    table[x=epoch, y expr=\thisrow{cfg1d}+\thisrow{cfg1s}, col sep=comma]{csv_tikz/qwen_other/gsm8k.csv};
  \addplot[name path=lgsm8kE, draw=none, forget plot]
    table[x=epoch, y expr=\thisrow{cfg1d}-\thisrow{cfg1s}, col sep=comma]{csv_tikz/qwen_other/gsm8k.csv};
  \addplot[colorE!15, draw=none, forget plot] fill between[of=ugsm8kE and lgsm8kE];
  \addplot[colorE, line width=0.8pt]
    table[x=epoch, y=cfg1d, col sep=comma]{csv_tikz/qwen_other/gsm8k.csv};
  \addplot[name path=ugsm8kF, draw=none, forget plot]
    table[x=epoch, y expr=\thisrow{cfg2d}+\thisrow{cfg2s}, col sep=comma]{csv_tikz/qwen_other/gsm8k.csv};
  \addplot[name path=lgsm8kF, draw=none, forget plot]
    table[x=epoch, y expr=\thisrow{cfg2d}-\thisrow{cfg2s}, col sep=comma]{csv_tikz/qwen_other/gsm8k.csv};
  \addplot[colorF!15, draw=none, forget plot] fill between[of=ugsm8kF and lgsm8kF];
  \addplot[colorF, line width=0.8pt]
    table[x=epoch, y=cfg2d, col sep=comma]{csv_tikz/qwen_other/gsm8k.csv};
  \addplot[name path=ugsm8kD, draw=none, forget plot]
    table[x=epoch, y expr=\thisrow{cfg3d}+\thisrow{cfg3s}, col sep=comma]{csv_tikz/qwen_other/gsm8k.csv};
  \addplot[name path=lgsm8kD, draw=none, forget plot]
    table[x=epoch, y expr=\thisrow{cfg3d}-\thisrow{cfg3s}, col sep=comma]{csv_tikz/qwen_other/gsm8k.csv};
  \addplot[colorD!15, draw=none, forget plot] fill between[of=ugsm8kD and lgsm8kD];
  \addplot[colorD, line width=0.8pt]
    table[x=epoch, y=cfg3d, col sep=comma]{csv_tikz/qwen_other/gsm8k.csv};
  \addplot[name path=ugsm8kB, draw=none, forget plot]
    table[x=epoch, y expr=\thisrow{cfg4d}+\thisrow{cfg4s}, col sep=comma]{csv_tikz/qwen_other/gsm8k.csv};
  \addplot[name path=lgsm8kB, draw=none, forget plot]
    table[x=epoch, y expr=\thisrow{cfg4d}-\thisrow{cfg4s}, col sep=comma]{csv_tikz/qwen_other/gsm8k.csv};
  \addplot[colorB!15, draw=none, forget plot] fill between[of=ugsm8kB and lgsm8kB];
  \addplot[colorB, line width=0.8pt]
    table[x=epoch, y=cfg4d, col sep=comma]{csv_tikz/qwen_other/gsm8k.csv};
  \addplot[name path=ugsm8kC, draw=none, forget plot]
    table[x=epoch, y expr=\thisrow{cfg5d}+\thisrow{cfg5s}, col sep=comma]{csv_tikz/qwen_other/gsm8k.csv};
  \addplot[name path=lgsm8kC, draw=none, forget plot]
    table[x=epoch, y expr=\thisrow{cfg5d}-\thisrow{cfg5s}, col sep=comma]{csv_tikz/qwen_other/gsm8k.csv};
  \addplot[colorC!15, draw=none, forget plot] fill between[of=ugsm8kC and lgsm8kC];
  \addplot[colorC, line width=0.8pt]
    table[x=epoch, y=cfg5d, col sep=comma]{csv_tikz/qwen_other/gsm8k.csv};

  \nextgroupplot[
    title={\tiny HellaSwag},
    ylabel={},
    xlabel={},
    axis x line*=bottom, axis y line*=left,
    xmin=0, xmax=40, ymin=-2.5, ymax=2.5,
    grid, grid style={line width=.2pt, draw=gray!30},
  ]
  \draw[gray,dashed,line width=0.5pt](axis cs:0,0)--(axis cs:40,0);
  \addplot[name path=uhellaA, draw=none, forget plot]
    table[x=epoch, y expr=\thisrow{cfg0d}+\thisrow{cfg0s}, col sep=comma]{csv_tikz/qwen_other/hellaswag.csv};
  \addplot[name path=lhellaA, draw=none, forget plot]
    table[x=epoch, y expr=\thisrow{cfg0d}-\thisrow{cfg0s}, col sep=comma]{csv_tikz/qwen_other/hellaswag.csv};
  \addplot[colorA!15, draw=none, forget plot] fill between[of=uhellaA and lhellaA];
  \addplot[colorA, line width=0.8pt, dashed]
    table[x=epoch, y=cfg0d, col sep=comma]{csv_tikz/qwen_other/hellaswag.csv};
  \addplot[name path=uhellaE, draw=none, forget plot]
    table[x=epoch, y expr=\thisrow{cfg1d}+\thisrow{cfg1s}, col sep=comma]{csv_tikz/qwen_other/hellaswag.csv};
  \addplot[name path=lhellaE, draw=none, forget plot]
    table[x=epoch, y expr=\thisrow{cfg1d}-\thisrow{cfg1s}, col sep=comma]{csv_tikz/qwen_other/hellaswag.csv};
  \addplot[colorE!15, draw=none, forget plot] fill between[of=uhellaE and lhellaE];
  \addplot[colorE, line width=0.8pt]
    table[x=epoch, y=cfg1d, col sep=comma]{csv_tikz/qwen_other/hellaswag.csv};
  \addplot[name path=uhellaF, draw=none, forget plot]
    table[x=epoch, y expr=\thisrow{cfg2d}+\thisrow{cfg2s}, col sep=comma]{csv_tikz/qwen_other/hellaswag.csv};
  \addplot[name path=lhellaF, draw=none, forget plot]
    table[x=epoch, y expr=\thisrow{cfg2d}-\thisrow{cfg2s}, col sep=comma]{csv_tikz/qwen_other/hellaswag.csv};
  \addplot[colorF!15, draw=none, forget plot] fill between[of=uhellaF and lhellaF];
  \addplot[colorF, line width=0.8pt]
    table[x=epoch, y=cfg2d, col sep=comma]{csv_tikz/qwen_other/hellaswag.csv};
  \addplot[name path=uhellaD, draw=none, forget plot]
    table[x=epoch, y expr=\thisrow{cfg3d}+\thisrow{cfg3s}, col sep=comma]{csv_tikz/qwen_other/hellaswag.csv};
  \addplot[name path=lhellaD, draw=none, forget plot]
    table[x=epoch, y expr=\thisrow{cfg3d}-\thisrow{cfg3s}, col sep=comma]{csv_tikz/qwen_other/hellaswag.csv};
  \addplot[colorD!15, draw=none, forget plot] fill between[of=uhellaD and lhellaD];
  \addplot[colorD, line width=0.8pt]
    table[x=epoch, y=cfg3d, col sep=comma]{csv_tikz/qwen_other/hellaswag.csv};
  \addplot[name path=uhellaB, draw=none, forget plot]
    table[x=epoch, y expr=\thisrow{cfg4d}+\thisrow{cfg4s}, col sep=comma]{csv_tikz/qwen_other/hellaswag.csv};
  \addplot[name path=lhellaB, draw=none, forget plot]
    table[x=epoch, y expr=\thisrow{cfg4d}-\thisrow{cfg4s}, col sep=comma]{csv_tikz/qwen_other/hellaswag.csv};
  \addplot[colorB!15, draw=none, forget plot] fill between[of=uhellaB and lhellaB];
  \addplot[colorB, line width=0.8pt]
    table[x=epoch, y=cfg4d, col sep=comma]{csv_tikz/qwen_other/hellaswag.csv};
  \addplot[name path=uhellaC, draw=none, forget plot]
    table[x=epoch, y expr=\thisrow{cfg5d}+\thisrow{cfg5s}, col sep=comma]{csv_tikz/qwen_other/hellaswag.csv};
  \addplot[name path=lhellaC, draw=none, forget plot]
    table[x=epoch, y expr=\thisrow{cfg5d}-\thisrow{cfg5s}, col sep=comma]{csv_tikz/qwen_other/hellaswag.csv};
  \addplot[colorC!15, draw=none, forget plot] fill between[of=uhellaC and lhellaC];
  \addplot[colorC, line width=0.8pt]
    table[x=epoch, y=cfg5d, col sep=comma]{csv_tikz/qwen_other/hellaswag.csv};

  \nextgroupplot[
    title={\tiny TruthfulQA},
    ylabel={\tiny $\Delta$ Acc.},
    xlabel={\tiny Epoch},
    axis x line*=bottom, axis y line*=left,
    xmin=0, xmax=40, ymin=-3.0, ymax=1.0,
    grid, grid style={line width=.2pt, draw=gray!30},
  ]
  \draw[gray,dashed,line width=0.5pt](axis cs:0,0)--(axis cs:40,0);
  \addplot[name path=utruthA, draw=none, forget plot]
    table[x=epoch, y expr=\thisrow{cfg0d}+\thisrow{cfg0s}, col sep=comma]{csv_tikz/qwen_other/truthfulqa_mc2.csv};
  \addplot[name path=ltruthA, draw=none, forget plot]
    table[x=epoch, y expr=\thisrow{cfg0d}-\thisrow{cfg0s}, col sep=comma]{csv_tikz/qwen_other/truthfulqa_mc2.csv};
  \addplot[colorA!15, draw=none, forget plot] fill between[of=utruthA and ltruthA];
  \addplot[colorA, line width=0.8pt, dashed]
    table[x=epoch, y=cfg0d, col sep=comma]{csv_tikz/qwen_other/truthfulqa_mc2.csv};
  \addplot[name path=utruthE, draw=none, forget plot]
    table[x=epoch, y expr=\thisrow{cfg1d}+\thisrow{cfg1s}, col sep=comma]{csv_tikz/qwen_other/truthfulqa_mc2.csv};
  \addplot[name path=ltruthE, draw=none, forget plot]
    table[x=epoch, y expr=\thisrow{cfg1d}-\thisrow{cfg1s}, col sep=comma]{csv_tikz/qwen_other/truthfulqa_mc2.csv};
  \addplot[colorE!15, draw=none, forget plot] fill between[of=utruthE and ltruthE];
  \addplot[colorE, line width=0.8pt]
    table[x=epoch, y=cfg1d, col sep=comma]{csv_tikz/qwen_other/truthfulqa_mc2.csv};
  \addplot[name path=utruthF, draw=none, forget plot]
    table[x=epoch, y expr=\thisrow{cfg2d}+\thisrow{cfg2s}, col sep=comma]{csv_tikz/qwen_other/truthfulqa_mc2.csv};
  \addplot[name path=ltruthF, draw=none, forget plot]
    table[x=epoch, y expr=\thisrow{cfg2d}-\thisrow{cfg2s}, col sep=comma]{csv_tikz/qwen_other/truthfulqa_mc2.csv};
  \addplot[colorF!15, draw=none, forget plot] fill between[of=utruthF and ltruthF];
  \addplot[colorF, line width=0.8pt]
    table[x=epoch, y=cfg2d, col sep=comma]{csv_tikz/qwen_other/truthfulqa_mc2.csv};
  \addplot[name path=utruthD, draw=none, forget plot]
    table[x=epoch, y expr=\thisrow{cfg3d}+\thisrow{cfg3s}, col sep=comma]{csv_tikz/qwen_other/truthfulqa_mc2.csv};
  \addplot[name path=ltruthD, draw=none, forget plot]
    table[x=epoch, y expr=\thisrow{cfg3d}-\thisrow{cfg3s}, col sep=comma]{csv_tikz/qwen_other/truthfulqa_mc2.csv};
  \addplot[colorD!15, draw=none, forget plot] fill between[of=utruthD and ltruthD];
  \addplot[colorD, line width=0.8pt]
    table[x=epoch, y=cfg3d, col sep=comma]{csv_tikz/qwen_other/truthfulqa_mc2.csv};
  \addplot[name path=utruthB, draw=none, forget plot]
    table[x=epoch, y expr=\thisrow{cfg4d}+\thisrow{cfg4s}, col sep=comma]{csv_tikz/qwen_other/truthfulqa_mc2.csv};
  \addplot[name path=ltruthB, draw=none, forget plot]
    table[x=epoch, y expr=\thisrow{cfg4d}-\thisrow{cfg4s}, col sep=comma]{csv_tikz/qwen_other/truthfulqa_mc2.csv};
  \addplot[colorB!15, draw=none, forget plot] fill between[of=utruthB and ltruthB];
  \addplot[colorB, line width=0.8pt]
    table[x=epoch, y=cfg4d, col sep=comma]{csv_tikz/qwen_other/truthfulqa_mc2.csv};
  \addplot[name path=utruthC, draw=none, forget plot]
    table[x=epoch, y expr=\thisrow{cfg5d}+\thisrow{cfg5s}, col sep=comma]{csv_tikz/qwen_other/truthfulqa_mc2.csv};
  \addplot[name path=ltruthC, draw=none, forget plot]
    table[x=epoch, y expr=\thisrow{cfg5d}-\thisrow{cfg5s}, col sep=comma]{csv_tikz/qwen_other/truthfulqa_mc2.csv};
  \addplot[colorC!15, draw=none, forget plot] fill between[of=utruthC and ltruthC];
  \addplot[colorC, line width=0.8pt]
    table[x=epoch, y=cfg5d, col sep=comma]{csv_tikz/qwen_other/truthfulqa_mc2.csv};

  \nextgroupplot[
    title={\tiny Math},
    ylabel={},
    xlabel={\tiny Epoch},
    axis x line*=bottom, axis y line*=left,
    xmin=0, xmax=40, ymin=-8.0, ymax=1.0,
    grid, grid style={line width=.2pt, draw=gray!30},
  ]
  \draw[gray,dashed,line width=0.5pt](axis cs:0,0)--(axis cs:40,0);
  \addplot[name path=uminerA, draw=none, forget plot]
    table[x=epoch, y expr=\thisrow{cfg0d}+\thisrow{cfg0s}, col sep=comma]{csv_tikz/qwen_other/minerva_math.csv};
  \addplot[name path=lminerA, draw=none, forget plot]
    table[x=epoch, y expr=\thisrow{cfg0d}-\thisrow{cfg0s}, col sep=comma]{csv_tikz/qwen_other/minerva_math.csv};
  \addplot[colorA!15, draw=none, forget plot] fill between[of=uminerA and lminerA];
  \addplot[colorA, line width=0.8pt, dashed]
    table[x=epoch, y=cfg0d, col sep=comma]{csv_tikz/qwen_other/minerva_math.csv};
  \addplot[name path=uminerE, draw=none, forget plot]
    table[x=epoch, y expr=\thisrow{cfg1d}+\thisrow{cfg1s}, col sep=comma]{csv_tikz/qwen_other/minerva_math.csv};
  \addplot[name path=lminerE, draw=none, forget plot]
    table[x=epoch, y expr=\thisrow{cfg1d}-\thisrow{cfg1s}, col sep=comma]{csv_tikz/qwen_other/minerva_math.csv};
  \addplot[colorE!15, draw=none, forget plot] fill between[of=uminerE and lminerE];
  \addplot[colorE, line width=0.8pt]
    table[x=epoch, y=cfg1d, col sep=comma]{csv_tikz/qwen_other/minerva_math.csv};
  \addplot[name path=uminerF, draw=none, forget plot]
    table[x=epoch, y expr=\thisrow{cfg2d}+\thisrow{cfg2s}, col sep=comma]{csv_tikz/qwen_other/minerva_math.csv};
  \addplot[name path=lminerF, draw=none, forget plot]
    table[x=epoch, y expr=\thisrow{cfg2d}-\thisrow{cfg2s}, col sep=comma]{csv_tikz/qwen_other/minerva_math.csv};
  \addplot[colorF!15, draw=none, forget plot] fill between[of=uminerF and lminerF];
  \addplot[colorF, line width=0.8pt]
    table[x=epoch, y=cfg2d, col sep=comma]{csv_tikz/qwen_other/minerva_math.csv};
  \addplot[name path=uminerD, draw=none, forget plot]
    table[x=epoch, y expr=\thisrow{cfg3d}+\thisrow{cfg3s}, col sep=comma]{csv_tikz/qwen_other/minerva_math.csv};
  \addplot[name path=lminerD, draw=none, forget plot]
    table[x=epoch, y expr=\thisrow{cfg3d}-\thisrow{cfg3s}, col sep=comma]{csv_tikz/qwen_other/minerva_math.csv};
  \addplot[colorD!15, draw=none, forget plot] fill between[of=uminerD and lminerD];
  \addplot[colorD, line width=0.8pt]
    table[x=epoch, y=cfg3d, col sep=comma]{csv_tikz/qwen_other/minerva_math.csv};
  \addplot[name path=uminerB, draw=none, forget plot]
    table[x=epoch, y expr=\thisrow{cfg4d}+\thisrow{cfg4s}, col sep=comma]{csv_tikz/qwen_other/minerva_math.csv};
  \addplot[name path=lminerB, draw=none, forget plot]
    table[x=epoch, y expr=\thisrow{cfg4d}-\thisrow{cfg4s}, col sep=comma]{csv_tikz/qwen_other/minerva_math.csv};
  \addplot[colorB!15, draw=none, forget plot] fill between[of=uminerB and lminerB];
  \addplot[colorB, line width=0.8pt]
    table[x=epoch, y=cfg4d, col sep=comma]{csv_tikz/qwen_other/minerva_math.csv};
  \addplot[name path=uminerC, draw=none, forget plot]
    table[x=epoch, y expr=\thisrow{cfg5d}+\thisrow{cfg5s}, col sep=comma]{csv_tikz/qwen_other/minerva_math.csv};
  \addplot[name path=lminerC, draw=none, forget plot]
    table[x=epoch, y expr=\thisrow{cfg5d}-\thisrow{cfg5s}, col sep=comma]{csv_tikz/qwen_other/minerva_math.csv};
  \addplot[colorC!15, draw=none, forget plot] fill between[of=uminerC and lminerC];
  \addplot[colorC, line width=0.8pt]
    table[x=epoch, y=cfg5d, col sep=comma]{csv_tikz/qwen_other/minerva_math.csv};

  \nextgroupplot[
    title={\tiny HumanEval},
    ylabel={},
    xlabel={\tiny Epoch},
    axis x line*=bottom, axis y line*=left,
    xmin=0, xmax=40, ymin=-9.0, ymax=3.0,
    grid, grid style={line width=.2pt, draw=gray!30},
    legend to name=figlegend,
    legend style={draw=gray!60, fill=white,
      nodes={scale=0.50, transform shape},
      text=black, cells={align=left}, row sep=0pt,
      inner sep=3pt, rounded corners=2pt},
  ]
  \draw[gray,dashed,line width=0.5pt](axis cs:0,0)--(axis cs:40,0);
  \addplot[name path=uhumanA, draw=none, forget plot]
    table[x=epoch, y expr=\thisrow{cfg0d}+\thisrow{cfg0s}, col sep=comma]{csv_tikz/qwen_other/humaneval_1.csv};
  \addplot[name path=lhumanA, draw=none, forget plot]
    table[x=epoch, y expr=\thisrow{cfg0d}-\thisrow{cfg0s}, col sep=comma]{csv_tikz/qwen_other/humaneval_1.csv};
  \addplot[colorA!15, draw=none, forget plot] fill between[of=uhumanA and lhumanA];
  \addplot[colorA, line width=0.8pt, dashed]
    table[x=epoch, y=cfg0d, col sep=comma]{csv_tikz/qwen_other/humaneval_1.csv};
  \addlegendentry{Qwen2.5-0.5B $\tau{=}1.25$};
  \addplot[name path=uhumanE, draw=none, forget plot]
    table[x=epoch, y expr=\thisrow{cfg1d}+\thisrow{cfg1s}, col sep=comma]{csv_tikz/qwen_other/humaneval_1.csv};
  \addplot[name path=lhumanE, draw=none, forget plot]
    table[x=epoch, y expr=\thisrow{cfg1d}-\thisrow{cfg1s}, col sep=comma]{csv_tikz/qwen_other/humaneval_1.csv};
  \addplot[colorE!15, draw=none, forget plot] fill between[of=uhumanE and lhumanE];
  \addplot[colorE, line width=0.8pt]
    table[x=epoch, y=cfg1d, col sep=comma]{csv_tikz/qwen_other/humaneval_1.csv};
  \addlegendentry{Qwen2.5-7B $\tau{=}1.25$};
  \addplot[name path=uhumanF, draw=none, forget plot]
    table[x=epoch, y expr=\thisrow{cfg2d}+\thisrow{cfg2s}, col sep=comma]{csv_tikz/qwen_other/humaneval_1.csv};
  \addplot[name path=lhumanF, draw=none, forget plot]
    table[x=epoch, y expr=\thisrow{cfg2d}-\thisrow{cfg2s}, col sep=comma]{csv_tikz/qwen_other/humaneval_1.csv};
  \addplot[colorF!15, draw=none, forget plot] fill between[of=uhumanF and lhumanF];
  \addplot[colorF, line width=0.8pt]
    table[x=epoch, y=cfg2d, col sep=comma]{csv_tikz/qwen_other/humaneval_1.csv};
  \addlegendentry{Qwen3-8B $\tau{=}1.25$};
  \addplot[name path=uhumanD, draw=none, forget plot]
    table[x=epoch, y expr=\thisrow{cfg3d}+\thisrow{cfg3s}, col sep=comma]{csv_tikz/qwen_other/humaneval_1.csv};
  \addplot[name path=lhumanD, draw=none, forget plot]
    table[x=epoch, y expr=\thisrow{cfg3d}-\thisrow{cfg3s}, col sep=comma]{csv_tikz/qwen_other/humaneval_1.csv};
  \addplot[colorD!15, draw=none, forget plot] fill between[of=uhumanD and lhumanD];
  \addplot[colorD, line width=0.8pt]
    table[x=epoch, y=cfg3d, col sep=comma]{csv_tikz/qwen_other/humaneval_1.csv};
  \addlegendentry{Llama-3.1-1B $\tau{=}1.25$};
  \addplot[name path=uhumanB, draw=none, forget plot]
    table[x=epoch, y expr=\thisrow{cfg4d}+\thisrow{cfg4s}, col sep=comma]{csv_tikz/qwen_other/humaneval_1.csv};
  \addplot[name path=lhumanB, draw=none, forget plot]
    table[x=epoch, y expr=\thisrow{cfg4d}-\thisrow{cfg4s}, col sep=comma]{csv_tikz/qwen_other/humaneval_1.csv};
  \addplot[colorB!15, draw=none, forget plot] fill between[of=uhumanB and lhumanB];
  \addplot[colorB, line width=0.8pt]
    table[x=epoch, y=cfg4d, col sep=comma]{csv_tikz/qwen_other/humaneval_1.csv};
  \addlegendentry{Llama-3.1-1B $\tau{=}1.0$};
  \addplot[name path=uhumanC, draw=none, forget plot]
    table[x=epoch, y expr=\thisrow{cfg5d}+\thisrow{cfg5s}, col sep=comma]{csv_tikz/qwen_other/humaneval_1.csv};
  \addplot[name path=lhumanC, draw=none, forget plot]
    table[x=epoch, y expr=\thisrow{cfg5d}-\thisrow{cfg5s}, col sep=comma]{csv_tikz/qwen_other/humaneval_1.csv};
  \addplot[colorC!15, draw=none, forget plot] fill between[of=uhumanC and lhumanC];
  \addplot[colorC, line width=0.8pt]
    table[x=epoch, y=cfg5d, col sep=comma]{csv_tikz/qwen_other/humaneval_1.csv};
  \addlegendentry{Llama-3.1-1B $\tau{=}0.75$};

  \nextgroupplot[axis lines=none, xtick=\empty, ytick=\empty,
    xmin=0, xmax=1, ymin=0, ymax=1, xlabel={}, ylabel={},]
\end{groupplot}
\node[anchor=center] at (group c4r2.center) {\pgfplotslegendfromname{figlegend}};
\end{tikzpicture}
}
\caption{\small \textbf{Synthetic utility is relational: self-generated data is strongest, same-lineage transfer outperforms a larger but differently trained model, and cross-family transfer is weakest.}
$\Delta$ performance of the Qwen2.5-0.5B student trained on synthetic data from four source models over 40 epochs. Dashed lines: self-generated (Qwen2.5-0.5B). Qwen3-8B is larger and more capable than Qwen2.5-7B, yet transfers worse, indicating that pretraining lineage predicts transfer better than source capability. Shaded bands show $\pm 1$ std.}
\label{fig:qwen_other}
\end{figure}

%% file: figure_nll.tex
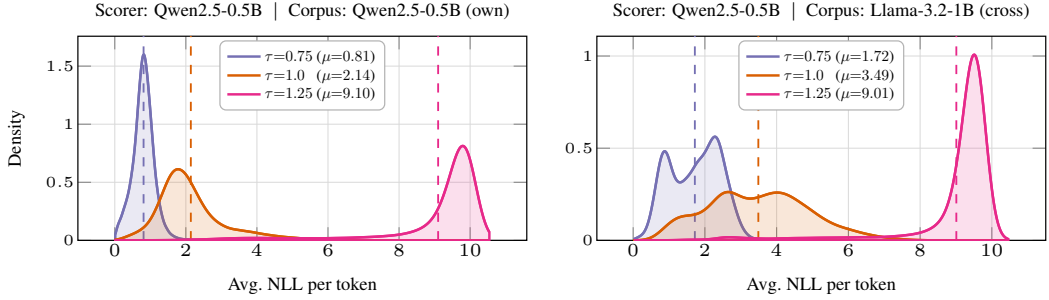
\begin{figure}[t]
\centering
\resizebox{\textwidth}{!}{%
\begin{tikzpicture}
\begin{groupplot}[
  group style={group size=2 by 1, horizontal sep=8mm},
  width=0.48\linewidth, height=0.28\linewidth,
  label style={font=\tiny},
  title style={font=\tiny, yshift=-1ex},
  tick label style={font=\fontsize{5}{5}\selectfont},
  ymin=0,
  grid, grid style={line width=.2pt, draw=gray!30},
  xlabel={\tiny Avg.\ NLL per token},
  scaled y ticks=false,
  yticklabel style={/pgf/number format/fixed, font=\fontsize{5}{5}\selectfont},
]

\nextgroupplot[
  title={\tiny Scorer: Qwen2.5-0.5B \ $|$ \ Corpus: Qwen2.5-0.5B (own)},
  ylabel={\tiny Density},
  legend style={at={(0.50,0.98)}, anchor=north,
    nodes={scale=0.50, transform shape}, draw=gray!60, fill=white,
    cells={align=left}, row sep=0pt, inner sep=2pt, rounded corners=2pt},
]
\addplot[colorC, fill=colorC, fill opacity=0.15, line width=0.8pt, smooth]
  table[x=x, y=t075, col sep=comma]{csv_tikz/nll/qwen_own.csv} \closedcycle;
\addplot[colorC, line width=0.8pt, smooth, forget plot]
  table[x=x, y=t075, col sep=comma]{csv_tikz/nll/qwen_own.csv};
\draw[colorC, dashed, line width=0.6pt] (axis cs:0.81,0) -- (axis cs:0.81,10);
\addplot[colorB, fill=colorB, fill opacity=0.15, line width=0.8pt, smooth]
  table[x=x, y=t10, col sep=comma]{csv_tikz/nll/qwen_own.csv} \closedcycle;
\addplot[colorB, line width=0.8pt, smooth, forget plot]
  table[x=x, y=t10, col sep=comma]{csv_tikz/nll/qwen_own.csv};
\draw[colorB, dashed, line width=0.6pt] (axis cs:2.14,0) -- (axis cs:2.14,10);
\addplot[colorD, fill=colorD, fill opacity=0.15, line width=0.8pt, smooth]
  table[x=x, y=t125, col sep=comma]{csv_tikz/nll/qwen_own.csv} \closedcycle;
\addplot[colorD, line width=0.8pt, smooth, forget plot]
  table[x=x, y=t125, col sep=comma]{csv_tikz/nll/qwen_own.csv};
\draw[colorD, dashed, line width=0.6pt] (axis cs:9.10,0) -- (axis cs:9.10,10);
\addlegendentry{$\tau{=}0.75$ ($\mu{=}0.81$)};
\addlegendentry{$\tau{=}1.0$\phantom{0} ($\mu{=}2.14$)};
\addlegendentry{$\tau{=}1.25$ ($\mu{=}9.10$)};

\nextgroupplot[
  title={\tiny Scorer: Qwen2.5-0.5B \ $|$ \ Corpus: Llama-3.2-1B (cross)},
  ylabel={},
  legend style={at={(0.50,0.98)}, anchor=north,
    nodes={scale=0.50, transform shape}, draw=gray!60, fill=white,
    cells={align=left}, row sep=0pt, inner sep=2pt, rounded corners=2pt},
]
\addplot[colorC, fill=colorC, fill opacity=0.15, line width=0.8pt, smooth]
  table[x=x, y=t075, col sep=comma]{csv_tikz/nll/qwen_cross.csv} \closedcycle;
\addplot[colorC, line width=0.8pt, smooth, forget plot]
  table[x=x, y=t075, col sep=comma]{csv_tikz/nll/qwen_cross.csv};
\draw[colorC, dashed, line width=0.6pt] (axis cs:1.72,0) -- (axis cs:1.72,10);
\addplot[colorB, fill=colorB, fill opacity=0.15, line width=0.8pt, smooth]
  table[x=x, y=t10, col sep=comma]{csv_tikz/nll/qwen_cross.csv} \closedcycle;
\addplot[colorB, line width=0.8pt, smooth, forget plot]
  table[x=x, y=t10, col sep=comma]{csv_tikz/nll/qwen_cross.csv};
\draw[colorB, dashed, line width=0.6pt] (axis cs:3.49,0) -- (axis cs:3.49,10);
\addplot[colorD, fill=colorD, fill opacity=0.15, line width=0.8pt, smooth]
  table[x=x, y=t125, col sep=comma]{csv_tikz/nll/qwen_cross.csv} \closedcycle;
\addplot[colorD, line width=0.8pt, smooth, forget plot]
  table[x=x, y=t125, col sep=comma]{csv_tikz/nll/qwen_cross.csv};
\draw[colorD, dashed, line width=0.6pt] (axis cs:9.01,0) -- (axis cs:9.01,10);
\addlegendentry{$\tau{=}0.75$ ($\mu{=}1.72$)};
\addlegendentry{$\tau{=}1.0$\phantom{0} ($\mu{=}3.49$)};
\addlegendentry{$\tau{=}1.25$ ($\mu{=}9.01$)};

\end{groupplot}
\end{tikzpicture}
}
\caption{\small \textbf{Likelihood under the student does not predict utility: two corpora with nearly identical mean NLL ($\mu{=}9.10$ vs $\mu{=}9.01$) produce opposite downstream effects.}
KDE of average NLL per token for synthetic corpora generated by Qwen2.5-0.5B (left, own) and Llama-3.2-1B (right, cross), all scored by Qwen2.5-0.5B. Dashed lines indicate distribution means. Despite converging at $\tau{=}1.25$, the own corpus substantially improves structured reasoning while the cross corpus does not (Figure~\ref{fig:qwen_other}). Symmetric distributions under the Llama scorer are reported in Appendix~\ref{app:llama_other}.}
\label{fig:nll_dist}
\end{figure}

%% file: fig_pythia_bench.tex
%
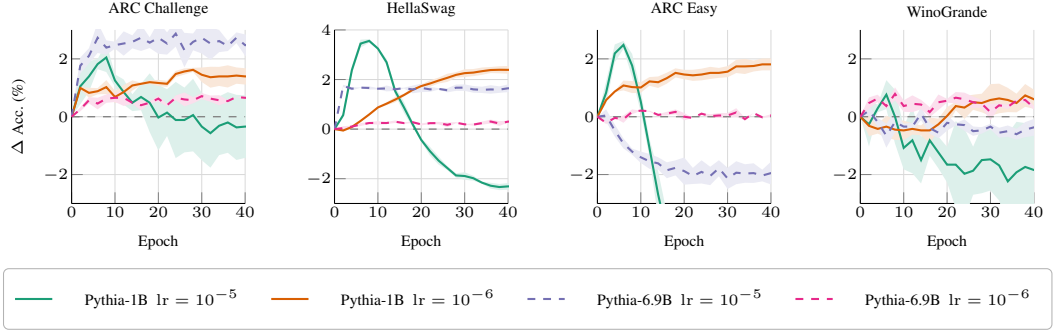
\begin{figure}[t]
\centering
\resizebox{\textwidth}{!}{%
\begin{tikzpicture}
\begin{groupplot}[
  group style={group size=4 by 1, horizontal sep=12mm},
  width=0.28\linewidth, height=0.28\linewidth,
  xtick={0,10,20,30,40},
  tick label style={font=\fontsize{4}{4}\selectfont},
  label style={font=\tiny},
  title style={font=\tiny, yshift=-1ex},
  xlabel={\tiny Epoch},
]

\nextgroupplot[
  title={\tiny ARC Challenge},
  ylabel={\tiny $\Delta$ Acc.\ (\%)},
  axis x line*=bottom, axis y line*=left,
  xmin=0, xmax=40, ymin=-3.0, ymax=3.0,
  grid, grid style={line width=.2pt, draw=gray!30},
]
\draw[gray,dashed,line width=0.5pt](axis cs:0,0)--(axis cs:40,0);
\addplot[name path=uarc0, draw=none, forget plot]
  table[x=epoch, y expr=\thisrow{cfg0d}+\thisrow{cfg0s}, col sep=comma]
  {csv_tikz/pythia1b/benchmarks/arc_challenge.csv};
\addplot[name path=larc0, draw=none, forget plot]
  table[x=epoch, y expr=\thisrow{cfg0d}-\thisrow{cfg0s}, col sep=comma]
  {csv_tikz/pythia1b/benchmarks/arc_challenge.csv};
\addplot[colorA!15, draw=none, forget plot] fill between[of=uarc0 and larc0];
\addplot[colorA, line width=0.8pt]
  table[x=epoch, y=cfg0d, col sep=comma]
  {csv_tikz/pythia1b/benchmarks/arc_challenge.csv};
\addplot[name path=uarc1, draw=none, forget plot]
  table[x=epoch, y expr=\thisrow{cfg1d}+\thisrow{cfg1s}, col sep=comma]
  {csv_tikz/pythia1b/benchmarks/arc_challenge.csv};
\addplot[name path=larc1, draw=none, forget plot]
  table[x=epoch, y expr=\thisrow{cfg1d}-\thisrow{cfg1s}, col sep=comma]
  {csv_tikz/pythia1b/benchmarks/arc_challenge.csv};
\addplot[colorB!15, draw=none, forget plot] fill between[of=uarc1 and larc1];
\addplot[colorB, line width=0.8pt]
  table[x=epoch, y=cfg1d, col sep=comma]
  {csv_tikz/pythia1b/benchmarks/arc_challenge.csv};
\addplot[name path=uarc2, draw=none, forget plot]
  table[x=epoch, y expr=\thisrow{cfg0d}+\thisrow{cfg0s}, col sep=comma]
  {csv_tikz/pythia6.9b/benchmarks/arc_challenge.csv};
\addplot[name path=larc2, draw=none, forget plot]
  table[x=epoch, y expr=\thisrow{cfg0d}-\thisrow{cfg0s}, col sep=comma]
  {csv_tikz/pythia6.9b/benchmarks/arc_challenge.csv};
\addplot[colorC!15, draw=none, forget plot] fill between[of=uarc2 and larc2];
\addplot[colorC, line width=0.8pt, dashed]
  table[x=epoch, y=cfg0d, col sep=comma]
  {csv_tikz/pythia6.9b/benchmarks/arc_challenge.csv};
\addplot[name path=uarc3, draw=none, forget plot]
  table[x=epoch, y expr=\thisrow{cfg1d}+\thisrow{cfg1s}, col sep=comma]
  {csv_tikz/pythia6.9b/benchmarks/arc_challenge.csv};
\addplot[name path=larc3, draw=none, forget plot]
  table[x=epoch, y expr=\thisrow{cfg1d}-\thisrow{cfg1s}, col sep=comma]
  {csv_tikz/pythia6.9b/benchmarks/arc_challenge.csv};
\addplot[colorD!15, draw=none, forget plot] fill between[of=uarc3 and larc3];
\addplot[colorD, line width=0.8pt, dashed]
  table[x=epoch, y=cfg1d, col sep=comma]
  {csv_tikz/pythia6.9b/benchmarks/arc_challenge.csv};

\nextgroupplot[
  title={\tiny HellaSwag},
  ylabel={},
  axis x line*=bottom, axis y line*=left,
  xmin=0, xmax=40, ymin=-3.0, ymax=4.0,
  grid, grid style={line width=.2pt, draw=gray!30},
]
\draw[gray,dashed,line width=0.5pt](axis cs:0,0)--(axis cs:40,0);
\addplot[name path=uhel0, draw=none, forget plot]
  table[x=epoch, y expr=\thisrow{cfg0d}+\thisrow{cfg0s}, col sep=comma]
  {csv_tikz/pythia1b/benchmarks/hellaswag.csv};
\addplot[name path=lhel0, draw=none, forget plot]
  table[x=epoch, y expr=\thisrow{cfg0d}-\thisrow{cfg0s}, col sep=comma]
  {csv_tikz/pythia1b/benchmarks/hellaswag.csv};
\addplot[colorA!15, draw=none, forget plot] fill between[of=uhel0 and lhel0];
\addplot[colorA, line width=0.8pt]
  table[x=epoch, y=cfg0d, col sep=comma]
  {csv_tikz/pythia1b/benchmarks/hellaswag.csv};
\addplot[name path=uhel1, draw=none, forget plot]
  table[x=epoch, y expr=\thisrow{cfg1d}+\thisrow{cfg1s}, col sep=comma]
  {csv_tikz/pythia1b/benchmarks/hellaswag.csv};
\addplot[name path=lhel1, draw=none, forget plot]
  table[x=epoch, y expr=\thisrow{cfg1d}-\thisrow{cfg1s}, col sep=comma]
  {csv_tikz/pythia1b/benchmarks/hellaswag.csv};
\addplot[colorB!15, draw=none, forget plot] fill between[of=uhel1 and lhel1];
\addplot[colorB, line width=0.8pt]
  table[x=epoch, y=cfg1d, col sep=comma]
  {csv_tikz/pythia1b/benchmarks/hellaswag.csv};
\addplot[name path=uhel2, draw=none, forget plot]
  table[x=epoch, y expr=\thisrow{cfg0d}+\thisrow{cfg0s}, col sep=comma]
  {csv_tikz/pythia6.9b/benchmarks/hellaswag.csv};
\addplot[name path=lhel2, draw=none, forget plot]
  table[x=epoch, y expr=\thisrow{cfg0d}-\thisrow{cfg0s}, col sep=comma]
  {csv_tikz/pythia6.9b/benchmarks/hellaswag.csv};
\addplot[colorC!15, draw=none, forget plot] fill between[of=uhel2 and lhel2];
\addplot[colorC, line width=0.8pt, dashed]
  table[x=epoch, y=cfg0d, col sep=comma]
  {csv_tikz/pythia6.9b/benchmarks/hellaswag.csv};
\addplot[name path=uhel3, draw=none, forget plot]
  table[x=epoch, y expr=\thisrow{cfg1d}+\thisrow{cfg1s}, col sep=comma]
  {csv_tikz/pythia6.9b/benchmarks/hellaswag.csv};
\addplot[name path=lhel3, draw=none, forget plot]
  table[x=epoch, y expr=\thisrow{cfg1d}-\thisrow{cfg1s}, col sep=comma]
  {csv_tikz/pythia6.9b/benchmarks/hellaswag.csv};
\addplot[colorD!15, draw=none, forget plot] fill between[of=uhel3 and lhel3];
\addplot[colorD, line width=0.8pt, dashed]
  table[x=epoch, y=cfg1d, col sep=comma]
  {csv_tikz/pythia6.9b/benchmarks/hellaswag.csv};

\nextgroupplot[
  title={\tiny ARC Easy},
  ylabel={},
  axis x line*=bottom, axis y line*=left,
  xmin=0, xmax=40, ymin=-3.0, ymax=3.0,
  grid, grid style={line width=.2pt, draw=gray!30},
]
\draw[gray,dashed,line width=0.5pt](axis cs:0,0)--(axis cs:40,0);
\addplot[name path=uae0, draw=none, forget plot]
  table[x=epoch, y expr=\thisrow{cfg0d}+\thisrow{cfg0s}, col sep=comma]
  {csv_tikz/pythia1b/benchmarks/arc_easy.csv};
\addplot[name path=lae0, draw=none, forget plot]
  table[x=epoch, y expr=\thisrow{cfg0d}-\thisrow{cfg0s}, col sep=comma]
  {csv_tikz/pythia1b/benchmarks/arc_easy.csv};
\addplot[colorA!15, draw=none, forget plot] fill between[of=uae0 and lae0];
\addplot[colorA, line width=0.8pt]
  table[x=epoch, y=cfg0d, col sep=comma]
  {csv_tikz/pythia1b/benchmarks/arc_easy.csv};
\addplot[name path=uae1, draw=none, forget plot]
  table[x=epoch, y expr=\thisrow{cfg1d}+\thisrow{cfg1s}, col sep=comma]
  {csv_tikz/pythia1b/benchmarks/arc_easy.csv};
\addplot[name path=lae1, draw=none, forget plot]
  table[x=epoch, y expr=\thisrow{cfg1d}-\thisrow{cfg1s}, col sep=comma]
  {csv_tikz/pythia1b/benchmarks/arc_easy.csv};
\addplot[colorB!15, draw=none, forget plot] fill between[of=uae1 and lae1];
\addplot[colorB, line width=0.8pt]
  table[x=epoch, y=cfg1d, col sep=comma]
  {csv_tikz/pythia1b/benchmarks/arc_easy.csv};
\addplot[name path=uae2, draw=none, forget plot]
  table[x=epoch, y expr=\thisrow{cfg0d}+\thisrow{cfg0s}, col sep=comma]
  {csv_tikz/pythia6.9b/benchmarks/arc_easy.csv};
\addplot[name path=lae2, draw=none, forget plot]
  table[x=epoch, y expr=\thisrow{cfg0d}-\thisrow{cfg0s}, col sep=comma]
  {csv_tikz/pythia6.9b/benchmarks/arc_easy.csv};
\addplot[colorC!15, draw=none, forget plot] fill between[of=uae2 and lae2];
\addplot[colorC, line width=0.8pt, dashed]
  table[x=epoch, y=cfg0d, col sep=comma]
  {csv_tikz/pythia6.9b/benchmarks/arc_easy.csv};
\addplot[name path=uae3, draw=none, forget plot]
  table[x=epoch, y expr=\thisrow{cfg1d}+\thisrow{cfg1s}, col sep=comma]
  {csv_tikz/pythia6.9b/benchmarks/arc_easy.csv};
\addplot[name path=lae3, draw=none, forget plot]
  table[x=epoch, y expr=\thisrow{cfg1d}-\thisrow{cfg1s}, col sep=comma]
  {csv_tikz/pythia6.9b/benchmarks/arc_easy.csv};
\addplot[colorD!15, draw=none, forget plot] fill between[of=uae3 and lae3];
\addplot[colorD, line width=0.8pt, dashed]
  table[x=epoch, y=cfg1d, col sep=comma]
  {csv_tikz/pythia6.9b/benchmarks/arc_easy.csv};

\nextgroupplot[
  title={\tiny WinoGrande},
  ylabel={},
  axis x line*=bottom, axis y line*=left,
  xmin=0, xmax=40, ymin=-3.0, ymax=3.0,
  grid, grid style={line width=.2pt, draw=gray!30},
]
\draw[gray,dashed,line width=0.5pt](axis cs:0,0)--(axis cs:40,0);
\addplot[name path=uwg0, draw=none, forget plot]
  table[x=epoch, y expr=\thisrow{cfg0d}+\thisrow{cfg0s}, col sep=comma]
  {csv_tikz/pythia1b/benchmarks/winogrande.csv};
\addplot[name path=lwg0, draw=none, forget plot]
  table[x=epoch, y expr=\thisrow{cfg0d}-\thisrow{cfg0s}, col sep=comma]
  {csv_tikz/pythia1b/benchmarks/winogrande.csv};
\addplot[colorA!15, draw=none, forget plot] fill between[of=uwg0 and lwg0];
\addplot[colorA, line width=0.8pt]
  table[x=epoch, y=cfg0d, col sep=comma]
  {csv_tikz/pythia1b/benchmarks/winogrande.csv};
\addplot[name path=uwg1, draw=none, forget plot]
  table[x=epoch, y expr=\thisrow{cfg1d}+\thisrow{cfg1s}, col sep=comma]
  {csv_tikz/pythia1b/benchmarks/winogrande.csv};
\addplot[name path=lwg1, draw=none, forget plot]
  table[x=epoch, y expr=\thisrow{cfg1d}-\thisrow{cfg1s}, col sep=comma]
  {csv_tikz/pythia1b/benchmarks/winogrande.csv};
\addplot[colorB!15, draw=none, forget plot] fill between[of=uwg1 and lwg1];
\addplot[colorB, line width=0.8pt]
  table[x=epoch, y=cfg1d, col sep=comma]
  {csv_tikz/pythia1b/benchmarks/winogrande.csv};
\addplot[name path=uwg2, draw=none, forget plot]
  table[x=epoch, y expr=\thisrow{cfg0d}+\thisrow{cfg0s}, col sep=comma]
  {csv_tikz/pythia6.9b/benchmarks/winogrande.csv};
\addplot[name path=lwg2, draw=none, forget plot]
  table[x=epoch, y expr=\thisrow{cfg0d}-\thisrow{cfg0s}, col sep=comma]
  {csv_tikz/pythia6.9b/benchmarks/winogrande.csv};
\addplot[colorC!15, draw=none, forget plot] fill between[of=uwg2 and lwg2];
\addplot[colorC, line width=0.8pt, dashed]
  table[x=epoch, y=cfg0d, col sep=comma]
  {csv_tikz/pythia6.9b/benchmarks/winogrande.csv};
\addplot[name path=uwg3, draw=none, forget plot]
  table[x=epoch, y expr=\thisrow{cfg1d}+\thisrow{cfg1s}, col sep=comma]
  {csv_tikz/pythia6.9b/benchmarks/winogrande.csv};
\addplot[name path=lwg3, draw=none, forget plot]
  table[x=epoch, y expr=\thisrow{cfg1d}-\thisrow{cfg1s}, col sep=comma]
  {csv_tikz/pythia6.9b/benchmarks/winogrande.csv};
\addplot[colorD!15, draw=none, forget plot] fill between[of=uwg3 and lwg3];
\addplot[colorD, line width=0.8pt, dashed]
  table[x=epoch, y=cfg1d, col sep=comma]
  {csv_tikz/pythia6.9b/benchmarks/winogrande.csv};

\end{groupplot}

\matrix [
  matrix of nodes,
  anchor=north,
  below=4pt of current bounding box.south,
  nodes={anchor=west, font=\fontsize{6}{7}\selectfont},
  column sep=8pt,
  ampersand replacement=\&,
  draw=gray!60,
  rounded corners=2pt,
  inner sep=4pt,
  fill=white,
] {
  \draw[colorA,  solid,  line width=0.8pt] (0,0)--(0.5,0); \&
  \node{Pythia-1B\enspace $\mathrm{lr}=10^{-5}$}; \&
  \draw[colorB, solid,  line width=0.8pt] (0,0)--(0.5,0); \&
  \node{Pythia-1B\enspace $\mathrm{lr}=10^{-6}$}; \&
  \draw[colorC,   dashed, line width=0.8pt] (0,0)--(0.5,0); \&
  \node{Pythia-6.9B\enspace $\mathrm{lr}=10^{-5}$}; \&
  \draw[colorD,  dashed, line width=0.8pt] (0,0)--(0.5,0); \&
  \node{Pythia-6.9B\enspace $\mathrm{lr}=10^{-6}$}; \\
};

\end{tikzpicture}
}

\caption{\small \textbf{Benchmark capability under self-training is preserved or improved across four evaluations.}
$\Delta$ accuracy relative to the frozen base model over 40 epochs on ARC-Challenge (25-shot), HellaSwag (5-shot), ARC-Easy (0-shot), and WinoGrande (5-shot).}

\label{fig:capability}
\end{figure}

%% file: fig_memo.tex
%
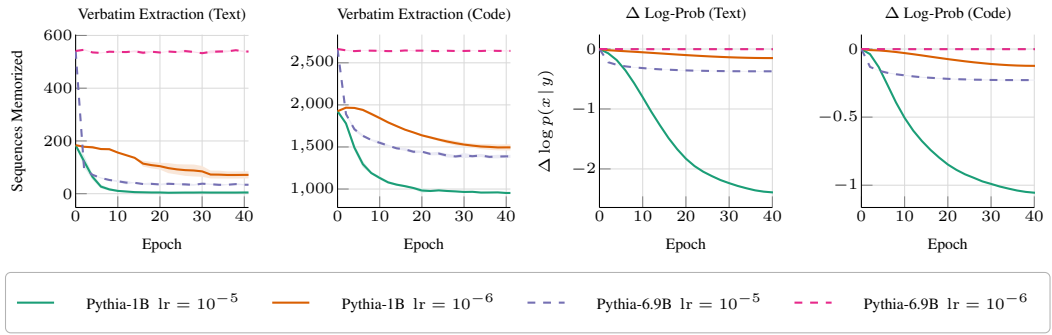
\begin{figure}[t]
\centering
\resizebox{\textwidth}{!}{%
\begin{tikzpicture}
\begin{groupplot}[
  group style={group size=4 by 1, horizontal sep=12mm},
  width=0.28\linewidth, height=0.28\linewidth,
  xtick={0,10,20,30,40},
  tick label style={font=\fontsize{4}{4}\selectfont},
  label style={font=\tiny},
  title style={font=\tiny, yshift=-1ex},
  xlabel={\tiny Epoch},
]

\nextgroupplot[
  title={\tiny Verbatim Extraction (Text)},
  ylabel={\tiny Sequences Memorized},
  axis x line*=bottom, axis y line*=left,
  xmin=0, xmax=41,
  grid, grid style={line width=.2pt, draw=gray!30},
]
\addplot[name path=utxt0, draw=none, forget plot]
  table[x=epoch, y expr=\thisrow{cfg0d}+\thisrow{cfg0s}, col sep=comma]
  {csv_tikz/pythia1b/memorization/combined_text.csv};
\addplot[name path=ltxt0, draw=none, forget plot]
  table[x=epoch, y expr=\thisrow{cfg0d}-\thisrow{cfg0s}, col sep=comma]
  {csv_tikz/pythia1b/memorization/combined_text.csv};
\addplot[colorA!15, draw=none, forget plot] fill between[of=utxt0 and ltxt0];
\addplot[colorA, line width=0.8pt]
  table[x=epoch, y=cfg0d, col sep=comma]
  {csv_tikz/pythia1b/memorization/combined_text.csv};
\addplot[name path=utxt1, draw=none, forget plot]
  table[x=epoch, y expr=\thisrow{cfg1d}+\thisrow{cfg1s}, col sep=comma]
  {csv_tikz/pythia1b/memorization/combined_text.csv};
\addplot[name path=ltxt1, draw=none, forget plot]
  table[x=epoch, y expr=\thisrow{cfg1d}-\thisrow{cfg1s}, col sep=comma]
  {csv_tikz/pythia1b/memorization/combined_text.csv};
\addplot[colorB!15, draw=none, forget plot] fill between[of=utxt1 and ltxt1];
\addplot[colorB, line width=0.8pt]
  table[x=epoch, y=cfg1d, col sep=comma]
  {csv_tikz/pythia1b/memorization/combined_text.csv};
\addplot[name path=utxt2, draw=none, forget plot]
  table[x=epoch, y expr=\thisrow{cfg0d}+\thisrow{cfg0s}, col sep=comma]
  {csv_tikz/pythia6.9b/memorization/combined_text.csv};
\addplot[name path=ltxt2, draw=none, forget plot]
  table[x=epoch, y expr=\thisrow{cfg0d}-\thisrow{cfg0s}, col sep=comma]
  {csv_tikz/pythia6.9b/memorization/combined_text.csv};
\addplot[colorC!15, draw=none, forget plot] fill between[of=utxt2 and ltxt2];
\addplot[colorC, line width=0.8pt, dashed]
  table[x=epoch, y=cfg0d, col sep=comma]
  {csv_tikz/pythia6.9b/memorization/combined_text.csv};
\addplot[name path=utxt3, draw=none, forget plot]
  table[x=epoch, y expr=\thisrow{cfg1d}+\thisrow{cfg1s}, col sep=comma]
  {csv_tikz/pythia6.9b/memorization/combined_text.csv};
\addplot[name path=ltxt3, draw=none, forget plot]
  table[x=epoch, y expr=\thisrow{cfg1d}-\thisrow{cfg1s}, col sep=comma]
  {csv_tikz/pythia6.9b/memorization/combined_text.csv};
\addplot[colorD!15, draw=none, forget plot] fill between[of=utxt3 and ltxt3];
\addplot[colorD, line width=0.8pt, dashed]
  table[x=epoch, y=cfg1d, col sep=comma]
  {csv_tikz/pythia6.9b/memorization/combined_text.csv};

\nextgroupplot[
  title={\tiny Verbatim Extraction (Code)},
  ylabel={},
  axis x line*=bottom, axis y line*=left,
  xmin=0, xmax=41,
  grid, grid style={line width=.2pt, draw=gray!30},
]
\addplot[name path=ucod0, draw=none, forget plot]
  table[x=epoch, y expr=\thisrow{cfg0d}+\thisrow{cfg0s}, col sep=comma]
  {csv_tikz/pythia1b/memorization/github.csv};
\addplot[name path=lcod0, draw=none, forget plot]
  table[x=epoch, y expr=\thisrow{cfg0d}-\thisrow{cfg0s}, col sep=comma]
  {csv_tikz/pythia1b/memorization/github.csv};
\addplot[colorA!15, draw=none, forget plot] fill between[of=ucod0 and lcod0];
\addplot[colorA, line width=0.8pt]
  table[x=epoch, y=cfg0d, col sep=comma]
  {csv_tikz/pythia1b/memorization/github.csv};
\addplot[name path=ucod1, draw=none, forget plot]
  table[x=epoch, y expr=\thisrow{cfg1d}+\thisrow{cfg1s}, col sep=comma]
  {csv_tikz/pythia1b/memorization/github.csv};
\addplot[name path=lcod1, draw=none, forget plot]
  table[x=epoch, y expr=\thisrow{cfg1d}-\thisrow{cfg1s}, col sep=comma]
  {csv_tikz/pythia1b/memorization/github.csv};
\addplot[colorB!15, draw=none, forget plot] fill between[of=ucod1 and lcod1];
\addplot[colorB, line width=0.8pt]
  table[x=epoch, y=cfg1d, col sep=comma]
  {csv_tikz/pythia1b/memorization/github.csv};
\addplot[name path=ucod2, draw=none, forget plot]
  table[x=epoch, y expr=\thisrow{cfg0d}+\thisrow{cfg0s}, col sep=comma]
  {csv_tikz/pythia6.9b/memorization/github.csv};
\addplot[name path=lcod2, draw=none, forget plot]
  table[x=epoch, y expr=\thisrow{cfg0d}-\thisrow{cfg0s}, col sep=comma]
  {csv_tikz/pythia6.9b/memorization/github.csv};
\addplot[colorC!15, draw=none, forget plot] fill between[of=ucod2 and lcod2];
\addplot[colorC, line width=0.8pt, dashed]
  table[x=epoch, y=cfg0d, col sep=comma]
  {csv_tikz/pythia6.9b/memorization/github.csv};
\addplot[name path=ucod3, draw=none, forget plot]
  table[x=epoch, y expr=\thisrow{cfg1d}+\thisrow{cfg1s}, col sep=comma]
  {csv_tikz/pythia6.9b/memorization/github.csv};
\addplot[name path=lcod3, draw=none, forget plot]
  table[x=epoch, y expr=\thisrow{cfg1d}-\thisrow{cfg1s}, col sep=comma]
  {csv_tikz/pythia6.9b/memorization/github.csv};
\addplot[colorD!15, draw=none, forget plot] fill between[of=ucod3 and lcod3];
\addplot[colorD, line width=0.8pt, dashed]
  table[x=epoch, y=cfg1d, col sep=comma]
  {csv_tikz/pythia6.9b/memorization/github.csv};

\nextgroupplot[
  title={\tiny $\Delta$ Log-Prob (Text)},
  ylabel={\tiny $\Delta \log p(x \!\mid\! y)$},
  axis x line*=bottom, axis y line*=left,
  xmin=0, xmax=40,
  grid, grid style={line width=.2pt, draw=gray!30},
]
\draw[gray,dashed,line width=0.5pt](axis cs:0,0)--(axis cs:40,0);
\addplot[name path=ulpt0, draw=none, forget plot]
  table[x=epoch, y expr=\thisrow{cfg0d}+\thisrow{cfg0s}, col sep=comma]
  {csv_tikz/pythia1b/exposure/combined_text.csv};
\addplot[name path=llpt0, draw=none, forget plot]
  table[x=epoch, y expr=\thisrow{cfg0d}-\thisrow{cfg0s}, col sep=comma]
  {csv_tikz/pythia1b/exposure/combined_text.csv};
\addplot[colorA!15, draw=none, forget plot] fill between[of=ulpt0 and llpt0];
\addplot[colorA, line width=0.8pt]
  table[x=epoch, y=cfg0d, col sep=comma]
  {csv_tikz/pythia1b/exposure/combined_text.csv};
\addplot[name path=ulpt1, draw=none, forget plot]
  table[x=epoch, y expr=\thisrow{cfg1d}+\thisrow{cfg1s}, col sep=comma]
  {csv_tikz/pythia1b/exposure/combined_text.csv};
\addplot[name path=llpt1, draw=none, forget plot]
  table[x=epoch, y expr=\thisrow{cfg1d}-\thisrow{cfg1s}, col sep=comma]
  {csv_tikz/pythia1b/exposure/combined_text.csv};
\addplot[colorB!15, draw=none, forget plot] fill between[of=ulpt1 and llpt1];
\addplot[colorB, line width=0.8pt]
  table[x=epoch, y=cfg1d, col sep=comma]
  {csv_tikz/pythia1b/exposure/combined_text.csv};
\addplot[name path=ulpt2, draw=none, forget plot]
  table[x=epoch, y expr=\thisrow{cfg0d}+\thisrow{cfg0s}, col sep=comma]
  {csv_tikz/pythia6.9b/exposure/combined_text.csv};
\addplot[name path=llpt2, draw=none, forget plot]
  table[x=epoch, y expr=\thisrow{cfg0d}-\thisrow{cfg0s}, col sep=comma]
  {csv_tikz/pythia6.9b/exposure/combined_text.csv};
\addplot[colorC!15, draw=none, forget plot] fill between[of=ulpt2 and llpt2];
\addplot[colorC, line width=0.8pt, dashed]
  table[x=epoch, y=cfg0d, col sep=comma]
  {csv_tikz/pythia6.9b/exposure/combined_text.csv};
\addplot[name path=ulpt3, draw=none, forget plot]
  table[x=epoch, y expr=\thisrow{cfg1d}+\thisrow{cfg1s}, col sep=comma]
  {csv_tikz/pythia6.9b/exposure/combined_text.csv};
\addplot[name path=llpt3, draw=none, forget plot]
  table[x=epoch, y expr=\thisrow{cfg1d}-\thisrow{cfg1s}, col sep=comma]
  {csv_tikz/pythia6.9b/exposure/combined_text.csv};
\addplot[colorD!15, draw=none, forget plot] fill between[of=ulpt3 and llpt3];
\addplot[colorD, line width=0.8pt, dashed]
  table[x=epoch, y=cfg1d, col sep=comma]
  {csv_tikz/pythia6.9b/exposure/combined_text.csv};

\nextgroupplot[
  title={\tiny $\Delta$ Log-Prob (Code)},
  ylabel={},
  axis x line*=bottom, axis y line*=left,
  xmin=0, xmax=40,
  grid, grid style={line width=.2pt, draw=gray!30},
]
\draw[gray,dashed,line width=0.5pt](axis cs:0,0)--(axis cs:40,0);
\addplot[name path=ulpc0, draw=none, forget plot]
  table[x=epoch, y expr=\thisrow{cfg0d}+\thisrow{cfg0s}, col sep=comma]
  {csv_tikz/pythia1b/exposure/github.csv};
\addplot[name path=llpc0, draw=none, forget plot]
  table[x=epoch, y expr=\thisrow{cfg0d}-\thisrow{cfg0s}, col sep=comma]
  {csv_tikz/pythia1b/exposure/github.csv};
\addplot[colorA!15, draw=none, forget plot] fill between[of=ulpc0 and llpc0];
\addplot[colorA, line width=0.8pt]
  table[x=epoch, y=cfg0d, col sep=comma]
  {csv_tikz/pythia1b/exposure/github.csv};
\addplot[name path=ulpc1, draw=none, forget plot]
  table[x=epoch, y expr=\thisrow{cfg1d}+\thisrow{cfg1s}, col sep=comma]
  {csv_tikz/pythia1b/exposure/github.csv};
\addplot[name path=llpc1, draw=none, forget plot]
  table[x=epoch, y expr=\thisrow{cfg1d}-\thisrow{cfg1s}, col sep=comma]
  {csv_tikz/pythia1b/exposure/github.csv};
\addplot[colorB!15, draw=none, forget plot] fill between[of=ulpc1 and llpc1];
\addplot[colorB, line width=0.8pt]
  table[x=epoch, y=cfg1d, col sep=comma]
  {csv_tikz/pythia1b/exposure/github.csv};
\addplot[name path=ulpc2, draw=none, forget plot]
  table[x=epoch, y expr=\thisrow{cfg0d}+\thisrow{cfg0s}, col sep=comma]
  {csv_tikz/pythia6.9b/exposure/github.csv};
\addplot[name path=llpc2, draw=none, forget plot]
  table[x=epoch, y expr=\thisrow{cfg0d}-\thisrow{cfg0s}, col sep=comma]
  {csv_tikz/pythia6.9b/exposure/github.csv};
\addplot[colorC!15, draw=none, forget plot] fill between[of=ulpc2 and llpc2];
\addplot[colorC, line width=0.8pt, dashed]
  table[x=epoch, y=cfg0d, col sep=comma]
  {csv_tikz/pythia6.9b/exposure/github.csv};
\addplot[name path=ulpc3, draw=none, forget plot]
  table[x=epoch, y expr=\thisrow{cfg1d}+\thisrow{cfg1s}, col sep=comma]
  {csv_tikz/pythia6.9b/exposure/github.csv};
\addplot[name path=llpc3, draw=none, forget plot]
  table[x=epoch, y expr=\thisrow{cfg1d}-\thisrow{cfg1s}, col sep=comma]
  {csv_tikz/pythia6.9b/exposure/github.csv};
\addplot[colorD!15, draw=none, forget plot] fill between[of=ulpc3 and llpc3];
\addplot[colorD, line width=0.8pt, dashed]
  table[x=epoch, y=cfg1d, col sep=comma]
  {csv_tikz/pythia6.9b/exposure/github.csv};

\end{groupplot}

\matrix [
  matrix of nodes,
  anchor=north,
  below=4pt of current bounding box.south,
  nodes={anchor=west, font=\fontsize{6}{7}\selectfont},
  column sep=8pt,
  ampersand replacement=\&,
  draw=gray!60,
  rounded corners=2pt,
  inner sep=4pt,
  fill=white,
] {
  \draw[colorA,  solid,  line width=0.8pt] (0,0)--(0.5,0); \&
  \node{Pythia-1B\enspace $\mathrm{lr}=10^{-5}$}; \&
  \draw[colorB, solid,  line width=0.8pt] (0,0)--(0.5,0); \&
  \node{Pythia-1B\enspace $\mathrm{lr}=10^{-6}$}; \&
  \draw[colorC,   dashed, line width=0.8pt] (0,0)--(0.5,0); \&
  \node{Pythia-6.9B\enspace $\mathrm{lr}=10^{-5}$}; \&
  \draw[colorD,  dashed, line width=0.8pt] (0,0)--(0.5,0); \&
  \node{Pythia-6.9B\enspace $\mathrm{lr}=10^{-6}$}; \\
};

\end{tikzpicture}
}

\caption{\small \textbf{Capability and memorization move in opposite directions.}
Left two panels: number of verbatim sequences extracted via the prefix-completion attack of \citet{carlini2021extracting} on text corpora (Enron, Pile-CC, Wikipedia) and code (GitHub).
Right two panels: average log-probability difference $\log p_{\theta_t}(x \!\mid\! y) - \log p_{\theta_{\mathrm{base}}}(x \!\mid\! y)$ of the true continuation under each checkpoint relative to the base model, evaluated on the same held-out sequences. Values below zero indicate the checkpoint assigns less probability to pretraining sequences. The extraction panels measure recoverability under greedy decoding; the log-probability panels confirm the effect is representational rather than a decoding artifact.
}

\label{fig:memo}
\end{figure}

%% file: figure_qwen_fewshot.tex
\begin{figure}[h]
\centering
\resizebox{\textwidth}{!}{%
\begin{tikzpicture}
\begin{groupplot}[
  group style={group size=4 by 1, horizontal sep=5mm, vertical sep=8mm},
  width=0.27\linewidth, height=0.24\linewidth,
  xtick={0,10,20,30,40},
  tick label style={font=\fontsize{4}{4}\selectfont},
  label style={font=\tiny},
  title style={font=\tiny, yshift=-1ex},
]

  \nextgroupplot[
    title={\tiny ARC Challenge},
    ylabel={\tiny $\Delta$ Acc.},
    xlabel={\tiny Epoch},
    axis x line*=bottom, axis y line*=left,
    xmin=0, xmax=40, ymin=-1.0, ymax=4.0,
    grid, grid style={line width=.2pt, draw=gray!30},
  ]
  \draw[gray,dashed,line width=0.5pt](axis cs:0,0)--(axis cs:40,0);
  \node[anchor=north west, font=\tiny, text=gray!70!black,
    fill=white, fill opacity=0.75, text opacity=1,
    inner sep=1pt, rounded corners=1pt]
    at (axis cs:1,3.8) {base$=36.1\%$};
  \addplot[name path=uarcA, draw=none, forget plot]
    table[x=epoch, y expr=\thisrow{cfg0d}+\thisrow{cfg0s}, col sep=comma]{csv_tikz/qwen_fewshot/arc_challenge.csv};
  \addplot[name path=larcA, draw=none, forget plot]
    table[x=epoch, y expr=\thisrow{cfg0d}-\thisrow{cfg0s}, col sep=comma]{csv_tikz/qwen_fewshot/arc_challenge.csv};
  \addplot[colorA!15, draw=none, forget plot] fill between[of=uarcA and larcA];
  \addplot[colorA, line width=0.8pt]
    table[x=epoch, y=cfg0d, col sep=comma]{csv_tikz/qwen_fewshot/arc_challenge.csv};
  \addplot[name path=uarcC, draw=none, forget plot]
    table[x=epoch, y expr=\thisrow{cfg3d}+\thisrow{cfg3s}, col sep=comma]{csv_tikz/qwen_fewshot/arc_challenge.csv};
  \addplot[name path=larcC, draw=none, forget plot]
    table[x=epoch, y expr=\thisrow{cfg3d}-\thisrow{cfg3s}, col sep=comma]{csv_tikz/qwen_fewshot/arc_challenge.csv};
  \addplot[colorC!15, draw=none, forget plot] fill between[of=uarcC and larcC];
  \addplot[colorC, line width=0.8pt]
    table[x=epoch, y=cfg3d, col sep=comma]{csv_tikz/qwen_fewshot/arc_challenge.csv};
  \addplot[name path=uarcB, draw=none, forget plot]
    table[x=epoch, y expr=\thisrow{cfg1d}+\thisrow{cfg1s}, col sep=comma]{csv_tikz/qwen_fewshot/arc_challenge.csv};
  \addplot[name path=larcB, draw=none, forget plot]
    table[x=epoch, y expr=\thisrow{cfg1d}-\thisrow{cfg1s}, col sep=comma]{csv_tikz/qwen_fewshot/arc_challenge.csv};
  \addplot[colorB!15, draw=none, forget plot] fill between[of=uarcB and larcB];
  \addplot[colorB, line width=0.8pt]
    table[x=epoch, y=cfg1d, col sep=comma]{csv_tikz/qwen_fewshot/arc_challenge.csv};
  \addplot[name path=uarcD, draw=none, forget plot]
    table[x=epoch, y expr=\thisrow{cfg2d}+\thisrow{cfg2s}, col sep=comma]{csv_tikz/qwen_fewshot/arc_challenge.csv};
  \addplot[name path=larcD, draw=none, forget plot]
    table[x=epoch, y expr=\thisrow{cfg2d}-\thisrow{cfg2s}, col sep=comma]{csv_tikz/qwen_fewshot/arc_challenge.csv};
  \addplot[colorD!15, draw=none, forget plot] fill between[of=uarcD and larcD];
  \addplot[colorD, line width=0.8pt]
    table[x=epoch, y=cfg2d, col sep=comma]{csv_tikz/qwen_fewshot/arc_challenge.csv};

  \nextgroupplot[
    title={\tiny GSM8K},
    ylabel={},
    xlabel={\tiny Epoch},
    axis x line*=bottom, axis y line*=left,
    xmin=0, xmax=40, ymin=-4.0, ymax=5.0,
    grid, grid style={line width=.2pt, draw=gray!30},
  ]
  \draw[gray,dashed,line width=0.5pt](axis cs:0,0)--(axis cs:40,0);
  \node[anchor=north east, font=\tiny, text=gray!70!black,
    fill=white, fill opacity=0.75, text opacity=1,
    inner sep=1pt, rounded corners=1pt]
    at (axis cs:39,4.8) {base$=33.7\%$};
  \addplot[name path=ugsmA, draw=none, forget plot]
    table[x=epoch, y expr=\thisrow{cfg0d}+\thisrow{cfg0s}, col sep=comma]{csv_tikz/qwen_fewshot/gsm8k.csv};
  \addplot[name path=lgsmA, draw=none, forget plot]
    table[x=epoch, y expr=\thisrow{cfg0d}-\thisrow{cfg0s}, col sep=comma]{csv_tikz/qwen_fewshot/gsm8k.csv};
  \addplot[colorA!15, draw=none, forget plot] fill between[of=ugsmA and lgsmA];
  \addplot[colorA, line width=0.8pt]
    table[x=epoch, y=cfg0d, col sep=comma]{csv_tikz/qwen_fewshot/gsm8k.csv};
  \addplot[name path=ugsmC, draw=none, forget plot]
    table[x=epoch, y expr=\thisrow{cfg3d}+\thisrow{cfg3s}, col sep=comma]{csv_tikz/qwen_fewshot/gsm8k.csv};
  \addplot[name path=lgsmC, draw=none, forget plot]
    table[x=epoch, y expr=\thisrow{cfg3d}-\thisrow{cfg3s}, col sep=comma]{csv_tikz/qwen_fewshot/gsm8k.csv};
  \addplot[colorC!15, draw=none, forget plot] fill between[of=ugsmC and lgsmC];
  \addplot[colorC, line width=0.8pt]
    table[x=epoch, y=cfg3d, col sep=comma]{csv_tikz/qwen_fewshot/gsm8k.csv};
  \addplot[name path=ugsmB, draw=none, forget plot]
    table[x=epoch, y expr=\thisrow{cfg1d}+\thisrow{cfg1s}, col sep=comma]{csv_tikz/qwen_fewshot/gsm8k.csv};
  \addplot[name path=lgsmB, draw=none, forget plot]
    table[x=epoch, y expr=\thisrow{cfg1d}-\thisrow{cfg1s}, col sep=comma]{csv_tikz/qwen_fewshot/gsm8k.csv};
  \addplot[colorB!15, draw=none, forget plot] fill between[of=ugsmB and lgsmB];
  \addplot[colorB, line width=0.8pt]
    table[x=epoch, y=cfg1d, col sep=comma]{csv_tikz/qwen_fewshot/gsm8k.csv};
  \addplot[name path=ugsmD, draw=none, forget plot]
    table[x=epoch, y expr=\thisrow{cfg2d}+\thisrow{cfg2s}, col sep=comma]{csv_tikz/qwen_fewshot/gsm8k.csv};
  \addplot[name path=lgsmD, draw=none, forget plot]
    table[x=epoch, y expr=\thisrow{cfg2d}-\thisrow{cfg2s}, col sep=comma]{csv_tikz/qwen_fewshot/gsm8k.csv};
  \addplot[colorD!15, draw=none, forget plot] fill between[of=ugsmD and lgsmD];
  \addplot[colorD, line width=0.8pt]
    table[x=epoch, y=cfg2d, col sep=comma]{csv_tikz/qwen_fewshot/gsm8k.csv};

  \nextgroupplot[
    title={\tiny HellaSwag},
    ylabel={},
    xlabel={\tiny Epoch},
    axis x line*=bottom, axis y line*=left,
    xmin=0, xmax=40, ymin=-1.0, ymax=3.5,
    grid, grid style={line width=.2pt, draw=gray!30},
  ]
  \draw[gray,dashed,line width=0.5pt](axis cs:0,0)--(axis cs:40,0);
  \node[anchor=north west, font=\tiny, text=gray!70!black,
    fill=white, fill opacity=0.75, text opacity=1,
    inner sep=1pt, rounded corners=1pt]
    at (axis cs:1,3.3) {base$=51.4\%$};
  \addplot[name path=uhelA, draw=none, forget plot]
    table[x=epoch, y expr=\thisrow{cfg0d}+\thisrow{cfg0s}, col sep=comma]{csv_tikz/qwen_fewshot/hellaswag.csv};
  \addplot[name path=lhelA, draw=none, forget plot]
    table[x=epoch, y expr=\thisrow{cfg0d}-\thisrow{cfg0s}, col sep=comma]{csv_tikz/qwen_fewshot/hellaswag.csv};
  \addplot[colorA!15, draw=none, forget plot] fill between[of=uhelA and lhelA];
  \addplot[colorA, line width=0.8pt]
    table[x=epoch, y=cfg0d, col sep=comma]{csv_tikz/qwen_fewshot/hellaswag.csv};
  \addplot[name path=uhelC, draw=none, forget plot]
    table[x=epoch, y expr=\thisrow{cfg3d}+\thisrow{cfg3s}, col sep=comma]{csv_tikz/qwen_fewshot/hellaswag.csv};
  \addplot[name path=lhelC, draw=none, forget plot]
    table[x=epoch, y expr=\thisrow{cfg3d}-\thisrow{cfg3s}, col sep=comma]{csv_tikz/qwen_fewshot/hellaswag.csv};
  \addplot[colorC!15, draw=none, forget plot] fill between[of=uhelC and lhelC];
  \addplot[colorC, line width=0.8pt]
    table[x=epoch, y=cfg3d, col sep=comma]{csv_tikz/qwen_fewshot/hellaswag.csv};
  \addplot[name path=uhelB, draw=none, forget plot]
    table[x=epoch, y expr=\thisrow{cfg1d}+\thisrow{cfg1s}, col sep=comma]{csv_tikz/qwen_fewshot/hellaswag.csv};
  \addplot[name path=lhelB, draw=none, forget plot]
    table[x=epoch, y expr=\thisrow{cfg1d}-\thisrow{cfg1s}, col sep=comma]{csv_tikz/qwen_fewshot/hellaswag.csv};
  \addplot[colorB!15, draw=none, forget plot] fill between[of=uhelB and lhelB];
  \addplot[colorB, line width=0.8pt]
    table[x=epoch, y=cfg1d, col sep=comma]{csv_tikz/qwen_fewshot/hellaswag.csv};
  \addplot[name path=uhelD, draw=none, forget plot]
    table[x=epoch, y expr=\thisrow{cfg2d}+\thisrow{cfg2s}, col sep=comma]{csv_tikz/qwen_fewshot/hellaswag.csv};
  \addplot[name path=lhelD, draw=none, forget plot]
    table[x=epoch, y expr=\thisrow{cfg2d}-\thisrow{cfg2s}, col sep=comma]{csv_tikz/qwen_fewshot/hellaswag.csv};
  \addplot[colorD!15, draw=none, forget plot] fill between[of=uhelD and lhelD];
  \addplot[colorD, line width=0.8pt]
    table[x=epoch, y=cfg2d, col sep=comma]{csv_tikz/qwen_fewshot/hellaswag.csv};

  \nextgroupplot[
    title={\tiny Minerva-MATH},
    ylabel={},
    xlabel={\tiny Epoch},
    axis x line*=bottom, axis y line*=left,
    xmin=0, xmax=40, ymin=-2.0, ymax=2.0,
    grid, grid style={line width=.2pt, draw=gray!30},
  ]
  \draw[gray,dashed,line width=0.5pt](axis cs:0,0)--(axis cs:40,0);
  \node[anchor=north east, font=\tiny, text=gray!70!black,
    fill=white, fill opacity=0.75, text opacity=1,
    inner sep=1pt, rounded corners=1pt]
    at (axis cs:39,1.8) {base$=16.5\%$};
  \addplot[name path=umathA, draw=none, forget plot]
    table[x=epoch, y expr=\thisrow{cfg0d}+\thisrow{cfg0s}, col sep=comma]{csv_tikz/qwen_fewshot/minerva_math.csv};
  \addplot[name path=lmathA, draw=none, forget plot]
    table[x=epoch, y expr=\thisrow{cfg0d}-\thisrow{cfg0s}, col sep=comma]{csv_tikz/qwen_fewshot/minerva_math.csv};
  \addplot[colorA!15, draw=none, forget plot] fill between[of=umathA and lmathA];
  \addplot[colorA, line width=0.8pt]
    table[x=epoch, y=cfg0d, col sep=comma]{csv_tikz/qwen_fewshot/minerva_math.csv};
  \addplot[name path=umathC, draw=none, forget plot]
    table[x=epoch, y expr=\thisrow{cfg3d}+\thisrow{cfg3s}, col sep=comma]{csv_tikz/qwen_fewshot/minerva_math.csv};
  \addplot[name path=lmathC, draw=none, forget plot]
    table[x=epoch, y expr=\thisrow{cfg3d}-\thisrow{cfg3s}, col sep=comma]{csv_tikz/qwen_fewshot/minerva_math.csv};
  \addplot[colorC!15, draw=none, forget plot] fill between[of=umathC and lmathC];
  \addplot[colorC, line width=0.8pt]
    table[x=epoch, y=cfg3d, col sep=comma]{csv_tikz/qwen_fewshot/minerva_math.csv};
  \addplot[name path=umathB, draw=none, forget plot]
    table[x=epoch, y expr=\thisrow{cfg1d}+\thisrow{cfg1s}, col sep=comma]{csv_tikz/qwen_fewshot/minerva_math.csv};
  \addplot[name path=lmathB, draw=none, forget plot]
    table[x=epoch, y expr=\thisrow{cfg1d}-\thisrow{cfg1s}, col sep=comma]{csv_tikz/qwen_fewshot/minerva_math.csv};
  \addplot[colorB!15, draw=none, forget plot] fill between[of=umathB and lmathB];
  \addplot[colorB, line width=0.8pt]
    table[x=epoch, y=cfg1d, col sep=comma]{csv_tikz/qwen_fewshot/minerva_math.csv};
  \addplot[name path=umathD, draw=none, forget plot]
    table[x=epoch, y expr=\thisrow{cfg2d}+\thisrow{cfg2s}, col sep=comma]{csv_tikz/qwen_fewshot/minerva_math.csv};
  \addplot[name path=lmathD, draw=none, forget plot]
    table[x=epoch, y expr=\thisrow{cfg2d}-\thisrow{cfg2s}, col sep=comma]{csv_tikz/qwen_fewshot/minerva_math.csv};
  \addplot[colorD!15, draw=none, forget plot] fill between[of=umathD and lmathD];
  \addplot[colorD, line width=0.8pt]
    table[x=epoch, y=cfg2d, col sep=comma]{csv_tikz/qwen_fewshot/minerva_math.csv};

\end{groupplot}

\matrix [
  matrix of nodes,
  anchor=north,
  below=4pt of current bounding box.south,
  nodes={anchor=west, font=\fontsize{6}{7}\selectfont},
  column sep=8pt,
  ampersand replacement=\&,
  draw=gray!60,
  rounded corners=2pt,
  inner sep=4pt,
  fill=white,
] {
  \draw[colorA, solid, line width=0.8pt] (0,0)--(0.5,0); \&
  \node{Real Data}; \&
  \draw[colorC, solid, line width=0.8pt] (0,0)--(0.5,0); \&
  \node{Synthetic $\tau{=}0.75$}; \&
  \draw[colorB, solid, line width=0.8pt] (0,0)--(0.5,0); \&
  \node{Synthetic $\tau{=}1.0$}; \&
  \draw[colorD, solid, line width=0.8pt] (0,0)--(0.5,0); \&
  \node{Synthetic $\tau{=}1.25$}; \\
};

\end{tikzpicture}
}
\caption{\small \textbf{Few-shot evaluation mirrors the zero-shot pattern: self-generated data produces gains on structured reasoning while generic replay does not.}
Qwen2.5-0.5B $\Delta$ accuracy under standard few-shot protocols (ARC-Challenge 25-shot, GSM8K 8-shot, HellaSwag 5-shot, Minerva-MATH 4-shot) over 40 epochs. Shaded bands show $\pm 1$ std.\ across subsets.}
\label{fig:qwen_fewshot}
\end{figure}
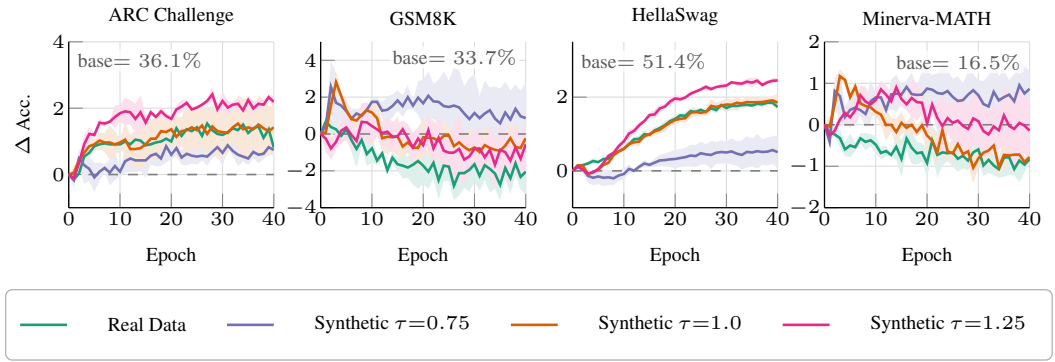

%% file: figure_llama_qwen.tex
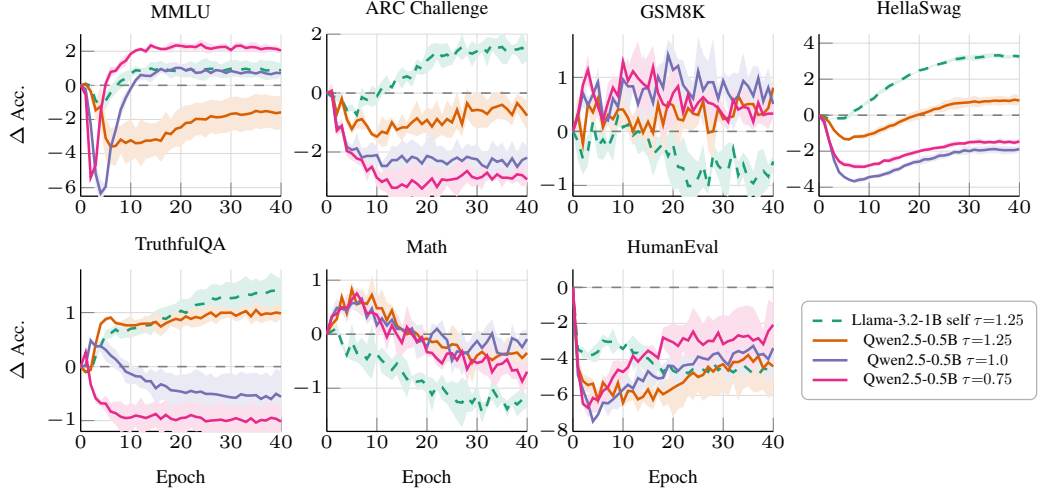
\begin{figure}[h]
\centering
\resizebox{\textwidth}{!}{%
\begin{tikzpicture}
\usepgfplotslibrary{fillbetween}
\begin{groupplot}[
  group style={group size=4 by 2, horizontal sep=5mm, vertical sep=8mm},
  width=0.27\linewidth, height=0.24\linewidth,
  xtick={0,10,20,30,40},
  tick label style={font=\fontsize{4}{4}\selectfont},
  label style={font=\tiny},
  title style={font=\tiny, yshift=-1ex},
]

  \nextgroupplot[
    title={\tiny MMLU},
    ylabel={\tiny $\Delta$ Acc.},
    xlabel={},
    axis x line*=bottom, axis y line*=left,
    xmin=0, xmax=40, ymin=-6.5, ymax=3.0,
    grid, grid style={line width=.2pt, draw=gray!30},
  ]
  \draw[gray,dashed,line width=0.5pt](axis cs:0,0)--(axis cs:40,0);
  \addplot[name path=ummlu0, draw=none, forget plot]
    table[x=epoch, y expr=\thisrow{cfg0d}+\thisrow{cfg0s}, col sep=comma]{csv_tikz/llama-qwen/mmlu.csv};
  \addplot[name path=lmmlu0, draw=none, forget plot]
    table[x=epoch, y expr=\thisrow{cfg0d}-\thisrow{cfg0s}, col sep=comma]{csv_tikz/llama-qwen/mmlu.csv};
  \addplot[colorA!15, draw=none, forget plot] fill between[of=ummlu0 and lmmlu0];
  \addplot[colorA, line width=0.8pt, dashed]
    table[x=epoch, y=cfg0d, col sep=comma]{csv_tikz/llama-qwen/mmlu.csv};
  \addplot[name path=ummlu1, draw=none, forget plot]
    table[x=epoch, y expr=\thisrow{cfg1d}+\thisrow{cfg1s}, col sep=comma]{csv_tikz/llama-qwen/mmlu.csv};
  \addplot[name path=lmmlu1, draw=none, forget plot]
    table[x=epoch, y expr=\thisrow{cfg1d}-\thisrow{cfg1s}, col sep=comma]{csv_tikz/llama-qwen/mmlu.csv};
  \addplot[colorB!15, draw=none, forget plot] fill between[of=ummlu1 and lmmlu1];
  \addplot[colorB, line width=0.8pt]
    table[x=epoch, y=cfg1d, col sep=comma]{csv_tikz/llama-qwen/mmlu.csv};
  \addplot[name path=ummlu2, draw=none, forget plot]
    table[x=epoch, y expr=\thisrow{cfg2d}+\thisrow{cfg2s}, col sep=comma]{csv_tikz/llama-qwen/mmlu.csv};
  \addplot[name path=lmmlu2, draw=none, forget plot]
    table[x=epoch, y expr=\thisrow{cfg2d}-\thisrow{cfg2s}, col sep=comma]{csv_tikz/llama-qwen/mmlu.csv};
  \addplot[colorC!15, draw=none, forget plot] fill between[of=ummlu2 and lmmlu2];
  \addplot[colorC, line width=0.8pt]
    table[x=epoch, y=cfg2d, col sep=comma]{csv_tikz/llama-qwen/mmlu.csv};
  \addplot[name path=ummlu3, draw=none, forget plot]
    table[x=epoch, y expr=\thisrow{cfg3d}+\thisrow{cfg3s}, col sep=comma]{csv_tikz/llama-qwen/mmlu.csv};
  \addplot[name path=lmmlu3, draw=none, forget plot]
    table[x=epoch, y expr=\thisrow{cfg3d}-\thisrow{cfg3s}, col sep=comma]{csv_tikz/llama-qwen/mmlu.csv};
  \addplot[colorD!15, draw=none, forget plot] fill between[of=ummlu3 and lmmlu3];
  \addplot[colorD, line width=0.8pt]
    table[x=epoch, y=cfg3d, col sep=comma]{csv_tikz/llama-qwen/mmlu.csv};

  \nextgroupplot[
    title={\tiny ARC Challenge},
    ylabel={},
    xlabel={},
    axis x line*=bottom, axis y line*=left,
    xmin=0, xmax=40, ymin=-3.5, ymax=2.0,
    grid, grid style={line width=.2pt, draw=gray!30},
  ]
  \draw[gray,dashed,line width=0.5pt](axis cs:0,0)--(axis cs:40,0);
  \addplot[name path=uarc0, draw=none, forget plot]
    table[x=epoch, y expr=\thisrow{cfg0d}+\thisrow{cfg0s}, col sep=comma]{csv_tikz/llama-qwen/arc_challenge.csv};
  \addplot[name path=larc0, draw=none, forget plot]
    table[x=epoch, y expr=\thisrow{cfg0d}-\thisrow{cfg0s}, col sep=comma]{csv_tikz/llama-qwen/arc_challenge.csv};
  \addplot[colorA!15, draw=none, forget plot] fill between[of=uarc0 and larc0];
  \addplot[colorA, line width=0.8pt, dashed]
    table[x=epoch, y=cfg0d, col sep=comma]{csv_tikz/llama-qwen/arc_challenge.csv};
  \addplot[name path=uarc1, draw=none, forget plot]
    table[x=epoch, y expr=\thisrow{cfg1d}+\thisrow{cfg1s}, col sep=comma]{csv_tikz/llama-qwen/arc_challenge.csv};
  \addplot[name path=larc1, draw=none, forget plot]
    table[x=epoch, y expr=\thisrow{cfg1d}-\thisrow{cfg1s}, col sep=comma]{csv_tikz/llama-qwen/arc_challenge.csv};
  \addplot[colorB!15, draw=none, forget plot] fill between[of=uarc1 and larc1];
  \addplot[colorB, line width=0.8pt]
    table[x=epoch, y=cfg1d, col sep=comma]{csv_tikz/llama-qwen/arc_challenge.csv};
  \addplot[name path=uarc2, draw=none, forget plot]
    table[x=epoch, y expr=\thisrow{cfg2d}+\thisrow{cfg2s}, col sep=comma]{csv_tikz/llama-qwen/arc_challenge.csv};
  \addplot[name path=larc2, draw=none, forget plot]
    table[x=epoch, y expr=\thisrow{cfg2d}-\thisrow{cfg2s}, col sep=comma]{csv_tikz/llama-qwen/arc_challenge.csv};
  \addplot[colorC!15, draw=none, forget plot] fill between[of=uarc2 and larc2];
  \addplot[colorC, line width=0.8pt]
    table[x=epoch, y=cfg2d, col sep=comma]{csv_tikz/llama-qwen/arc_challenge.csv};
  \addplot[name path=uarc3, draw=none, forget plot]
    table[x=epoch, y expr=\thisrow{cfg3d}+\thisrow{cfg3s}, col sep=comma]{csv_tikz/llama-qwen/arc_challenge.csv};
  \addplot[name path=larc3, draw=none, forget plot]
    table[x=epoch, y expr=\thisrow{cfg3d}-\thisrow{cfg3s}, col sep=comma]{csv_tikz/llama-qwen/arc_challenge.csv};
  \addplot[colorD!15, draw=none, forget plot] fill between[of=uarc3 and larc3];
  \addplot[colorD, line width=0.8pt]
    table[x=epoch, y=cfg3d, col sep=comma]{csv_tikz/llama-qwen/arc_challenge.csv};

  \nextgroupplot[
    title={\tiny GSM8K},
    ylabel={},
    xlabel={},
    axis x line*=bottom, axis y line*=left,
    xmin=0, xmax=40, ymin=-1.2, ymax=1.8,
    grid, grid style={line width=.2pt, draw=gray!30},
  ]
  \draw[gray,dashed,line width=0.5pt](axis cs:0,0)--(axis cs:40,0);
  \addplot[name path=ugsm0, draw=none, forget plot]
    table[x=epoch, y expr=\thisrow{cfg0d}+\thisrow{cfg0s}, col sep=comma]{csv_tikz/llama-qwen/gsm8k.csv};
  \addplot[name path=lgsm0, draw=none, forget plot]
    table[x=epoch, y expr=\thisrow{cfg0d}-\thisrow{cfg0s}, col sep=comma]{csv_tikz/llama-qwen/gsm8k.csv};
  \addplot[colorA!15, draw=none, forget plot] fill between[of=ugsm0 and lgsm0];
  \addplot[colorA, line width=0.8pt, dashed]
    table[x=epoch, y=cfg0d, col sep=comma]{csv_tikz/llama-qwen/gsm8k.csv};
  \addplot[name path=ugsm1, draw=none, forget plot]
    table[x=epoch, y expr=\thisrow{cfg1d}+\thisrow{cfg1s}, col sep=comma]{csv_tikz/llama-qwen/gsm8k.csv};
  \addplot[name path=lgsm1, draw=none, forget plot]
    table[x=epoch, y expr=\thisrow{cfg1d}-\thisrow{cfg1s}, col sep=comma]{csv_tikz/llama-qwen/gsm8k.csv};
  \addplot[colorB!15, draw=none, forget plot] fill between[of=ugsm1 and lgsm1];
  \addplot[colorB, line width=0.8pt]
    table[x=epoch, y=cfg1d, col sep=comma]{csv_tikz/llama-qwen/gsm8k.csv};
  \addplot[name path=ugsm2, draw=none, forget plot]
    table[x=epoch, y expr=\thisrow{cfg2d}+\thisrow{cfg2s}, col sep=comma]{csv_tikz/llama-qwen/gsm8k.csv};
  \addplot[name path=lgsm2, draw=none, forget plot]
    table[x=epoch, y expr=\thisrow{cfg2d}-\thisrow{cfg2s}, col sep=comma]{csv_tikz/llama-qwen/gsm8k.csv};
  \addplot[colorC!15, draw=none, forget plot] fill between[of=ugsm2 and lgsm2];
  \addplot[colorC, line width=0.8pt]
    table[x=epoch, y=cfg2d, col sep=comma]{csv_tikz/llama-qwen/gsm8k.csv};
  \addplot[name path=ugsm3, draw=none, forget plot]
    table[x=epoch, y expr=\thisrow{cfg3d}+\thisrow{cfg3s}, col sep=comma]{csv_tikz/llama-qwen/gsm8k.csv};
  \addplot[name path=lgsm3, draw=none, forget plot]
    table[x=epoch, y expr=\thisrow{cfg3d}-\thisrow{cfg3s}, col sep=comma]{csv_tikz/llama-qwen/gsm8k.csv};
  \addplot[colorD!15, draw=none, forget plot] fill between[of=ugsm3 and lgsm3];
  \addplot[colorD, line width=0.8pt]
    table[x=epoch, y=cfg3d, col sep=comma]{csv_tikz/llama-qwen/gsm8k.csv};

  \nextgroupplot[
    title={\tiny HellaSwag},
    ylabel={},
    xlabel={},
    axis x line*=bottom, axis y line*=left,
    xmin=0, xmax=40, ymin=-4.5, ymax=4.5,
    grid, grid style={line width=.2pt, draw=gray!30},
  ]
  \draw[gray,dashed,line width=0.5pt](axis cs:0,0)--(axis cs:40,0);
  \addplot[name path=uhella0, draw=none, forget plot]
    table[x=epoch, y expr=\thisrow{cfg0d}+\thisrow{cfg0s}, col sep=comma]{csv_tikz/llama-qwen/hellaswag.csv};
  \addplot[name path=lhella0, draw=none, forget plot]
    table[x=epoch, y expr=\thisrow{cfg0d}-\thisrow{cfg0s}, col sep=comma]{csv_tikz/llama-qwen/hellaswag.csv};
  \addplot[colorA!15, draw=none, forget plot] fill between[of=uhella0 and lhella0];
  \addplot[colorA, line width=0.8pt, dashed]
    table[x=epoch, y=cfg0d, col sep=comma]{csv_tikz/llama-qwen/hellaswag.csv};
  \addplot[name path=uhella1, draw=none, forget plot]
    table[x=epoch, y expr=\thisrow{cfg1d}+\thisrow{cfg1s}, col sep=comma]{csv_tikz/llama-qwen/hellaswag.csv};
  \addplot[name path=lhella1, draw=none, forget plot]
    table[x=epoch, y expr=\thisrow{cfg1d}-\thisrow{cfg1s}, col sep=comma]{csv_tikz/llama-qwen/hellaswag.csv};
  \addplot[colorB!15, draw=none, forget plot] fill between[of=uhella1 and lhella1];
  \addplot[colorB, line width=0.8pt]
    table[x=epoch, y=cfg1d, col sep=comma]{csv_tikz/llama-qwen/hellaswag.csv};
  \addplot[name path=uhella2, draw=none, forget plot]
    table[x=epoch, y expr=\thisrow{cfg2d}+\thisrow{cfg2s}, col sep=comma]{csv_tikz/llama-qwen/hellaswag.csv};
  \addplot[name path=lhella2, draw=none, forget plot]
    table[x=epoch, y expr=\thisrow{cfg2d}-\thisrow{cfg2s}, col sep=comma]{csv_tikz/llama-qwen/hellaswag.csv};
  \addplot[colorC!15, draw=none, forget plot] fill between[of=uhella2 and lhella2];
  \addplot[colorC, line width=0.8pt]
    table[x=epoch, y=cfg2d, col sep=comma]{csv_tikz/llama-qwen/hellaswag.csv};
  \addplot[name path=uhella3, draw=none, forget plot]
    table[x=epoch, y expr=\thisrow{cfg3d}+\thisrow{cfg3s}, col sep=comma]{csv_tikz/llama-qwen/hellaswag.csv};
  \addplot[name path=lhella3, draw=none, forget plot]
    table[x=epoch, y expr=\thisrow{cfg3d}-\thisrow{cfg3s}, col sep=comma]{csv_tikz/llama-qwen/hellaswag.csv};
  \addplot[colorD!15, draw=none, forget plot] fill between[of=uhella3 and lhella3];
  \addplot[colorD, line width=0.8pt]
    table[x=epoch, y=cfg3d, col sep=comma]{csv_tikz/llama-qwen/hellaswag.csv};

  \nextgroupplot[
    title={\tiny TruthfulQA},
    ylabel={\tiny $\Delta$ Acc.},
    xlabel={\tiny Epoch},
    axis x line*=bottom, axis y line*=left,
    xmin=0, xmax=40, ymin=-1.2, ymax=1.8,
    grid, grid style={line width=.2pt, draw=gray!30},
  ]
  \draw[gray,dashed,line width=0.5pt](axis cs:0,0)--(axis cs:40,0);
  \addplot[name path=utruth0, draw=none, forget plot]
    table[x=epoch, y expr=\thisrow{cfg0d}+\thisrow{cfg0s}, col sep=comma]{csv_tikz/llama-qwen/truthfulqa_mc2.csv};
  \addplot[name path=ltruth0, draw=none, forget plot]
    table[x=epoch, y expr=\thisrow{cfg0d}-\thisrow{cfg0s}, col sep=comma]{csv_tikz/llama-qwen/truthfulqa_mc2.csv};
  \addplot[colorA!15, draw=none, forget plot] fill between[of=utruth0 and ltruth0];
  \addplot[colorA, line width=0.8pt, dashed]
    table[x=epoch, y=cfg0d, col sep=comma]{csv_tikz/llama-qwen/truthfulqa_mc2.csv};
  \addplot[name path=utruth1, draw=none, forget plot]
    table[x=epoch, y expr=\thisrow{cfg1d}+\thisrow{cfg1s}, col sep=comma]{csv_tikz/llama-qwen/truthfulqa_mc2.csv};
  \addplot[name path=ltruth1, draw=none, forget plot]
    table[x=epoch, y expr=\thisrow{cfg1d}-\thisrow{cfg1s}, col sep=comma]{csv_tikz/llama-qwen/truthfulqa_mc2.csv};
  \addplot[colorB!15, draw=none, forget plot] fill between[of=utruth1 and ltruth1];
  \addplot[colorB, line width=0.8pt]
    table[x=epoch, y=cfg1d, col sep=comma]{csv_tikz/llama-qwen/truthfulqa_mc2.csv};
  \addplot[name path=utruth2, draw=none, forget plot]
    table[x=epoch, y expr=\thisrow{cfg2d}+\thisrow{cfg2s}, col sep=comma]{csv_tikz/llama-qwen/truthfulqa_mc2.csv};
  \addplot[name path=ltruth2, draw=none, forget plot]
    table[x=epoch, y expr=\thisrow{cfg2d}-\thisrow{cfg2s}, col sep=comma]{csv_tikz/llama-qwen/truthfulqa_mc2.csv};
  \addplot[colorC!15, draw=none, forget plot] fill between[of=utruth2 and ltruth2];
  \addplot[colorC, line width=0.8pt]
    table[x=epoch, y=cfg2d, col sep=comma]{csv_tikz/llama-qwen/truthfulqa_mc2.csv};
  \addplot[name path=utruth3, draw=none, forget plot]
    table[x=epoch, y expr=\thisrow{cfg3d}+\thisrow{cfg3s}, col sep=comma]{csv_tikz/llama-qwen/truthfulqa_mc2.csv};
  \addplot[name path=ltruth3, draw=none, forget plot]
    table[x=epoch, y expr=\thisrow{cfg3d}-\thisrow{cfg3s}, col sep=comma]{csv_tikz/llama-qwen/truthfulqa_mc2.csv};
  \addplot[colorD!15, draw=none, forget plot] fill between[of=utruth3 and ltruth3];
  \addplot[colorD, line width=0.8pt]
    table[x=epoch, y=cfg3d, col sep=comma]{csv_tikz/llama-qwen/truthfulqa_mc2.csv};

  \nextgroupplot[
    title={\tiny Math},
    ylabel={},
    xlabel={\tiny Epoch},
    axis x line*=bottom, axis y line*=left,
    xmin=0, xmax=40, ymin=-1.8, ymax=1.2,
    grid, grid style={line width=.2pt, draw=gray!30},
  ]
  \draw[gray,dashed,line width=0.5pt](axis cs:0,0)--(axis cs:40,0);
  \addplot[name path=uminer0, draw=none, forget plot]
    table[x=epoch, y expr=\thisrow{cfg0d}+\thisrow{cfg0s}, col sep=comma]{csv_tikz/llama-qwen/minerva_math.csv};
  \addplot[name path=lminer0, draw=none, forget plot]
    table[x=epoch, y expr=\thisrow{cfg0d}-\thisrow{cfg0s}, col sep=comma]{csv_tikz/llama-qwen/minerva_math.csv};
  \addplot[colorA!15, draw=none, forget plot] fill between[of=uminer0 and lminer0];
  \addplot[colorA, line width=0.8pt, dashed]
    table[x=epoch, y=cfg0d, col sep=comma]{csv_tikz/llama-qwen/minerva_math.csv};
  \addplot[name path=uminer1, draw=none, forget plot]
    table[x=epoch, y expr=\thisrow{cfg1d}+\thisrow{cfg1s}, col sep=comma]{csv_tikz/llama-qwen/minerva_math.csv};
  \addplot[name path=lminer1, draw=none, forget plot]
    table[x=epoch, y expr=\thisrow{cfg1d}-\thisrow{cfg1s}, col sep=comma]{csv_tikz/llama-qwen/minerva_math.csv};
  \addplot[colorB!15, draw=none, forget plot] fill between[of=uminer1 and lminer1];
  \addplot[colorB, line width=0.8pt]
    table[x=epoch, y=cfg1d, col sep=comma]{csv_tikz/llama-qwen/minerva_math.csv};
  \addplot[name path=uminer2, draw=none, forget plot]
    table[x=epoch, y expr=\thisrow{cfg2d}+\thisrow{cfg2s}, col sep=comma]{csv_tikz/llama-qwen/minerva_math.csv};
  \addplot[name path=lminer2, draw=none, forget plot]
    table[x=epoch, y expr=\thisrow{cfg2d}-\thisrow{cfg2s}, col sep=comma]{csv_tikz/llama-qwen/minerva_math.csv};
  \addplot[colorC!15, draw=none, forget plot] fill between[of=uminer2 and lminer2];
  \addplot[colorC, line width=0.8pt]
    table[x=epoch, y=cfg2d, col sep=comma]{csv_tikz/llama-qwen/minerva_math.csv};
  \addplot[name path=uminer3, draw=none, forget plot]
    table[x=epoch, y expr=\thisrow{cfg3d}+\thisrow{cfg3s}, col sep=comma]{csv_tikz/llama-qwen/minerva_math.csv};
  \addplot[name path=lminer3, draw=none, forget plot]
    table[x=epoch, y expr=\thisrow{cfg3d}-\thisrow{cfg3s}, col sep=comma]{csv_tikz/llama-qwen/minerva_math.csv};
  \addplot[colorD!15, draw=none, forget plot] fill between[of=uminer3 and lminer3];
  \addplot[colorD, line width=0.8pt]
    table[x=epoch, y=cfg3d, col sep=comma]{csv_tikz/llama-qwen/minerva_math.csv};

  \nextgroupplot[
    title={\tiny HumanEval},
    ylabel={},
    xlabel={\tiny Epoch},
    axis x line*=bottom, axis y line*=left,
    xmin=0, xmax=40, ymin=-8.0, ymax=1.0,
    grid, grid style={line width=.2pt, draw=gray!30},
    legend to name=figlegendllama,
    legend style={draw=gray!60, fill=white,
      nodes={scale=0.50, transform shape},
      text=black, cells={align=left}, row sep=0pt,
      inner sep=3pt, rounded corners=2pt},
  ]
  \draw[gray,dashed,line width=0.5pt](axis cs:0,0)--(axis cs:40,0);
  \addplot[name path=uhuman0, draw=none, forget plot]
    table[x=epoch, y expr=\thisrow{cfg0d}+\thisrow{cfg0s}, col sep=comma]{csv_tikz/llama-qwen/humaneval_1.csv};
  \addplot[name path=lhuman0, draw=none, forget plot]
    table[x=epoch, y expr=\thisrow{cfg0d}-\thisrow{cfg0s}, col sep=comma]{csv_tikz/llama-qwen/humaneval_1.csv};
  \addplot[colorA!15, draw=none, forget plot] fill between[of=uhuman0 and lhuman0];
  \addplot[colorA, line width=0.8pt, dashed]
    table[x=epoch, y=cfg0d, col sep=comma]{csv_tikz/llama-qwen/humaneval_1.csv};
  \addlegendentry{Llama-3.2-1B self $\tau{=}1.25$};
  \addplot[name path=uhuman1, draw=none, forget plot]
    table[x=epoch, y expr=\thisrow{cfg1d}+\thisrow{cfg1s}, col sep=comma]{csv_tikz/llama-qwen/humaneval_1.csv};
  \addplot[name path=lhuman1, draw=none, forget plot]
    table[x=epoch, y expr=\thisrow{cfg1d}-\thisrow{cfg1s}, col sep=comma]{csv_tikz/llama-qwen/humaneval_1.csv};
  \addplot[colorB!15, draw=none, forget plot] fill between[of=uhuman1 and lhuman1];
  \addplot[colorB, line width=0.8pt]
    table[x=epoch, y=cfg1d, col sep=comma]{csv_tikz/llama-qwen/humaneval_1.csv};
  \addlegendentry{Qwen2.5-0.5B $\tau{=}1.25$};
  \addplot[name path=uhuman2, draw=none, forget plot]
    table[x=epoch, y expr=\thisrow{cfg2d}+\thisrow{cfg2s}, col sep=comma]{csv_tikz/llama-qwen/humaneval_1.csv};
  \addplot[name path=lhuman2, draw=none, forget plot]
    table[x=epoch, y expr=\thisrow{cfg2d}-\thisrow{cfg2s}, col sep=comma]{csv_tikz/llama-qwen/humaneval_1.csv};
  \addplot[colorC!15, draw=none, forget plot] fill between[of=uhuman2 and lhuman2];
  \addplot[colorC, line width=0.8pt]
    table[x=epoch, y=cfg2d, col sep=comma]{csv_tikz/llama-qwen/humaneval_1.csv};
  \addlegendentry{Qwen2.5-0.5B  $\tau{=}1.0$};
  \addplot[name path=uhuman3, draw=none, forget plot]
    table[x=epoch, y expr=\thisrow{cfg3d}+\thisrow{cfg3s}, col sep=comma]{csv_tikz/llama-qwen/humaneval_1.csv};
  \addplot[name path=lhuman3, draw=none, forget plot]
    table[x=epoch, y expr=\thisrow{cfg3d}-\thisrow{cfg3s}, col sep=comma]{csv_tikz/llama-qwen/humaneval_1.csv};
  \addplot[colorD!15, draw=none, forget plot] fill between[of=uhuman3 and lhuman3];
  \addplot[colorD, line width=0.8pt]
    table[x=epoch, y=cfg3d, col sep=comma]{csv_tikz/llama-qwen/humaneval_1.csv};
  \addlegendentry{Qwen2.5-0.5B  $\tau{=}0.75$};

  \nextgroupplot[axis lines=none, xtick=\empty, ytick=\empty,
    xmin=0, xmax=1, ymin=0, ymax=1, xlabel={}, ylabel={},]

\end{groupplot}
\node[anchor=center] at (group c4r2.center) {\pgfplotslegendfromname{figlegendllama}};
\end{tikzpicture}
}
\caption{\small \textbf{Qwen data is not a generic upgrade: a LLaMA student trained on Qwen teacher data fails to outperform LLaMA-self, confirming that utility is tied to student--source compatibility.}
$\Delta$ performance of the Llama-3.2-1B student trained on self-generated data vs.\ Qwen2.5-0.5B teacher data at three temperatures over 40 epochs. Dashed line: self-generated (Llama-3.2-1B, $\tau{=}1.25$). Shaded bands show $\pm 1$ std.}
\label{fig:llama_qwen}
\end{figure}

%% file: figure_nll_llama.tex
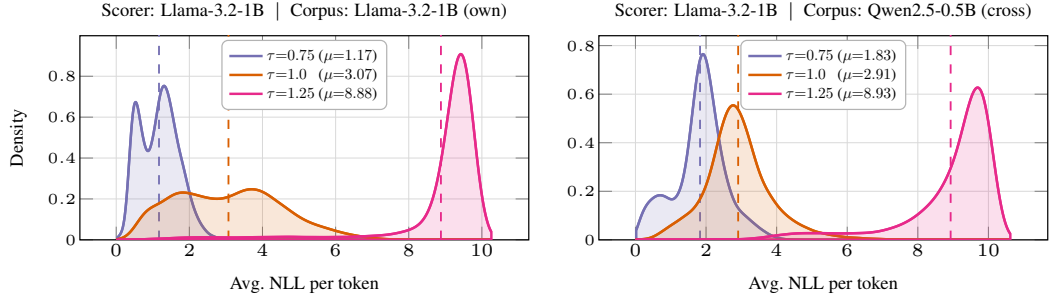
\begin{figure}[h]
\centering
\resizebox{\textwidth}{!}{%
\begin{tikzpicture}
\begin{groupplot}[
  group style={group size=2 by 1, horizontal sep=8mm},
  width=0.48\linewidth, height=0.28\linewidth,
  label style={font=\tiny},
  title style={font=\tiny, yshift=-1ex},
  tick label style={font=\fontsize{5}{5}\selectfont},
  ymin=0,
  grid, grid style={line width=.2pt, draw=gray!30},
  xlabel={\tiny Avg.\ NLL per token},
  scaled y ticks=false,
  yticklabel style={/pgf/number format/fixed, font=\fontsize{5}{5}\selectfont},
]

\nextgroupplot[
  title={\tiny Scorer: Llama-3.2-1B \ $|$ \ Corpus: Llama-3.2-1B (own)},
  ylabel={\tiny Density},
  legend style={at={(0.50,0.98)}, anchor=north,
    nodes={scale=0.50, transform shape}, draw=gray!60, fill=white,
    cells={align=left}, row sep=0pt, inner sep=2pt, rounded corners=2pt},
]
\addplot[colorC, fill=colorC, fill opacity=0.15, line width=0.8pt, smooth]
  table[x=x, y=t075, col sep=comma]{csv_tikz/nll/llama_own.csv} \closedcycle;
\addplot[colorC, line width=0.8pt, smooth, forget plot]
  table[x=x, y=t075, col sep=comma]{csv_tikz/nll/llama_own.csv};
\draw[colorC, dashed, line width=0.6pt] (axis cs:1.17,0) -- (axis cs:1.17,10);
\addplot[colorB, fill=colorB, fill opacity=0.15, line width=0.8pt, smooth]
  table[x=x, y=t10, col sep=comma]{csv_tikz/nll/llama_own.csv} \closedcycle;
\addplot[colorB, line width=0.8pt, smooth, forget plot]
  table[x=x, y=t10, col sep=comma]{csv_tikz/nll/llama_own.csv};
\draw[colorB, dashed, line width=0.6pt] (axis cs:3.07,0) -- (axis cs:3.07,10);
\addplot[colorD, fill=colorD, fill opacity=0.15, line width=0.8pt, smooth]
  table[x=x, y=t125, col sep=comma]{csv_tikz/nll/llama_own.csv} \closedcycle;
\addplot[colorD, line width=0.8pt, smooth, forget plot]
  table[x=x, y=t125, col sep=comma]{csv_tikz/nll/llama_own.csv};
\draw[colorD, dashed, line width=0.6pt] (axis cs:8.88,0) -- (axis cs:8.88,10);
\addlegendentry{$\tau{=}0.75$ ($\mu{=}1.17$)};
\addlegendentry{$\tau{=}1.0$\phantom{0} ($\mu{=}3.07$)};
\addlegendentry{$\tau{=}1.25$ ($\mu{=}8.88$)};

\nextgroupplot[
  title={\tiny Scorer: Llama-3.2-1B \ $|$ \ Corpus: Qwen2.5-0.5B (cross)},
  ylabel={},
  legend style={at={(0.50,0.98)}, anchor=north,
    nodes={scale=0.50, transform shape}, draw=gray!60, fill=white,
    cells={align=left}, row sep=0pt, inner sep=2pt, rounded corners=2pt},
]
\addplot[colorC, fill=colorC, fill opacity=0.15, line width=0.8pt, smooth]
  table[x=x, y=t075, col sep=comma]{csv_tikz/nll/llama_cross.csv} \closedcycle;
\addplot[colorC, line width=0.8pt, smooth, forget plot]
  table[x=x, y=t075, col sep=comma]{csv_tikz/nll/llama_cross.csv};
\draw[colorC, dashed, line width=0.6pt] (axis cs:1.83,0) -- (axis cs:1.83,10);
\addplot[colorB, fill=colorB, fill opacity=0.15, line width=0.8pt, smooth]
  table[x=x, y=t10, col sep=comma]{csv_tikz/nll/llama_cross.csv} \closedcycle;
\addplot[colorB, line width=0.8pt, smooth, forget plot]
  table[x=x, y=t10, col sep=comma]{csv_tikz/nll/llama_cross.csv};
\draw[colorB, dashed, line width=0.6pt] (axis cs:2.91,0) -- (axis cs:2.91,10);
\addplot[colorD, fill=colorD, fill opacity=0.15, line width=0.8pt, smooth]
  table[x=x, y=t125, col sep=comma]{csv_tikz/nll/llama_cross.csv} \closedcycle;
\addplot[colorD, line width=0.8pt, smooth, forget plot]
  table[x=x, y=t125, col sep=comma]{csv_tikz/nll/llama_cross.csv};
\draw[colorD, dashed, line width=0.6pt] (axis cs:8.93,0) -- (axis cs:8.93,10);
\addlegendentry{$\tau{=}0.75$ ($\mu{=}1.83$)};
\addlegendentry{$\tau{=}1.0$\phantom{0} ($\mu{=}2.91$)};
\addlegendentry{$\tau{=}1.25$ ($\mu{=}8.93$)};

\end{groupplot}
\end{tikzpicture}
}
\caption{\small \textbf{The NLL convergence pattern mirrors the Qwen-scorer case: own and cross corpora become indistinguishable in likelihood at high temperature yet produce different downstream effects.}
KDE of average NLL per token for synthetic corpora generated by Llama-3.2-1B (left, own) and Qwen2.5-0.5B (right, cross), all scored by Llama-3.2-1B. Dashed lines indicate distribution means. The pattern mirrors the Qwen-scorer case in Figure~\ref{fig:nll_dist}.}
 
\label{fig:nll_dist_llama}
\end{figure}

%% file: reference.bib
@inproceedings{alemohammad2024self,
  title = {Self-Consuming Generative Models Go {MAD}},
  author = {Sina Alemohammad and Josue Casco-Rodriguez and Lorenzo Luzi and Ahmed Imtiaz Humayun and Hossein Babaei and Daniel LeJeune and Ali Siahkoohi and Richard G. Baraniuk},
  booktitle = {International Conference on Learning Representations},
  year = {2024},
  eprint = {2307.01850},
  archivePrefix = {arXiv},
  primaryClass = {cs.LG},
  url = {https://openreview.net/forum?id=ShjMHfmPs0}
}

@inproceedings{biderman2023pythia,
  title = {Pythia: A Suite for Analyzing Large Language Models Across Training and Scaling},
  author = {Stella Biderman and Hailey Schoelkopf and Quentin Gregory Anthony and Herbie Bradley and Kyle O'Brien and Eric Hallahan and Mohammad Aflah Khan and Shivanshu Purohit and Usvsn Sai Prashanth and Edward Raff and Aviya Skowron and Lintang Sutawika and Oskar Van Der Wal},
  booktitle = {Proceedings of the 40th International Conference on Machine Learning},
  series = {Proceedings of Machine Learning Research},
  volume = {202},
  pages = {2397--2430},
  publisher = {PMLR},
  year = {2023},
  url = {https://proceedings.mlr.press/v202/biderman23a.html}
}

@inproceedings{carlini2021extracting,
  title = {Extracting Training Data from Large Language Models},
  author = {Nicholas Carlini and Florian Tram{\`e}r and Eric Wallace and Matthew Jagielski and Ariel Herbert-Voss and Katherine Lee and Adam Roberts and Tom Brown and Dawn Song and {\'U}lfar Erlingsson and Alina Oprea and Colin Raffel},
  booktitle = {30th USENIX Security Symposium (USENIX Security 21)},
  pages = {2633--2650},
  publisher = {USENIX Association},
  year = {2021},
  eprint = {2012.07805},
  archivePrefix = {arXiv},
  primaryClass = {cs.CR},
  url = {https://www.usenix.org/conference/usenixsecurity21/presentation/carlini-extracting}
}

@inproceedings{feng2025beyond,
  title = {Beyond Model Collapse: Scaling Up with Synthesized Data Requires Verification},
  author = {Yunzhen Feng and Elvis Dohmatob and Pu Yang and Francois Charton and Julia Kempe},
  booktitle = {International Conference on Learning Representations},
  year = {2025},
  eprint = {2406.07515},
  archivePrefix = {arXiv},
  primaryClass = {cs.LG},
  url = {https://openreview.net/forum?id=MQXrTMonT1}
}

@inproceedings{karan2025reasoning,
  title = {Reasoning with Sampling: Your Base Model is Smarter Than You Think},
  author = {Aayush Karan and Yilun Du},
  booktitle = {International Conference on Learning Representations},
  year = {2026},
  eprint = {2510.14901},
  archivePrefix = {arXiv},
  primaryClass = {cs.LG},
  url = {https://openreview.net/forum?id=Vsgq2ldr4K}
}

@misc{li2025selftrain,
  title = {A Model Can Help Itself: Reward-Free Self-Training for {LLM} Reasoning},
  author = {Mengqi Li and Lei Zhao and Anthony Man-Cho So and Ruoyu Sun and Xiao Li},
  year = {2025},
  eprint = {2510.18814},
  archivePrefix = {arXiv},
  primaryClass = {cs.LG},
  url = {https://arxiv.org/abs/2510.18814}
}

@misc{shao2025spurious,
  title = {Spurious Rewards: Rethinking Training Signals in {RLVR}},
  author = {Rulin Shao and Shuyue Stella Li and Rui Xin and Scott Geng and Yiping Wang and Sewoong Oh and Simon Shaolei Du and Nathan Lambert and Sewon Min and Ranjay Krishna and Yulia Tsvetkov and Hannaneh Hajishirzi and Pang Wei Koh and Luke Zettlemoyer},
  year = {2025},
  eprint = {2506.10947},
  archivePrefix = {arXiv},
  primaryClass = {cs.AI},
  url = {https://arxiv.org/abs/2506.10947}
}

@article{shumailov2023curse,
  title = {{AI} models collapse when trained on recursively generated data},
  author = {Ilia Shumailov and Zakhar Shumaylov and Yiren Zhao and Nicolas Papernot and Ross Anderson and Yarin Gal},
  journal = {Nature},
  volume = {631},
  pages = {755--759},
  year = {2024},
  doi = {10.1038/s41586-024-07566-y},
  url = {https://www.nature.com/articles/s41586-024-07566-y}
}

@article{singh2024beyond,
  title = {Beyond Human Data: Scaling Self-Training for Problem-Solving with Language Models},
  author = {Avi Singh and John D. Co-Reyes and Rishabh Agarwal and Ankesh Anand and Piyush Patil and Xavier Garcia and Peter J. Liu and James Harrison and Jaehoon Lee and Kelvin Xu and Aaron Parisi and Abhishek Kumar and Alexander A. Alemi and Alex Rizkowsky and Azade Nova and Ben Adlam and Bernd Bohnet and Gamaleldin Fathy Elsayed and Hanie Sedghi and Igor Mordatch and Isabelle Simpson and Izzeddin Gur and Jasper Snoek and Jeffrey Pennington and Jiri Hron and Kathleen Kenealy and Kevin Swersky and Kshiteej Mahajan and Laura Culp and Lechao Xiao and Maxwell L. Bileschi and Noah Constant and Roman Novak and Rosanne Liu and Tris Warkentin and Yundi Qian and Yamini Bansal and Ethan Dyer and Behnam Neyshabur and Jascha Sohl-Dickstein and Noah Fiedel},
  journal = {Transactions on Machine Learning Research},
  year = {2024},
  url = {https://openreview.net/forum?id=lNAyUngGFK}
}

@inproceedings{zelikman2022star,
  title = {{STaR}: Bootstrapping Reasoning With Reasoning},
  author = {Eric Zelikman and Yuhuai Wu and Jesse Mu and Noah Goodman},
  booktitle = {Advances in Neural Information Processing Systems 35},
  year = {2022},
  eprint = {2203.14465},
  archivePrefix = {arXiv},
  primaryClass = {cs.LG},
  url = {https://papers.nips.cc/paper_files/paper/2022/hash/639a9a172c044fbb64175b5fad42e9a5-Abstract-Conference.html}
}

@misc{zhang2026selfdistill,
  title = {Embarrassingly Simple Self-Distillation Improves Code Generation},
  author = {Ruixiang Zhang and Richard He Bai and Huangjie Zheng and Navdeep Jaitly and Ronan Collobert and Yizhe Zhang},
  year = {2026},
  eprint = {2604.01193},
  archivePrefix = {arXiv},
  primaryClass = {cs.CL},
  url = {https://arxiv.org/abs/2604.01193}
}

@inproceedings{hendrycks2021mmlu,
  title = {Measuring Massive Multitask Language Understanding},
  author = {Dan Hendrycks and Collin Burns and Steven Basart and Andy Zou and Mantas Mazeika and Dawn Song and Jacob Steinhardt},
  booktitle = {International Conference on Learning Representations},
  year = {2021},
  eprint = {2009.03300},
  archivePrefix = {arXiv},
  primaryClass = {cs.CL},
  url = {https://openreview.net/forum?id=d7KBjmI3GmQ}
}

@misc{clark2018arc,
  title = {Think you have Solved Question Answering? Try {ARC}, the {AI2} Reasoning Challenge},
  author = {Peter Clark and Isaac Cowhey and Oren Etzioni and Tushar Khot and Ashish Sabharwal and Carissa Schoenick and Oyvind Tafjord},
  year = {2018},
  eprint = {1803.05457},
  archivePrefix = {arXiv},
  primaryClass = {cs.AI},
  url = {https://arxiv.org/abs/1803.05457}
}

@misc{cobbe2021gsm8k,
  title = {Training Verifiers to Solve Math Word Problems},
  author = {Karl Cobbe and Vineet Kosaraju and Mohammad Bavarian and Mark Chen and Heewoo Jun and Lukasz Kaiser and Matthias Plappert and Jerry Tworek and Jacob Hilton and Reiichiro Nakano and Christopher Hesse and John Schulman},
  year = {2021},
  eprint = {2110.14168},
  archivePrefix = {arXiv},
  primaryClass = {cs.LG},
  url = {https://arxiv.org/abs/2110.14168}
}

@inproceedings{zellers2019hellaswag,
  title = {{HellaSwag}: Can a Machine Really Finish Your Sentence?},
  author = {Rowan Zellers and Ari Holtzman and Yonatan Bisk and Ali Farhadi and Yejin Choi},
  booktitle = {Proceedings of the 57th Annual Meeting of the Association for Computational Linguistics},
  pages = {4791--4800},
  publisher = {Association for Computational Linguistics},
  year = {2019},
  doi = {10.18653/v1/P19-1472},
  url = {https://aclanthology.org/P19-1472/}
}

@inproceedings{lin2022truthfulqa,
  title = {{TruthfulQA}: Measuring How Models Mimic Human Falsehoods},
  author = {Stephanie Lin and Jacob Hilton and Owain Evans},
  booktitle = {Proceedings of the 60th Annual Meeting of the Association for Computational Linguistics (Volume 1: Long Papers)},
  pages = {3214--3252},
  publisher = {Association for Computational Linguistics},
  year = {2022},
  doi = {10.18653/v1/2022.acl-long.229},
  url = {https://aclanthology.org/2022.acl-long.229/}
}

@inproceedings{lewkowycz2022minerva,
  title = {Solving Quantitative Reasoning Problems with Language Models},
  author = {Aitor Lewkowycz and Anders Andreassen and David Dohan and Ethan Dyer and Henryk Michalewski and Vinay Ramasesh and Ambrose Slone and Cem Anil and Imanol Schlag and Theo Gutman-Solo and Yuhuai Wu and Behnam Neyshabur and Guy Gur-Ari and Vedant Misra},
  booktitle = {Advances in Neural Information Processing Systems 35},
  year = {2022},
  url = {https://papers.nips.cc/paper_files/paper/2022/hash/18abbeef8cfe9203fdf9053c9c4fe191-Abstract-Conference.html}
}

@misc{chen2021humaneval,
  title = {Evaluating Large Language Models Trained on Code},
  author = {Mark Chen and Jerry Tworek and Heewoo Jun and Qiming Yuan and Henrique Ponde de Oliveira Pinto and Jared Kaplan and Harri Edwards and Yuri Burda and Nicholas Joseph and Greg Brockman and Alex Ray and Raul Puri and Gretchen Krueger and Michael Petrov and Heidy Khlaaf and Girish Sastry and Pamela Mishkin and Brooke Chan and Scott Gray and Nick Ryder and Mikhail Pavlov and Alethea Power and Lukasz Kaiser and Mohammad Bavarian and Clemens Winter and Philippe Tillet and Felipe Petroski Such and Dave Cummings and Matthias Plappert and Fotios Chantzis and Elizabeth Barnes and Ariel Herbert-Voss and William Hebgen Guss and Alex Nichol and Alex Paino and Nikolas Tezak and Jie Tang and Igor Babuschkin and Suchir Balaji and Shantanu Jain and William Saunders and Christopher Hesse and Andrew N. Carr and Jan Leike and Josh Achiam and Vedant Misra and Evan Morikawa and Alec Radford and Matthew Knight and Miles Brundage and Mira Murati and Katie Mayer and Peter Welinder and Bob McGrew and Dario Amodei and Sam McCandlish and Ilya Sutskever and Wojciech Zaremba},
  year = {2021},
  eprint = {2107.03374},
  archivePrefix = {arXiv},
  primaryClass = {cs.LG},
  url = {https://arxiv.org/abs/2107.03374}
}

@misc{spiesberger2026soft,
  title = {Soft Contamination Means Benchmarks Test Shallow Generalization},
  author = {Ari Spiesberger and Juan J. Vazquez and Nicky Pochinkov and Tom{\'a}{\v{s}} Gaven{\v{c}}iak and Peli Grietzer and Gavin Leech and Nandi Schoots},
  year = {2026},
  eprint = {2602.12413},
  archivePrefix = {arXiv},
  primaryClass = {cs.LG},
  url = {https://arxiv.org/abs/2602.12413}
}

@misc{babakhin2025llamaembed,
  title = {{Llama-Embed-Nemotron-8B}: A Universal Text Embedding Model for Multilingual and Cross-Lingual Tasks},
  author = {Yauhen Babakhin and Radek Osmulski and Ronay Ak and Gabriel de Souza Pereira Moreira and Mengyao Xu and Benedikt Schifferer and Bo Liu and Even Oldridge},
  year = {2025},
  eprint = {2511.07025},
  archivePrefix = {arXiv},
  primaryClass = {cs.CL},
  url = {https://arxiv.org/abs/2511.07025}
}

@inproceedings{carlini2022quantifying,
  title = {Quantifying Memorization Across Neural Language Models},
  author = {Nicholas Carlini and Daphne Ippolito and Matthew Jagielski and Katherine Lee and Florian Tram{\`e}r and Chiyuan Zhang},
  booktitle = {International Conference on Learning Representations},
  year = {2023},
  eprint = {2202.07646},
  archivePrefix = {arXiv},
  primaryClass = {cs.LG},
  url = {https://openreview.net/forum?id=TatRHT_1cK}
}

@misc{alemohammad2024selfimproving,
  title = {Self-Improving Diffusion Models With Synthetic Data},
  author = {Sina Alemohammad and Ahmed Imtiaz Humayun and Shruti Agarwal and John Collomosse and Richard G. Baraniuk},
  year = {2024},
  eprint = {2408.16333},
  archivePrefix = {arXiv},
  primaryClass = {cs.LG},
  url = {https://arxiv.org/abs/2408.16333}
}

@inproceedings{alemohammad2025neon,
  title = {Neon: Negative Extrapolation From Self-Training Improves Image Generation},
  author = {Sina Alemohammad and Zhangyang Wang and Richard G. Baraniuk},
  booktitle = {International Conference on Learning Representations},
  year = {2026},
  eprint = {2510.03597},
  archivePrefix = {arXiv},
  primaryClass = {cs.GR},
  url = {https://openreview.net/forum?id=kpLRYtPGt3}
}

@misc{ji2026scalable,
  title = {Scalable Power Sampling: Unlocking Efficient, Training-Free Reasoning for {LLMs} via Distribution Sharpening},
  author = {Xiaotong Ji and Rasul Tutunov and Matthieu Zimmer and Haitham Bou Ammar},
  year = {2026},
  eprint = {2601.21590},
  archivePrefix = {arXiv},
  primaryClass = {cs.LG},
  doi = {10.48550/arXiv.2601.21590},
  url = {https://arxiv.org/abs/2601.21590}
}

@misc{tan2026prerl,
  title = {From $P(y|x)$ to $P(y)$: Investigating Reinforcement Learning in Pre-train Space},
  author = {Yuqiao Tan and Minzheng Wang and Bo Liu and Zichen Liu and Tian Liang and Shizhu He and Jun Zhao and Kang Liu},
  year = {2026},
  eprint = {2604.14142},
  archivePrefix = {arXiv},
  primaryClass = {cs.LG},
  doi = {10.48550/arXiv.2604.14142},
  url = {https://arxiv.org/abs/2604.14142}
}

@inproceedings{jang2022knowledge,
  title = {Knowledge Unlearning for Mitigating Privacy Risks in Language Models},
  author = {Joel Jang and Dongkeun Yoon and Sohee Yang and Sungmin Cha and Moontae Lee and Lajanugen Logeswaran and Minjoon Seo},
  booktitle = {Proceedings of the 61st Annual Meeting of the Association for Computational Linguistics (Volume 1: Long Papers)},
  pages = {14389--14408},
  address = {Toronto, Canada},
  publisher = {Association for Computational Linguistics},
  year = {2023},
  doi = {10.18653/v1/2023.acl-long.805},
  url = {https://aclanthology.org/2023.acl-long.805/}
}

@inproceedings{maini2024tofu,
  title = {{TOFU}: A Task of Fictitious Unlearning for {LLMs}},
  author = {Pratyush Maini and Zhili Feng and Avi Schwarzschild and Zachary C. Lipton and J. Zico Kolter},
  booktitle = {First Conference on Language Modeling},
  year = {2024},
  eprint = {2401.06121},
  archivePrefix = {arXiv},
  primaryClass = {cs.LG},
  doi = {10.48550/arXiv.2401.06121},
  url = {https://openreview.net/forum?id=B41hNBoWLo}
}

@inproceedings{scholten2026pmc,
  title = {Model Collapse Is Not a Bug but a Feature in Machine Unlearning for {LLMs}},
  author = {Yan Scholten and Sophie Xhonneux and Leo Schwinn and Stephan G{\"u}nnemann},
  booktitle = {International Conference on Learning Representations},
  year = {2026},
  eprint = {2507.04219},
  archivePrefix = {arXiv},
  primaryClass = {cs.LG},
  url = {https://openreview.net/forum?id=1MCQzboBrR}
}

@inproceedings{zheng2024llamafactory,
  title={LlamaFactory: Unified Efficient Fine-Tuning of 100+ Language Models},
  author={Yaowei Zheng and Richong Zhang and Junhao Zhang and Yanhan Ye and Zheyan Luo and Zhangchi Feng and Yongqiang Ma},
  booktitle={Proceedings of the 62nd Annual Meeting of the Association for Computational Linguistics (Volume 3: System Demonstrations)},
  address={Bangkok, Thailand},
  publisher={Association for Computational Linguistics},
  year={2024},
  url={http://arxiv.org/abs/2403.13372}
}

@misc{eval-harness,
  author       = {Gao, Leo and Tow, Jonathan and Abbasi, Baber and Biderman, Stella and Black, Sid and DiPofi, Anthony and Foster, Charles and Golding, Laurence and Hsu, Jeffrey and Le Noac'h, Alain and Li, Haonan and McDonell, Kyle and Muennighoff, Niklas and Ociepa, Chris and Phang, Jason and Reynolds, Laria and Schoelkopf, Hailey and Skowron, Aviya and Sutawika, Lintang and Tang, Eric and Thite, Anish and Wang, Ben and Wang, Kevin and Zou, Andy},
  title        = {The Language Model Evaluation Harness},
  month        = 12,
  year         = 2023,
  publisher    = {Zenodo},
  version      = {v0.4.0},
  doi          = {10.5281/zenodo.10256836},
  url          = {https://zenodo.org/records/10256836}
}

@inproceedings{10.1145/3600006.3613165,
author = {Kwon, Woosuk and Li, Zhuohan and Zhuang, Siyuan and Sheng, Ying and Zheng, Lianmin and Yu, Cody Hao and Gonzalez, Joseph and Zhang, Hao and Stoica, Ion},
title = {Efficient Memory Management for Large Language Model Serving with PagedAttention},
year = {2023},
isbn = {9798400702297},
publisher = {Association for Computing Machinery},
address = {New York, NY, USA},
url = {https://doi.org/10.1145/3600006.3613165},
doi = {10.1145/3600006.3613165},
booktitle = {Proceedings of the 29th Symposium on Operating Systems Principles},
pages = {611–626},
numpages = {16},
location = {Koblenz, Germany},
series = {SOSP '23}
}

@INPROCEEDINGS{9355301,
  author={Rajbhandari, Samyam and Rasley, Jeff and Ruwase, Olatunji and He, Yuxiong},
  booktitle={SC20: International Conference for High Performance Computing, Networking, Storage and Analysis}, 
  title={ZeRO: Memory optimizations Toward Training Trillion Parameter Models}, 
  year={2020},
  volume={},
  number={},
  pages={1-16},
  doi={10.1109/SC41405.2020.00024}}

@inproceedings{
langlais2026common,
title={Common Corpus: The Largest Collection of Ethical Data for {LLM} Pre-Training},
author={Pierre-Carl Langlais and Pavel Chizhov and Catherine Arnett and Carlos Rosas Hinostroza and Mattia Nee and Eliot Krzysztof Jones and Ir{\`e}ne Girard and David Mach and Anastasia Stasenko and Ivan P. Yamshchikov},
booktitle={The Fourteenth International Conference on Learning Representations},
year={2026},
url={https://openreview.net/forum?id=0wSlFpMsGb}
}
